\documentclass{article}

     \usepackage[final,dandb]{neurips_2025}

\usepackage[utf8]{inputenc} 
\usepackage[T1]{fontenc}    
\usepackage{url}            
\usepackage{booktabs}       
\usepackage{amsfonts}       
\usepackage{nicefrac}       
\usepackage{microtype}      
\usepackage{xcolor}         
\usepackage{colortbl}
\usepackage{makecell}
\usepackage{multirow}
\usepackage{tabularx}
 \usepackage[misc]{ifsym}

\usepackage{subcaption}
\usepackage[most]{tcolorbox}
\usepackage{adjustbox}
\usepackage{wrapfig}
\usepackage{colortbl}
\usepackage{makecell}
\usepackage{multirow}
\usepackage{tabularx}
\usepackage{array}
\usepackage{pifont}
\usepackage{float}
\usepackage{xspace}
\usepackage{svg}
\usepackage{dsfont}
\usepackage{placeins}
\newcommand{\ourbench}{Paper2Poster}
\newcommand{\ouralg}{PosterAgent}
\newcommand{\kqa}{PaperQuiz}

\newcommand{\ie}{\textit{i.e.,}}
\newcommand{\eg}{\textit{e.g.,}}

\definecolor{citecolor}{HTML}{0071bc}
\usepackage[pagebackref=false,breaklinks=true,letterpaper=true,colorlinks,citecolor=citecolor,bookmarks=false]{hyperref}

\usepackage{etoc}
\etocdepthtag.toc{mtchapter}
\etocsettagdepth{mtchapter}{subsection}
\etocsettagdepth{mtappendix}{none}

\title{\ourbench: Towards Multimodal Poster Automation from Scientific Papers}

\author{\textsuperscript{$1,3$}Wei Pang\thanks{Equal contribution. \Letter\ Corresponding to:  xihe@uwaterloo.ca, kevin.qh.lin@gmail.com}~, 
\textsuperscript{$2$}Kevin Qinghong Lin\footnotemark[1]~~\textsuperscript{\Letter}, 
\textsuperscript{$1$}Xiangru Jian\footnotemark[1]~, 
\textsuperscript{$1,3$}Xi He\textsuperscript{\Letter},
\textsuperscript{$2$}Philip Torr
\\
\textsuperscript{$1$} University of Waterloo\quad 
\textsuperscript{$2$} University of Oxford\quad
\textsuperscript{$3$} Vector Institute
\\
Project Page: \url{https://paper2poster.github.io}
}
\definecolor{prompt}{RGB}{223, 223, 192}
\definecolor{prompt-frame}{RGB}{137, 137, 90}
\definecolor{prompt2}{RGB}{223, 223, 192}
\definecolor{prompt2-frame}{RGB}{137, 137, 90}
\definecolor{prompt3}{RGB}{212, 238, 179}
\definecolor{prompt3-frame}{RGB}{117, 146, 77}
\definecolor{prompt4}{RGB}{212, 238, 179}
\definecolor{prompt4-frame}{RGB}{117, 146, 77}
\definecolor{prompt5}{RGB}{212, 238, 179}
\definecolor{prompt5-frame}{RGB}{117, 146, 77}
\usepackage{fontawesome5}
\tcbset{
    prompt_func/.style={
        enhanced,
        breakable,
        colback=prompt,        
        colframe=prompt-frame,            
        fontupper=\normalsize,               
        boxrule=1pt,                       
        arc=4mm,                           
        left=1mm, right=1mm, top=1mm, bottom=1mm, 
        boxsep=4pt,                        
        before skip=10pt, after skip=10pt, 
        overlay={
            \node[anchor=north west, xshift=4pt, text=white] at (frame.north west) {\faDatabase};
        },
        title={~~~~~~~\textbf{Prompt used with GPT-4o to generate annotation for Element Grounding with functional setting}},
        coltitle=white,
        fonttitle=\bfseries\small
    }
}

\tcbset{
    prompt_aethetic_element/.style={
        enhanced,
        breakable,
        colback=prompt,        
        colframe=prompt-frame,            
        fontupper=\normalsize,               
        boxrule=1pt,                       
        arc=4mm,                           
        left=1mm, right=1mm, top=1mm, bottom=1mm, 
        boxsep=4pt,                        
        before skip=10pt, after skip=10pt, 
        title={\faDatabase~~~\textbf{Prompt: Element Quality Judge}},
        coltitle=white,
        fonttitle=\bfseries\small
    }
}

\tcbset{
    prompt_aethetic_Layout/.style={
        enhanced,
        breakable,
        colback=prompt,        
        colframe=prompt-frame,            
        fontupper=\normalsize,               
        boxrule=1pt,                       
        arc=4mm,                           
        left=1mm, right=1mm, top=1mm, bottom=1mm, 
        boxsep=4pt,                        
        before skip=10pt, after skip=10pt, 
        title={\faDatabase~~~\textbf{Prompt: Layout Balance Judge}},
        coltitle=white,
        fonttitle=\bfseries\small
    }
}

\tcbset{prompt_aethetic_engagement/.style={
        enhanced,
        breakable,
        colback=prompt,        
        colframe=prompt-frame,            
        fontupper=\normalsize,               
        boxrule=1pt,                       
        arc=4mm,                           
        left=1mm, right=1mm, top=1mm, bottom=1mm, 
        boxsep=4pt,                        
        before skip=10pt, after skip=10pt, 
        title={\faDatabase~~~\textbf{Prompt: Engagement Judge}},
        coltitle=white,
        fonttitle=\bfseries\small
    }
}

\tcbset{prompt_info_clarity/.style={
        enhanced,
        breakable,
        colback=prompt,        
        colframe=prompt-frame,            
        fontupper=\normalsize,               
        boxrule=1pt,                       
        arc=4mm,                           
        left=1mm, right=1mm, top=1mm, bottom=1mm, 
        boxsep=4pt,                        
        before skip=10pt, after skip=10pt, 
        title={\faDatabase~~~\textbf{Prompt: Clarity Judge}},
        coltitle=white,
        fonttitle=\bfseries\small
    }
}

\tcbset{prompt_info_content/.style={
        enhanced,
        breakable,
        colback=prompt,        
        colframe=prompt-frame,            
        fontupper=\normalsize,               
        boxrule=1pt,                       
        arc=4mm,                           
        left=1mm, right=1mm, top=1mm, bottom=1mm, 
        boxsep=4pt,                        
        before skip=10pt, after skip=10pt, 
        title={\faDatabase~~~\textbf{Prompt: Content Completeness Judge}},
        coltitle=white,
        fonttitle=\bfseries\small
    }
}

\tcbset{prompt_info_logic/.style={
        enhanced,
        breakable,
        colback=prompt,        
        colframe=prompt-frame,            
        fontupper=\normalsize,               
        boxrule=1pt,                       
        arc=4mm,                           
        left=1mm, right=1mm, top=1mm, bottom=1mm, 
        boxsep=4pt,                        
        before skip=10pt, after skip=10pt, 
        title={\faDatabase~~~\textbf{Prompt: Logical Flow Judge}},
        coltitle=white,
        fonttitle=\bfseries\small
    }
}

\tcbset{prompt_verbatim_question/.style={
        enhanced,
        breakable,
        colback=prompt,        
        colframe=prompt-frame,            
        fontupper=\normalsize,               
        boxrule=1pt,                       
        arc=4mm,                           
        left=1mm, right=1mm, top=1mm, bottom=1mm, 
        boxsep=4pt,                        
        before skip=10pt, after skip=10pt, 
        title={\faDatabase~~~\textbf{Prompt: Generate Verbatim QA}},
        coltitle=white,
        fonttitle=\bfseries\small
    }
}

\tcbset{prompt_interpretive_question/.style={
        enhanced,
        breakable,
        colback=prompt,        
        colframe=prompt-frame,            
        fontupper=\normalsize,               
        boxrule=1pt,                       
        arc=4mm,                           
        left=1mm, right=1mm, top=1mm, bottom=1mm, 
        boxsep=4pt,                        
        before skip=10pt, after skip=10pt, 
        title={\faDatabase~~~\textbf{Prompt: Generate Interpretive QA}},
        coltitle=white,
        fonttitle=\bfseries\small
    }
}

\tcbset{prompt_answer_agent/.style={
        enhanced,
        breakable,
        colback=prompt,        
        colframe=prompt-frame,            
        fontupper=\normalsize,               
        boxrule=1pt,                       
        arc=4mm,                           
        left=1mm, right=1mm, top=1mm, bottom=1mm, 
        boxsep=4pt,                        
        before skip=10pt, after skip=10pt, 
        title={\faDatabase~~~\textbf{Prompt: Answer Questions}},
        coltitle=white,
        fonttitle=\bfseries\small
    }
}

\tcbset{prompt_4o_image/.style={
        enhanced,
        breakable,
        colback=prompt,        
        colframe=prompt-frame,            
        fontupper=\normalsize,               
        boxrule=1pt,                       
        arc=4mm,                           
        left=1mm, right=1mm, top=1mm, bottom=1mm, 
        boxsep=4pt,                        
        before skip=10pt, after skip=10pt, 
        title={\faDatabase~~~\textbf{Prompt: \texttt{4o-Image}}},
        coltitle=white,
        fonttitle=\bfseries\small
    }
}

\tcbset{prompt_owl_4o/.style={
        enhanced,
        breakable,
        colback=prompt,        
        colframe=prompt-frame,            
        fontupper=\normalsize,               
        boxrule=1pt,                       
        arc=4mm,                           
        left=1mm, right=1mm, top=1mm, bottom=1mm, 
        boxsep=4pt,                        
        before skip=10pt, after skip=10pt, 
        title={\faDatabase~~~\textbf{Prompt: \texttt{OWL-4o}}},
        coltitle=white,
        fonttitle=\bfseries\small
    }
}

\tcbset{prompt_4o_html/.style={
        enhanced,
        breakable,
        colback=prompt,        
        colframe=prompt-frame,            
        fontupper=\normalsize,               
        boxrule=1pt,                       
        arc=4mm,                           
        left=1mm, right=1mm, top=1mm, bottom=1mm, 
        boxsep=4pt,                        
        before skip=10pt, after skip=10pt, 
        title={\faDatabase~~~\textbf{Prompt: \texttt{4o-HTML}}},
        coltitle=white,
        fonttitle=\bfseries\small
    }
}

\tcbset{prompt_parser_summarizer/.style={
        enhanced,
        breakable,
        colback=prompt,        
        colframe=prompt-frame,            
        fontupper=\normalsize,               
        boxrule=1pt,                       
        arc=4mm,                           
        left=1mm, right=1mm, top=1mm, bottom=1mm, 
        boxsep=4pt,                        
        before skip=10pt, after skip=10pt, 
        title={\faDatabase~~~\textbf{Prompt: Paper Summarizer}},
        coltitle=white,
        fonttitle=\bfseries\small
    }
}

\tcbset{prompt_parser_filter/.style={
        enhanced,
        breakable,
        colback=prompt,        
        colframe=prompt-frame,            
        fontupper=\normalsize,               
        boxrule=1pt,                       
        arc=4mm,                           
        left=1mm, right=1mm, top=1mm, bottom=1mm, 
        boxsep=4pt,                        
        before skip=10pt, after skip=10pt, 
        title={\faDatabase~~~\textbf{Prompt: Figure Filter}},
        coltitle=white,
        fonttitle=\bfseries\small
    }
}

\tcbset{prompt_planner_matching/.style={
        enhanced,
        breakable,
        colback=prompt,        
        colframe=prompt-frame,            
        fontupper=\normalsize,               
        boxrule=1pt,                       
        arc=4mm,                           
        left=1mm, right=1mm, top=1mm, bottom=1mm, 
        boxsep=4pt,                        
        before skip=10pt, after skip=10pt, 
        title={\faDatabase~~~\textbf{Prompt: Asset Matching}},
        coltitle=white,
        fonttitle=\bfseries\small
    }
}

\tcbset{prompt_planner_painter/.style={
        enhanced,
        breakable,
        colback=prompt,        
        colframe=prompt-frame,            
        fontupper=\normalsize,               
        boxrule=1pt,                       
        arc=4mm,                           
        left=1mm, right=1mm, top=1mm, bottom=1mm, 
        boxsep=4pt,                        
        before skip=10pt, after skip=10pt, 
        title={\faDatabase~~~\textbf{Prompt: Painter}},
        coltitle=white,
        fonttitle=\bfseries\small
    }
}

\tcbset{prompt_planner_commenter/.style={
        enhanced,
        breakable,
        colback=prompt,        
        colframe=prompt-frame,            
        fontupper=\normalsize,               
        boxrule=1pt,                       
        arc=4mm,                           
        left=1mm, right=1mm, top=1mm, bottom=1mm, 
        boxsep=4pt,                        
        before skip=10pt, after skip=10pt, 
        title={\faDatabase~~~\textbf{Prompt: Commenter}},
        coltitle=white,
        fonttitle=\bfseries\small
    }
}
\definecolor{gt_models}{RGB}{240, 240, 240}
\definecolor{direct_gen}{RGB}{200, 255, 200}
\definecolor{multi_agent}{RGB}{185, 235, 255}
\definecolor{poster_agent}{RGB}{255, 219, 187}

\begin{document}

\maketitle
\vspace{-12pt}

\begin{abstract}
Academic poster generation is a crucial yet challenging task in scientific communication, requiring the compression of long-context interleaved documents into a single, visually coherent page. 
To address this challenge, we introduce the first benchmark and metric suite for poster generation, which pairs recent conference papers with author-designed posters and evaluates outputs on 
(i) Visual Quality—semantic alignment with human posters, 
(ii) Textual Coherence—language fluency, 
(iii) Holistic Assessment—six fine-grained aesthetic and informational criteria scored by a VLM-as-judge, 
and notably (iv) \textbf{\kqa}—the poster’s ability to \textit{convey core paper content as measured by VLMs answering generated quizzes}. 
Building on this benchmark, we propose \textbf{\ouralg}, a top‐down, visual‐in‐the‐loop multi‐agent pipeline: the \textit{(a) Parser} distills the paper into a structured asset library; the \textit{(b) Planner} aligns text–visual pairs into a binary‐tree layout that preserves reading order and spatial balance; and the \textit{(c) Painter–Commenter} loop refines each panel by executing rendering code and using VLM feedback to eliminate overflow and ensure alignment.
In our comprehensive evaluation, we find that GPT‐4o outputs—though visually appealing at first glance—often exhibit noisy text and poor PaperQuiz scores, and 
we find that reader engagement is the primary aesthetic bottleneck, as human‐designed posters rely largely on visual semantics to convey meaning.
Our \textit{fully open-source} variants (\eg~based on the Qwen-2.5 series) outperform existing 4o-driven multi-agent systems across nearly all metrics, while using \textbf{$\mathbf{87\%}$ fewer tokens}. It transforms a $22$-page paper into a finalized yet editable ‘\texttt{.pptx}’ poster — all for just $\$0.005$.
These findings chart clear directions for the next generation of fully automated poster‐generation models.
The code and datasets are available at \url{https://github.com/Paper2Poster/Paper2Poster}.
\end{abstract}
\vspace{-12pt}

\section{Introduction}
\label{sec:intro}
Academic posters play a pivotal role in scientific communication, enabling rapid dissemination of key findings at conferences where attendees have only minutes to grasp core insights from the full papers.
Despite significant progress in automated slide generation -- with systems such as PPTAgent~\cite{pptagent} and D2S~\cite{d2s} pioneering text-to-slide pipelines -- poster creation~\cite{posterbot,genposter,posta} remains an underexplored and substantially more challenging task. Unlike slide decks, which distribute content across multiple, single-message slides, academic posters must condense an entire paper into a single, visually coherent page.
This requires 
\textbf{(i)} handling a much longer multi-modal context~\cite{postersum}, 
\textbf{(ii)} tightly interleaving text and graphics to convey complex ideas at a glance~\cite{posterbot,posta}, and 
\textbf{(iii)} respecting stringent spatial constraints to avoid text overflow or layout collapse~\cite{relationdif,genposter}. 
These factors make VLM- or LLM-only approaches insufficient: without explicit visual feedback like humans, it is difficult to reason about spatial layouts, maintain logical flow within a confined canvas, ensuring legibility and aesthetic.

\begin{figure}[htbp]
  \centering
  \includegraphics[width=\textwidth]{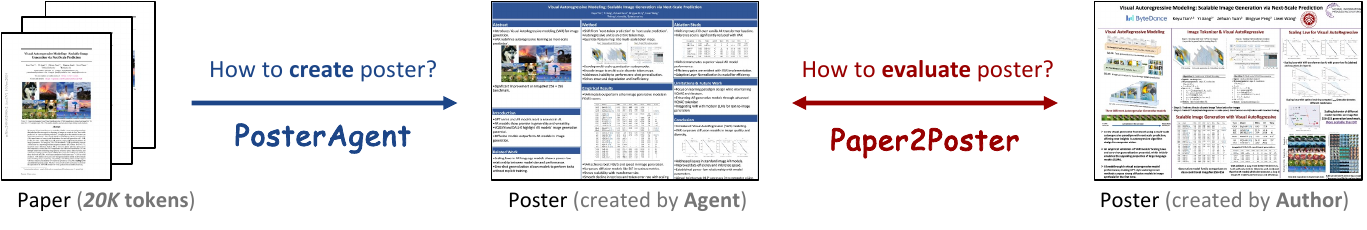}
  \vspace{-1em}
\caption{\textbf{Overview of this work.} We address two core challenges in scientific poster generation: \textbf{Left: How to create a poster from a paper}—we propose \textbf{PosterAgent} (Sec.~\ref{sec:posteragent}), a framework that transforms long-context scientific papers (20K+ tokens) into structured visual posters; and \textbf{Right: How to evaluate poster quality}—we introduce the \textbf{Paper2Poster} benchmark (Sec.~\ref{sec:paper2poster}), which enables systematic comparison between agent-generated and author-designed posters.}
\vspace{-2em}
  \label{fig:teaser}
\end{figure}

To systematically evaluate poster generation, we propose the \textbf{\ourbench} Benchmark, the first benchmark and metric suite for this novel task. Our benchmark comprises recent conference papers paired with author-designed posters, along with a human-and-model evaluation protocol that measures  
\textbf{(i)} Visual Quality — how well the generated poster aligns visually with the human-designed version.
\textbf{(ii)} Textual Coherence — the clarity and fluency of the poster’s language.
\textbf{(iii)} Holistic Assessment — the overall aesthetic and informational quality, rated across six fine-grained dimensions by VLM as Judge.
Notably, \textbf{(iv) \kqa} — motivated by the poster’s role as a bridge between authors and readers, this metric evaluates \textit{how effectively the poster alone conveys core paper content} by simulating diverse reader comprehension using VLMs to answer questions derived from the paper.

To tackle multimodal context compression in \ourbench, we introduce \textbf{\ouralg}, a multi-agent framework that first globally organizes document content and then performs panel-level refinements—while weaving visual feedback into every stage.
Starting with the \textbf{Parser}, we ingest the full paper PDF and transform it into an asset library of section-level text summaries and extracted figures and tables.
Next, the \textbf{Planner} semantically matches each synopsis to its corresponding visual asset and generates a binary‑tree layout, allocating panels by estimated content length while preserving reading order and spatial balance. 
Finally, the \textbf{Painter–Commenter} loop refines each panel: the Painter distills section‑figure pairs into concise bullet points and renders draft panels via \texttt{python‑pptx} code, and the Commenter—a VLM with zoom‑in reference prompts—provides targeted feedback to correct text overflow and spatial alignment. This \textit{top‑down, visual‑in‑the‑loop} design produces concise, coherent posters without manual tuning.  

Using \ourbench, we comprehensively evaluate human-designed (oracle) posters, state-of-the-art generative models (\eg~GPT-4o), and multi-agent solutions, revealing several key insights: \textit{(i)} GPT-4o’s outputs, though visually appealing at first glance, suffer from noisy or incoherent text, yielding high perplexity and poor PaperQuiz performance; \textit{(ii)} VLM-based judging shows the primary aesthetic bottleneck is Engagement rather than informational content, since human posters convey meaning predominantly through visual semantics; \textit{(iii)} PaperQuiz proves a reliable metric—VLM reader scores correlate closely with human evaluations, and more capable VLMs achieve higher scores on well-designed posters;
and \textit{(iv)} our Paper2Poster pipeline, built on a \textit{fully open-source} toolbox (\eg~\texttt{Qwen-2.5-VL-7B}), surpasses existing GPT-4o–based multi-agent approaches on nearly all metrics while consuming \textbf{$\mathbf{87\%}$ fewer tokens}.
Our findings illuminate pathways for the next generation of models and agent systems aimed at fully automated poster generation.
\vspace{-9pt}  

\section{Related Work}
\label{sec:relatedwork}
\subsection{Visual Design Automation}
Recent advances in multi-modal learning have driven significant progress in automating visual design tasks. These tasks commonly fall into two broad categories:
\textbf{(i) Text-rich Image Generation.} Tasks such as poster generation~\cite{posta, glyphdraw2, planrender,posterbot} have greatly benefited from diffusion-based approaches~\cite{planrender,relationdif,textatlas}, which enable the synthesis of detailed visuals conditioned on natural language descriptions. 
However, ensuring the quality and fidelity of embedded textual content via an end-to-end pixel generative model remains a major challenge, as generated text at the pixel level appears blurry and hard to read.
\textbf{(ii) Complex Visual Layouts.}
Tasks like website designing~\cite{webdraw,design2code,uilayout,bigdocs} or slide generation~\cite{pptagent,enhancepre,slidespawn,pre4human,slidegen,d2s} involve intricate visual structures and require integrating diverse components. 
To handle such complexity, mainstream approaches~\cite{pptagent,autopresent} often employ agentic workflows that rely heavily on code generation and tool usage to assemble complete visual outputs.
In contrast, our \ourbench\ addresses a more demanding yet highly practical setting: scientific visual design based on academic papers. This involves \textit{long-context, interleaved multi-modal,  inputs and outputs}, posing substantial challenges in both effectiveness and computational efficiency.

\subsection{Vision-Language Agents}
Recent progress has revealed the promising potential of LLMs beyond pure language understanding. Techniques such as ReAct \cite{react,mmreat} have demonstrated that LLMs can serve as autonomous agents, capable of solving complex tasks through step-by-step reasoning and dynamic interaction via coding \cite{opendevin, sweagent}, API function calling \cite{toolformer,octotools}, or UI interface interaction \cite{showui,uitars,uivision}. 
Despite these advances, general-purpose agents still struggle with professional tasks~\cite{videogui} as they require serious, accurate interaction and domain-specific knowledge. One closely related application is slide automation~\cite{autopresent, pptagent}, where agents translate brief textual queries into executable Python code (\eg~via \texttt{python-pptx}) to render presentation slides. 
However, our \ourbench\ setting is significantly more challenging: instead of a text prompt, we take full-length academic papers as inputs and generate compact, well-structured posters as output. This novel task requires careful design of both evaluation metrics and an effective, practical automation workflow.

\vspace{-8pt}  
\section{\ourbench\ Benchmark}
\label{sec:paper2poster}
\vspace{-1pt}
\subsection{Task Definition}
\vspace{-1pt}
Given a scientific paper composed of interleaved text, figures, and tables, the goal is to automatically generate a single-page academic poster that faithfully conveys the paper’s core content in a visually coherent and spatially efficient format. This task presents several unique challenges:
\textbf{\textit{a}. Long-Context Long-Horizon Task}: Scientific papers span multiple pages and thousands of words. Summarizing key insights while preserving coherence demands hierarchical understanding and selective abstraction. The complexity further necessitates long-horizon reasoning and multiple iterative interactions, making the task especially challenging.
\textbf{\textit{b}. Interleaved Multimodal Inputs}: Papers integrate numerous figures, tables, and charts, each semantically linked to the surrounding text. Successful poster generation demands the ability to extract, interpret, and align these multimodal elements in a contextually appropriate manner.
\textbf{\textit{c}. Layout-Aware Multimodal Outputs}: Unlike tasks focused solely on text (\eg~blog) or vision, poster generation requires producing interleaved text–image outputs within a constrained spatial layout. This necessitates joint reasoning over language, visual content, and layout to prevent overflow, imbalance, and logical misalignment.

\vspace{-1pt}
\subsection{Data Curation}
\vspace{-1pt}
\noindent \textbf{Data Source.}
We focus exclusively on AI papers for three key reasons: (1) they are relatively recent and undergo rigorous peer review, ensuring high scientific quality; (2) they offer diverse content across subfields—such as image-rich computer vision, text-centric NLP, and theory papers with numerous equations—providing a broad range of input modalities. To support this, we adopt the \textsc{PosterSum} dataset~\cite{postersum}, which contains a large collection of paper–poster pairs from recent AI conferences including ICML, NeurIPS, and ICLR (2022–2024). We specifically use the test split to reduce the risk of overlap with training data.

\noindent \textbf{Diverse Sampling.} 
Based on the initial candidate set, we apply two filtering criteria to curate high-quality data:
(1) \textbf{Length Control}: We deliberately include longer papers, including supplementary material, selecting PDFs that exceed $15$ pages and extend up to $50$ pages.
(2) \textbf{Latest Version}: We manually retrieve the most recent PDF version for each paper to ensure the dataset reflects final camera-ready submissions.
From the filtered set, we construct the final \ourbench\ dataset consisting of 100 paper–poster pairs, stratified by publication year to ensure temporal balance: $33$ pairs from 2022, $33$ from 2023, and $34$ from 2024. To further enhance diversity, we also stratify by source venue—selecting $35$ papers from NeurIPS, $37$ from ICML, and $28$ from ICLR, ensuring broad coverage across these leading conferences. 

\begin{figure}[ht]
  \centering
  \begin{subfigure}[b]{0.33\textwidth}
    \includegraphics[width=\textwidth]{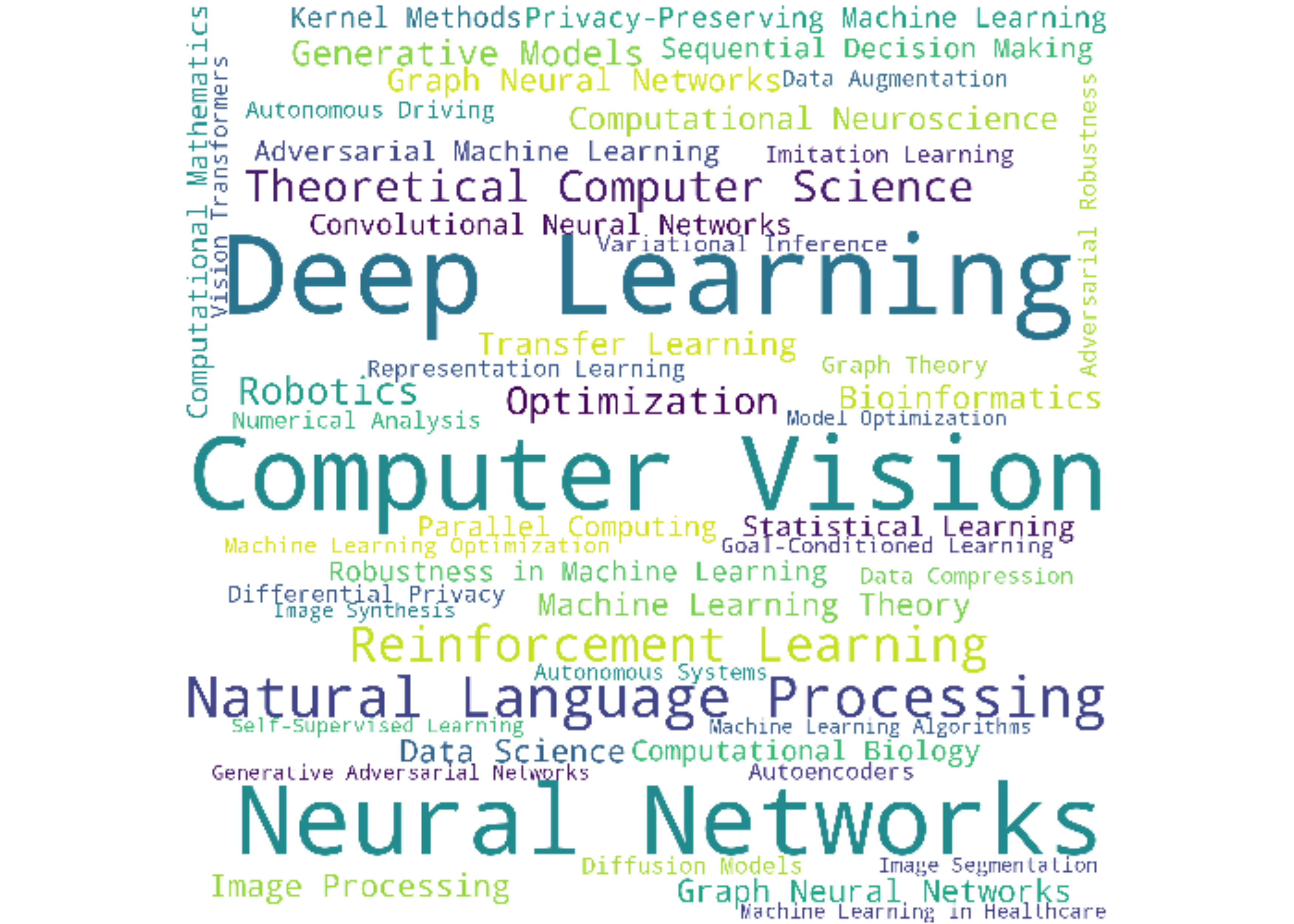}
    \vspace{-0.9em}
    \caption{Word cloud of topics}
    \label{fig:word_cloud}
  \end{subfigure}\hfill
  \begin{subfigure}[b]{0.3\textwidth}
    \includegraphics[width=\textwidth]{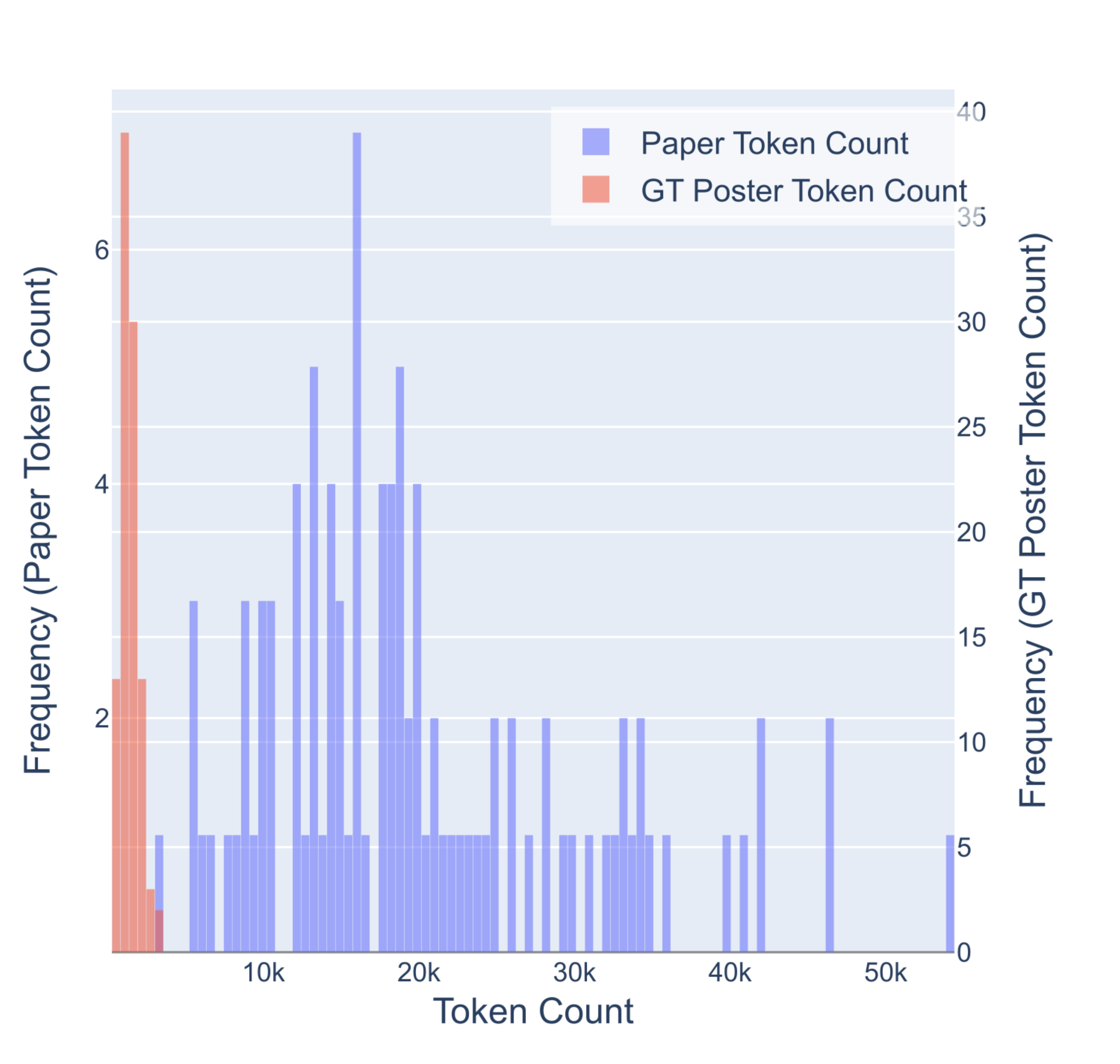}
    \vspace{-2em}
    \caption{$\#$ of tokens}
    \label{fig:token_count}
  \end{subfigure}\hfill
  \begin{subfigure}[b]{0.3\textwidth}
    \includegraphics[width=\textwidth]{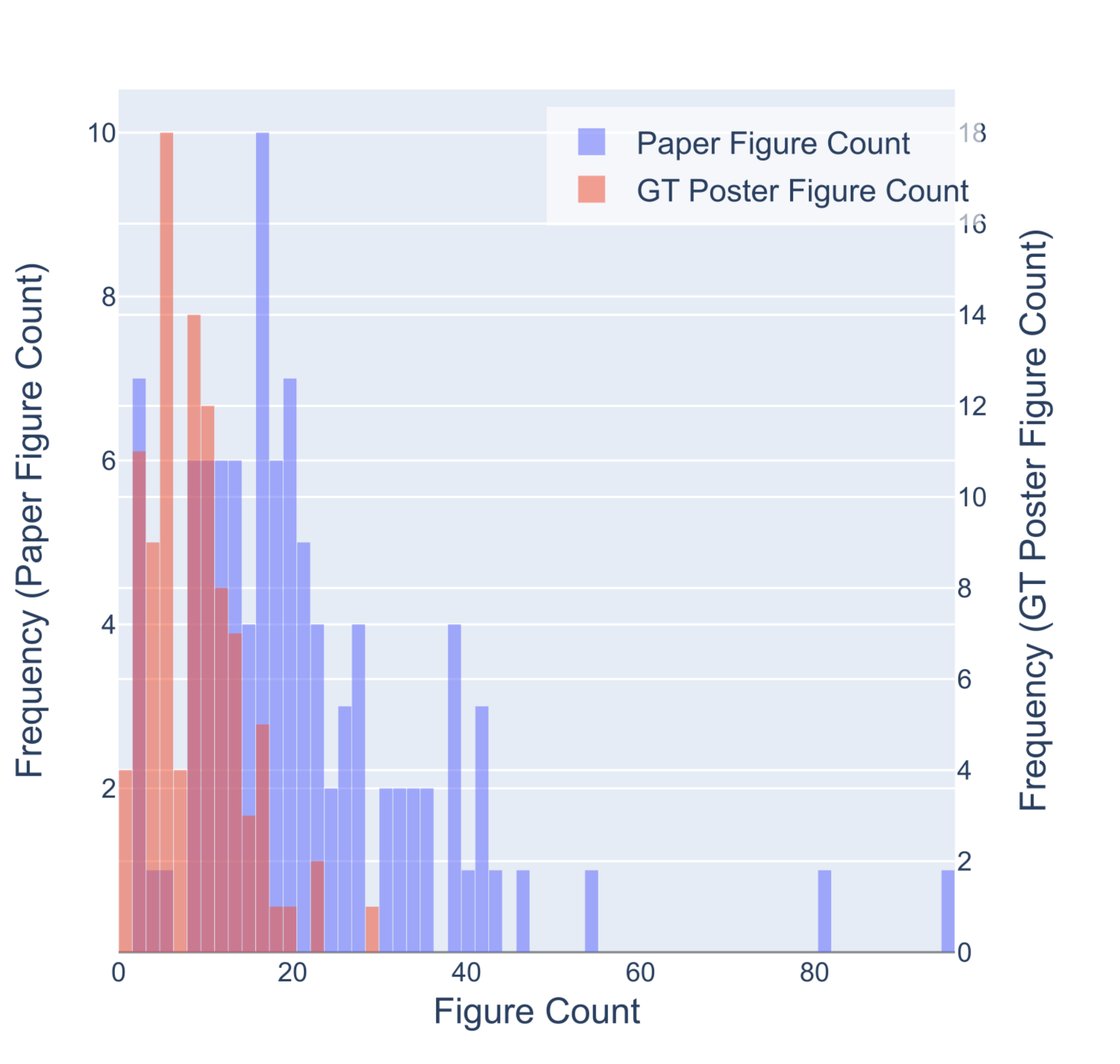}
    \vspace{-2em}
    \caption{$\#$ of figures}
    \label{fig:figure_count}
  \end{subfigure}
  \caption{\textbf{Data Statistics of \ourbench.}
  (a) Word cloud illustrating the diversity of research topics.
  (b) Textual Token statistics and Figure count statistics for input papers \textit{vs.} posters provided by authors.
  Overall, these statistics highlight that {Paper2Poster} is a multimodal context compression task, requiring effective abstraction of both textual and visual content.}\vspace{-1em} 
  \label{fig:data_stats_four-panel}
\end{figure}

\noindent \textbf{Data Statistics.}
Overall, \ourbench~comprises 100 paper-poster pairs spanning $280$ distinct topics across domains such as Computer Vision ($19\%$), Natural Language Processing ($17\%$), and Reinforcement Learning ($10\%$), ensuring comprehensive coverage across subfields.
As illustrated in Fig.~\ref{fig:data_stats_four-panel}~(a-b), the input papers contain an average of $12155.7$ words across $22.6$ pages, amounting to approximately $20370.3$ tokens, with an average of $22.59$ figures per paper. 
In Fig.~\ref{fig:data_stats_four-panel}~(c-d), the corresponding author-designed posters include an average of $774.1$ words ($1416.2$ tokens) and $8.7$ figures. This reflects a textual compression ratio of approximately $14.4\times$ and a figure reduction ratio of about $2.6\times$ from paper to poster.

\subsection{Evaluation Metrics}
\begin{figure}[htbp]
  \centering
  \includegraphics[width=\textwidth]{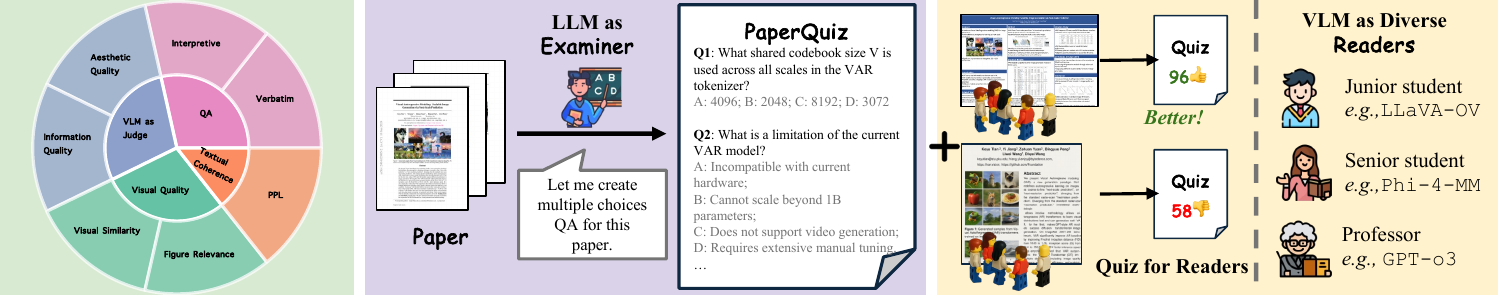}
  \caption{
  \textbf{Left}: Overview of the evaluation framework in \ourbench.
    \textbf{Middle}: We automatically generate multiple-choice questions from each paper using an LLM (o3), forming the our \textbf{\kqa}\ evaluation.
    \textbf{Right}: In \kqa, we simulate multiple reader by allowing VLMs—representing different expertise levels (\eg~student, professor)—to read each generated poster and answer the quiz. The poster that achieves the highest average score is considered the most effective in conveying the paper's content.}
    \vspace{-1em}
  \label{fig:eval}
\end{figure}

To systematically measure the quality of generated posters, we establish a comprehensive evaluation framework that covers four essential dimensions as shown in Fig.~\ref{fig:eval} (left): \textbf{(i) visual quality}, \textbf{(ii) textual coherence}, \textbf{(iii) quality assessment via VLM} (\ie VLM-as-judge), and notably our proposed \textbf{(iv) \kqa} which measures how effectively the poster conveys the paper’s core knowledge.

\noindent \textbf{(i) Visual Quality.}
\label{visquality}
The visual presentation of a poster directly impacts reader comprehension and engagement.
To evaluate visual quality from both global and local perspectives, we employ two metrics:
(1) We measure "\textit{Visual Similarity}" between the generated and the author-designed posters as ground-truth using CLIP image embeddings.
This metric captures high-level visual–textual correspondence to assess whether outputs are truly "poster-like" rather than article-like layouts, though it is not a direct measure of aesthetic quality.
This approach is favored over traditional distribution-based metrics (such as FID used in prior works~\cite{autopresent,pptagent}), as it assesses instance-level semantic consistency.
(2) We measure "\textit{Figure Relevance}" by computing the average CLIP similarity between figures and their corresponding text sections in the original paper. This metric ensures figures are contextually appropriate and effectively integrated, assigning zero relevance to posters lacking visual content.
For both metrics, we employ AltCLIP~\cite{altclip} due to its robustness in handling longer sequences alignment.
We complement CLIP with aspect-level VLM-as-Judge evaluation (see below) to capture fine-grained visual quality that CLIP may not fully capture.
Detailed definition of both metrics can be found in Appendix~\ref{app:vis_qual}.

\noindent \textbf{(ii) Textual Coherence.}
\label{txtcoherence}
Clear and fluent text is essential for poster readability and comprehension. We therefore quantify textual coherence by computing the standard "\emph{Perplexity}" (PPL) of the entire poster text under \texttt{Llama-2-7b-hf}. Lower PPL indicates more predictable, coherent language.
Importantly, PPL assesses \textit{fluency and local coherence} rather than semantic similarity to a reference, making it well-suited for our abstractive poster generation task where content should be reorganized and compressed rather than copied.
A detailed definition is provided in Appendix~\ref{app:text_coh}.

\noindent \textbf{(iii) Holistic  Assessment (VLM-as-Judge).}
\label{vlmjudge}
To evaluate overall poster effectiveness in fine-grained dimension, we prompt a VLM (\eg\ GPT‑4o) as an automated judge by outputting score ($1$–$5$). For each poster image, the model assigns $6$ criterion-level scores:
$3$ under “\emph{Aesthetic Score}”—\{Element Quality, Layout Balance, Engagement\}, and $3$ under “\emph{Information Score}”—\{Clarity, Content Completeness, Logical Flow\}. 
This direct, image-centric evaluation preserves fidelity to both visual design and content, while also capturing informativeness. It provides fine-grained feedback to guide future poster design. Full prompt templates and scoring protocols are detailed in Appendix~\ref{app:vlm_judge}.  

\noindent \textbf{(iv) \kqa.} 
Given the poster’s central role in communicating the content of its source paper—serving as a bridge between authors and readers—we design an evaluation protocol that simulates this communication scenario.  
As shown in Fig.~\ref{fig:eval} (middle), each paper PDF is first submitted to o3 as examiner to generate 100 multiple-choice questions per paper: 50 \emph{verbatim} questions (directly answerable from the text, spanning 13 content aspects) and 50 \emph{interpretive} questions (targeting high-level comprehension across 10 conceptual dimensions).  
Next, as illustrated in Fig.~\ref{fig:eval} (right), we present each poster image to six VLMs (both open- and closed-source), simulating a range of reader standards from casual to expert. These models then answer the quiz based solely on the poster content. By comparing their quiz scores across different poster variants, we identify which poster best conveys the original paper content.
Given that a poster is a visual medium rather than plain text like a note, we further adjust the raw Quiz scores \(\mathbf{s}_r \in [0, 100]\) by incorporating a length-based penalty, resulting in a penalized score \(\mathbf{s}_a \in [0, 200]\):
\[
\mathbf{s}_a = \mathbf{s}_r \left(1 + \tfrac{1}{\max(1,\,L/W)}\right),
\]
where \(L\) denotes the total text length of the poster, and \(W\) is the median text length of human-designed (ground-truth) posters.
This penalty function is designed with three goals: (i) discourage overly long posters (\(L \gg W\) yields \(\mathbf{s}_a \to \mathbf{s}_r\), losing the bonus), (ii) avoid harsh punishment (as \(L \to \infty\), \(\mathbf{s}_a\) remains at \(\mathbf{s}_r\), not approaching zero), and (iii) prevent rewarding extreme brevity (when \(L \leq W\), the bonus is capped at \(\mathbf{s}_a = 2\mathbf{s}_r\), so further shortening provides no additional gain). By anchoring the penalty to human-designed poster lengths, we ensure that posters are neither excessively verbose nor sacrifice informative content for brevity.
Further details on metric design, question curation, evaluation workflow, and scoring procedures can be found in Appendix~\ref{app:qa}.

\begin{figure}[!t]
  \centering
  \includegraphics[width=\textwidth]{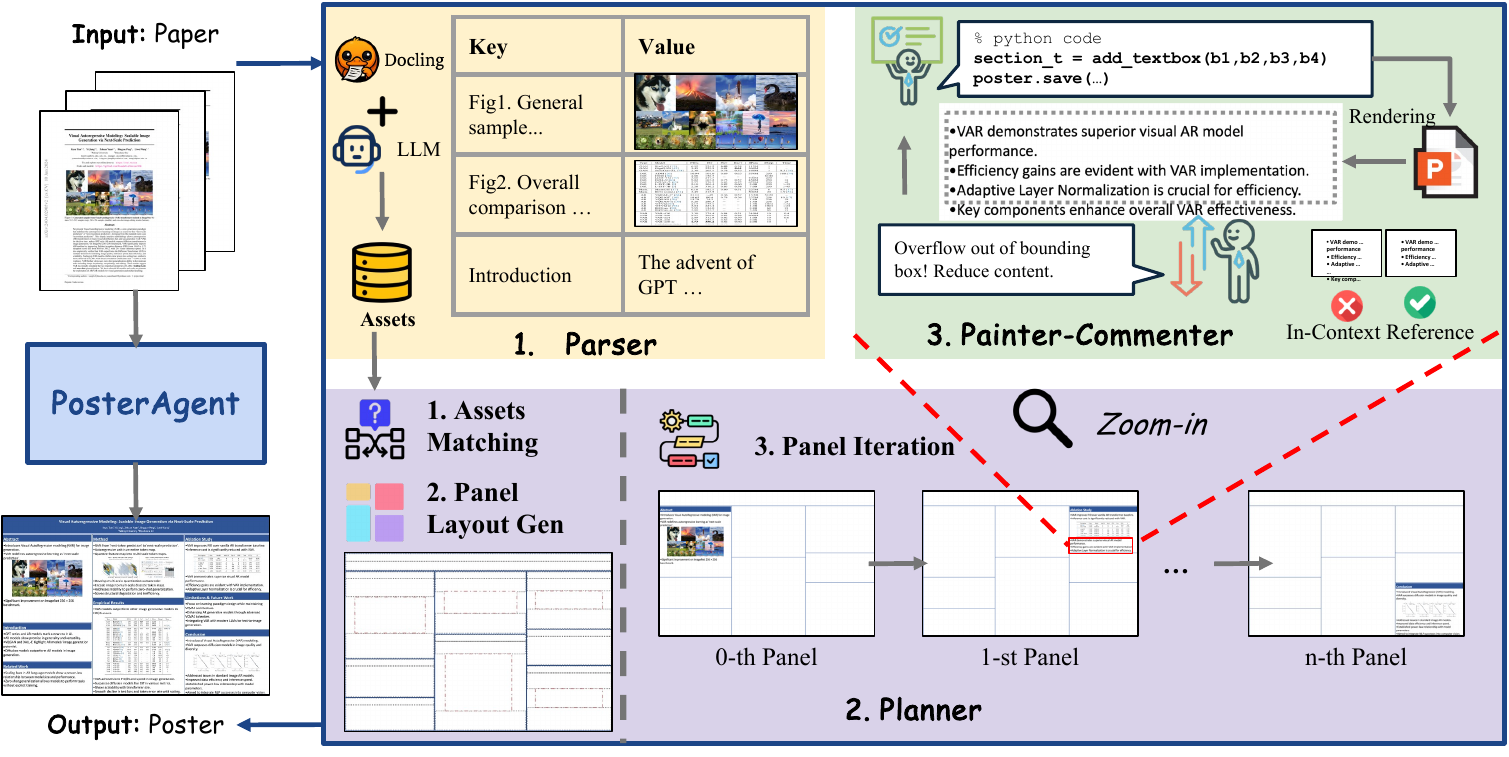}
\caption{
\textbf{Illustration of the PosterAgent pipeline.}
Given an input paper, PosterAgent generates a structured academic poster through three modules:
\textbf{1. Parser:} Extracts key textual and visual assets using a combination of tools and LLM-based summarization, resulting in a structured asset library.
\textbf{2. Planner:} Matches assets and arranges them into coherent layouts, iteratively generating panels with a \textit{zoom-in} operation.
\textbf{3. Painter–Commenter:} The Painter generates panel-level bullet-content along with executable code, and renders the visual output, while the Commenter—a VLM with in-context reference—provides feedback to ensure layout coherence and prevent content overflow.
}\vspace{-1.5em} 
\label{fig:posteragent}
\end{figure}

\vspace{-10pt}  
\section{\ouralg}
\label{sec:posteragent}
\vspace{-8pt}  
\noindent\textbf{Overview.}
Identifying the challenges posed by the \ourbench, we formulate it as a problem of multimodal context compression, and introduce \ouralg, a multi-agent pipeline that adopts a “\textit{Top-down}” design philosophy: it first globally restructures the entire document into concise, coherent sections, followed by local refinements for fine-grained, panel-level control.
As shown in Fig.~\ref{fig:posteragent}. The pipeline consists of three key components:  
\textbf{1. Parser:} Extracts key textual and visual content by tools and LLM-based summarization to build an asset library.  
\textbf{2. Planner:} Aligns assets and arranges them into coherent layouts, generating panels iteratively with a \textit{zoom-in} mechanism.  
\textbf{3. Painter–Commenter:} The Painter produces panel-level bullet points and executable code for rendering, while a VLM as Commenter—ensures layout coherence and avoids overflow.

\subsection{Parser: \textit{global organization}}\vspace{-0.5em}
Given a paper, the first step is to \textit{globally} organize the information into a structured format to support subsequent processing. This is handled by the \textbf{Parser}, which performs a coarse-grained compression by ingesting the raw PDF and producing an \textbf{asset library} across two modalities:
\textbf{(1) Text assets} that capture the document hierarchy like human first glance focus on section heading—each key is a section heading and the associated value a paragraph‑level synopsis; 
\textbf{(2) Visual assets} built in parallel, where figure or table captions serve as keys and the extracted image files are stored as values. 
We leverage \textsc{Marker}\cite{marker} and \textsc{Docling}\cite{docling} to convert each page into Markdown, which is then processed by an LLM to generate a structured, JSON-like outline. This transformation compresses the raw text into a compact asset library that preserves essential semantics while significantly reducing size, enabling more efficient downstream iteration and layout generation.
\vspace{-0.7em}

\subsection{Planner: \textit{local organization}}\vspace{-0.5em}
With the visual and text assets collected by the Parser, the next step is to select the relevant content and begin constructing the poster. Rather than generating the entire poster in one shot, we emphasize the importance of layout configuration and adopt an iterative, \emph{section-by-section} completion process—mirroring how humans typically start with a template and sequentially fill in each section.

\noindent \textbf{Asset matching.}
This step aims to associate visual assets with corresponding textual content—for example, matching a teaser image to the introduction paragraph. We employ an LLM to semantically align each visual asset with its most relevant section from the asset library, resulting in a set of $(\text{section}, \text{figure})$ pairs.

\noindent \textbf{Layout generation.}
An essential step is determining the panel-level layout, which requires precise absolute coordinates while accounting for the relative informativeness of each section. We found that directly predicting numerical coordinates using an LLM was unstable. Therefore, we adopt the binary-tree layout strategy~\cite{genposter}, which reliably translates hierarchical constraints into panel bounding boxes by estimating content length (\eg, word number, figure size), maintaining reading order, and preserving aspect ratio—ensuring each poster section corresponds to a well-defined panel.

\noindent \textbf{Panel iteration.}
Once the paper layout is configured, the next stage is to populate each panel with content. To ensure precise control, the Planner iterates over each section’s synopsis and condenses it into concise, hierarchically structured bullet points—creating a compact format well-suited for poster panels.
Inspired by how humans design posters—initially filling in content and iteratively refining it based on visual feedback—we introduce the {\textit{Painter-Commenter loop}} (Sec.~\ref{sec:coder_commenter}), which mimics this process while maintaining visual clarity and appeal.
After all panels undergo this process, the finalized poster is produced.
\vspace{-0.7em}

\subsection{Painter–Commenter: \textit{local refinement}}\label{sec:coder_commenter}\vspace{-0.5em}
For each panel, the \textbf{Painter} converts its asset pair \ie~$(\text{section}, \text{figure})$ into executable code instructions and invokes the runtime environment to render a draft panel image.
Particularly, the Painter comprises two modules: (i) an LLM that ingests the section synopsis and distills it into a concise set of bullet points, and (ii) a deterministic code generator that leverages the \texttt{python-pptx} library together with predefined helper functions to generate presentation code, which is subsequently executed and rendered into an image of the current panel.

However, in practice, a single pass rarely produces a flawless panel. To address this, we pair the Painter with a \textbf{Commenter}—a VLM that evaluates the quality of the rendered panel image.
While VLMs are promising, they often hallucinate in visual design tasks, leading to unreliable judgments. To mitigate this, we employ a \textit{Zoom-in} strategy that focuses attention on the panel region.
Additionally, we enhance the Commenter with an \emph{in-context reference} prompt containing two examples: one with severe overflow and one with an ideal layout. Guided by these references, the Commenter provides targeted visual feedback—such as “overflow,” “too blank,” or “good to go”—which informs the Painter’s next revision.
This loop continues until the Commenter signals success or a maximum number of iterations is reached, ensuring each panel is accurate, readable, and visually well-balanced.

\vspace{-0.7em}
\section{Experiments} \label{sec:exp}
\vspace{-0.5em}
\subsection{Baselines and Settings}
\label{sec:baseline_setting}  \vspace{-0.5em}
We evaluate four categories of baselines:
\textbf{(i) Oracle methods}, which serve as upper bounds—"\textit{Paper}" (the original PDF with maximum informativeness) for content fidelity, and "\textit{GT Poster}" (the author-designed poster from \ourbench) as the best possible presentation in terms of human understanding and layout quality;
\textbf{(ii) End-to-end methods}, where GPT-4o directly generates posters either through text-based rendering—"\texttt{4o-HTML}" (Markdown-to-HTML)—or image generation—"\texttt{4o-Image}" (poster graphics produced via GPT-4o's web interface);
\textbf{(iii) Multi-agent workflows}, which decompose the task using specialized toolkits—"\texttt{OWL}"\cite{owl2025}, a general-purpose PDF-to-HTML converter, and "\texttt{PPTAgent}"\cite{pptagent}, a Python-pptx-based slide generator, where candidate posters are selected via manual inspection;
\textbf{(iv) \ouralg}, our proposed approach—\texttt{\ouralg-4o} uses GPT-4o for both internal LLM and VLM commenter, while \texttt{\ouralg-Qwen} is a purely \textit{open-source} solution, employs Qwen-2.5-7B for text generation and Qwen-2.5-VL-7B for commenter. Additional backbones are evaluated to study the generalizability of our method, which is detailed in Appendix~\ref{append:additional_backbones}.

\subsection{Main Results} \label{subsec:main_res}
\noindent\textbf{Visual Quality \& Text Coherence.}
In the left part of Tab.~\ref{tab:judge_table}, we evaluate visual quality and textual coherence. Interestingly, while \texttt{4o-Image} achieves the highest visual similarity, it also records the worst perplexity, suggesting that \textit{although the generated posters may appear visually appealing at first glance, they often contain noisy or incoherent text}. As expected, the original paper performs best in terms of textual coherence. Notably, the author-designed poster (GT) still shows relatively high PPL, indicating that \textit{authors often prioritize visual appeal and reader engagement by conveying information through visual rather than textual means}.
Our \texttt{\ouralg}\ achieves the highest figure relevance compared to \texttt{PPTAgent},
\setlength{\intextsep}{1pt}      
\setlength{\columnsep}{4pt}      
\begin{wrapfigure}{l}{0.4\columnwidth}
  \flushleft                   
  \includegraphics[
    width          = \linewidth
  ]{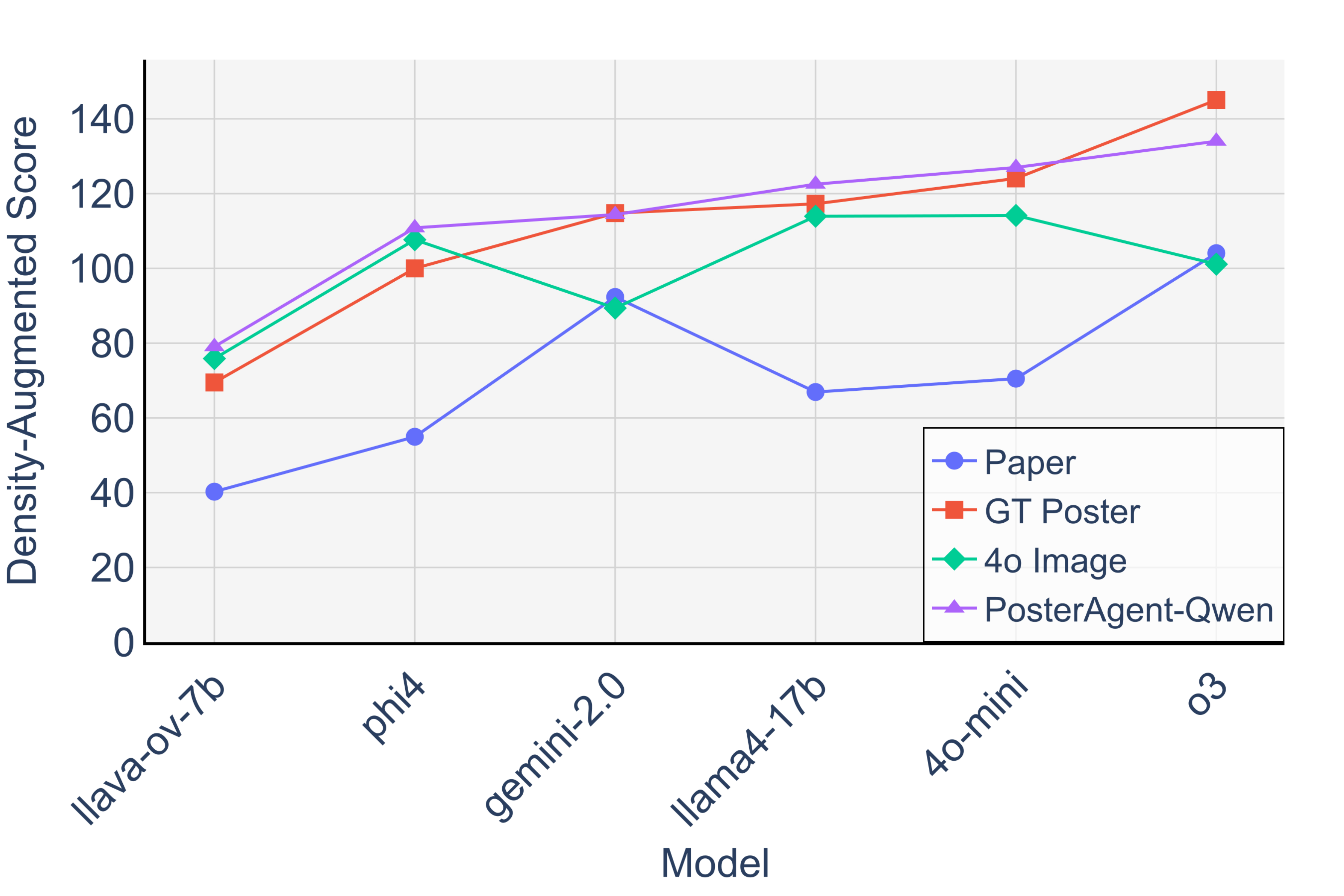}

  \caption{\textbf{\kqa's Avg. scores across different Reader VLMs} (x-axis) for each poster type (legend lines). Refer to Append. Tab. \ref{tab:abbrev_fullname} for full model names.}
  \label{fig:qa_across_models}
  
\end{wrapfigure}
primarily due to our visual-semantic-aware asset library construction and asset matching. It also ranks second in visual similarity, closely following the human-designed poster. 
Above results highlight that each metric captures only a specific aspect of quality and has its limitations. Therefore, we turn to the VLM-as-Judge and {PaperQuiz} next.

\begin{table*}[t]
  \centering
  \setlength{\tabcolsep}{1.8pt} 
  \resizebox{\textwidth}{!}
  {
  \begin{tabular}{l *{3}{c} *{9}{c}}
    \toprule
    \multirow{5}{*}{\textbf{Model}}
      & \multicolumn{3}{c}{\textbf{Vis. quality \& Txt. coherence}}
      & \multicolumn{9}{c}{\textbf{VLM-as-Judge}} \\[1ex]
    \cmidrule(lr){2-4}\cmidrule(lr){5-13}
      & \multirow{3}{*}{Vis. Sim.~$\uparrow$}
      & \multirow{3}{*}{PPL~$\downarrow$}
      & \multirow{3}{*}{Fig. Rel.~$\uparrow$}
      & \multicolumn{4}{c}{\textbf{Aesthetic score}~$\uparrow$}
      & \multicolumn{4}{c}{\textbf{Information score}~$\uparrow$}
      & \multirow{3}{*}{\textbf{Overall}~$\uparrow$} \\[1ex]
    \cmidrule(lr){5-8}\cmidrule(lr){9-12}
      &  &  & 
      & Element & Layout & Engage. & \textbf{Avg.}
      & Clarity & Content & Logic & \textbf{Avg.}
      & \\[0.5ex]
    \midrule
    \textit{{Oracle methods}} \\
    \rowcolor{gt_models!50}
    Paper
      & 0.53 & \textbf{4.60} & \underline{0.22}
      & \underline{4.05} & \underline{3.89} & 2.80 & \cellcolor{gt_models}\textbf{3.58}
      & 4.00 & \textbf{4.68} & \textbf{3.98} & \cellcolor{gt_models}\textbf{4.22}
      & \cellcolor{gt_models}{\textbf{3.90}} \\[1ex]

    \rowcolor{gt_models!50}
    GT Poster
      & \textcolor{gray}{1.00} & 11.26 & 0.21
      & \textbf{4.07} & \textbf{3.90} & 2.70 & \cellcolor{gt_models}\underline{3.56}
      & \textbf{4.09} & \underline{3.96} & \underline{3.89} & \cellcolor{gt_models}\underline{3.98}
      & \cellcolor{gt_models}{\underline{3.77}} \\[1ex]

    \midrule
    \textit{{End-to-end methods}} \\

    \rowcolor{direct_gen!50}
    \texttt{4o-HTML}
      & 0.52 & 9.86 & – 
      & 3.53 & 3.82 & 2.72 & \cellcolor{direct_gen}3.36
      & 3.94 & 3.64 & 3.47 & \cellcolor{direct_gen}3.68
      & \cellcolor{direct_gen}{3.52} \\[1ex]

    \rowcolor{direct_gen!50}
    \texttt{4o-Image}
      & \textbf{0.76} & 77.13 & 0.21
      & 2.93 & 3.02 & 2.75 & \cellcolor{direct_gen}2.90
      & 1.05 & 2.04 & 2.22 & \cellcolor{direct_gen}1.77
      & \cellcolor{direct_gen}{2.33} \\[1ex]

    \midrule
    \textit{{Multi-Agent methods}} \\

    \rowcolor{multi_agent!50}
    \texttt{OWL-4o}
      & 0.54 & 11.46 & –
      & 2.76 & 3.62 & 2.56 & \cellcolor{multi_agent}2.98
      & 3.92 & 2.89 & 3.36 & \cellcolor{multi_agent}3.39
      & \cellcolor{multi_agent}{3.19} \\[1ex]

    \rowcolor{multi_agent!50}
    \texttt{PPTAgent-4o}
      & 0.50 & \underline{6.20} & 0.16
      & 2.49 & 3.05 & 2.45 & \cellcolor{multi_agent}2.66
      & 2.05 & 1.26 & 1.38 & \cellcolor{multi_agent}1.56
      & \cellcolor{multi_agent}{2.11} \\[1ex]

    \midrule
    \textit{{\ouralg\ variants}} \\[0.3ex]

    \rowcolor{poster_agent!50}
    \texttt{\ouralg-4o}
      & \underline{0.75} & 8.31 & \textbf{0.24}
      & 3.95 & 3.86 & \textbf{2.93} & \cellcolor{poster_agent}\textbf{3.58}
      & \underline{4.03} & \underline{3.96} & 3.60 & \cellcolor{poster_agent}3.86
      & \cellcolor{poster_agent}{3.72} \\[1ex]

    \rowcolor{poster_agent!50}
    \texttt{\ouralg-Qwen}
      & \underline{0.75} & 8.81 & \textbf{0.24}
      & 3.93 & 3.67 & \underline{2.89} & \cellcolor{poster_agent}3.50
      & 3.95 & 3.85 & \underline{3.68} & \cellcolor{poster_agent}3.83
      & \cellcolor{poster_agent}{3.66} \\

    \bottomrule
  \end{tabular}%
  }
\caption{\textbf{Detailed evaluation of \ourbench\ across four categories of baselines,} including \textit{Visual Quality \& Text Coherence} and \textit{VLM-as-Judge} for fine-grained assessments. 
Oracle methods together (Paper or author-designed poster) serve as upper bounds in theory and strong baselines empirically. 
}\vspace{-1em}  
  \label{tab:judge_table}
\end{table*}

\noindent\textbf{VLM as Judge Metrics.}
In the right part of Tab.~\ref{tab:judge_table}, we conduct a comprehensive evaluation using a suite of metrics.
We find that both the Paper and GT Poster achieve the highest aesthetic and information scores. In contrast, \texttt{4o-Image} performs poorly in terms of information, aligning with findings from previous PPL studies.
Overall, \texttt{\ouralg-4o} achieves an average score of $3.72$, reaching a level comparable to that of human-designed posters. Variants of \ouralg\ that use GPT-4o as the visual commenter outperform those using \texttt{Qwen2.5-VL-7B}, highlighting the superior visual perception capabilities of \texttt{4o}, particularly in panel refinement tasks such as preventing text overflow.

\texttt{PPTAgent} frequently fails to replace placeholder content or fill templates properly, leading to meaningless text or large blank areas, and thus receives low scores in both aesthetics and informativeness. 
Despite not generating images, \texttt{4o-HTML} yields the highest aesthetic score among baselines, owing to its clean and structured layout. 
Overall, we found that \textit{the primary bottleneck in existing poster generation lies in Engagement}, where all variants score below 3. In contrast, most variants achieve good Information scores, likely due to the robust long-context handling capabilities of GPT-4o. All \ouralg\ variants—even those using Qwen2.5-7B—surpass baselines in information quality, demonstrating the effectiveness of our content planning and generation framework in mitigating limitations of less capable LLMs. Although \texttt{PPTAgent} is also powered by GPT-4o, its rigid template-filling mechanism often fails to properly populate content, leading to poor performance.

\noindent\textbf{\kqa.}
As shown in Tab.~\ref{tab:qa_table}, we draw several key observations:
\textit{\textbf{(i)}} \textit{Verbatim} questions are generally more challenging than those assessing broader understanding and interpretation.
\textit{\textbf{(ii)}} Without textual brevity penalties, \texttt{Paper} achieves the highest overall score. When the penalty is applied, the \texttt{GT Poster} performs best. This highlights both the comprehensiveness of the full paper and the value of concise, well-designed posters. It also reinforces how the \kqa{} setup reflects poster generation as a process of \textit{effective context compression}, where careful condensation rather than sheer content volume is rewarded.
\textit{\textbf{(iii)}} GPT-4o supplies strong base ability. Its \texttt{4o-HTML} variant outperforms \texttt{OWL-4o}, and even its purely visual \texttt{4o-Image} generation surpasses \texttt{PPTAgent-4o}. Our proposed \texttt{\ouralg{}} variants consistently achieve the best scores.
\textit{\textbf{(iv)}} Across all methods, performance on open-source reader models is consistently lower than on closed-source ones. This suggests that stronger perceptual ability correlates with better poster comprehension.
\textit{\textbf{(v)}} Notably, both \texttt{4o-HTML} and \texttt{OWL-4o}, despite leveraging GPT-4o and generating lengthy, figure-free, blog-style outputs, are outperformed in \textit{raw accuracy} by our \texttt{\ouralg-Qwen} variant, even though they are exempt from brevity penalties. 
This result further affirms that \kqa{} evaluates more than content volume; presentation quality matters. 
Our \texttt{\ouralg-Qwen} surpasses more resource-intensive baselines despite relying on the relatively weaker \texttt{Qwen-2.5-VL-7B}, due to two key design choices: 
(a) a structured, multi-step compression process that enables even weaker LMs to distill information with minimal loss; and (b) a layout that presents information clearly and with a logical reading order, making it easy for VLM-based readers to locate and interpret key points, similar to how clear visual structure supports efficient comprehension for human poster readers.

\begin{table*}[t]
  \centering
  \setlength{\tabcolsep}{1.8pt}
  \resizebox{\textwidth}{!}{%
  \begin{tabular}{l *{7}{c} *{3}{c}}
    \toprule
      & \multicolumn{7}{c}{\textbf{Raw Accuracy}}
      & \multicolumn{3}{c}{\textbf{Density-Augmented Score}} \\[1ex]
    \cmidrule(lr){2-8}\cmidrule(lr){9-11}
    \textbf{Model}
      & \multicolumn{3}{c}{\textbf{Verbatim}~$\uparrow$}
      & \multicolumn{3}{c}{\textbf{Interpretive}~$\uparrow$}
      & \multirow{2}{*}{\textbf{Overall}~$\uparrow$}
      & \multirow{2}{*}{\textbf{V-Avg}~$\uparrow$} & \multirow{2}{*}{\textbf{I-Avg}~$\uparrow$} & \multirow{2}{*}{\textbf{Overall}~$\uparrow$}
      \\[0.5ex]
      & open-source & closed-source & \textbf{V-Avg}
      & open-source & closed-source & \textbf{I-Avg}
      &
      & & &
      \\[0.5ex]
    \midrule
    \textit{Oracle methods} \\
    \rowcolor{gt_models!50}
    Paper
      & 51.45 & \textbf{82.95}       & \cellcolor{gt_models}\textbf{67.20}
      & 48.48 & \textbf{81.61}       & \cellcolor{gt_models}\underline{65.05}
      & \cellcolor{gt_models}\textbf{66.12}
      & \cellcolor{gt_models}72.69 & \cellcolor{gt_models}70.34 & \cellcolor{gt_models}71.52
      \\[1ex]
    \rowcolor{gt_models!50}
    GT Poster
      & 51.75 & \underline{58.10}    & \cellcolor{gt_models}\underline{54.93}
      & 49.19 & 77.55                & \cellcolor{gt_models}63.37
      & \cellcolor{gt_models}\underline{59.15}
      & \cellcolor{gt_models}\textbf{103.56} & \cellcolor{gt_models}120.00 & \cellcolor{gt_models}111.78
      \\[1ex]
    \midrule
    \textit{End-to-end methods} \\
    \rowcolor{direct_gen!50}
    \texttt{4o-HTML}
      & \underline{52.45} & 48.00 & \cellcolor{direct_gen}50.23
      & 50.78 & 75.14 & \cellcolor{direct_gen}62.96
      & \cellcolor{direct_gen}56.59
      & \cellcolor{direct_gen}95.72  & \cellcolor{direct_gen}120.55 & \cellcolor{direct_gen}108.13
      \\[1ex]
    \rowcolor{direct_gen!50}
    \texttt{4o-Image}
      & 48.97 & 30.89 & \cellcolor{direct_gen}39.93
      & 50.19 & 70.67 & \cellcolor{direct_gen}60.43
      & \cellcolor{direct_gen}50.18
      & \cellcolor{direct_gen}79.86  & \cellcolor{direct_gen}120.86 & \cellcolor{direct_gen}100.36
      \\[1ex]
    \midrule
    \textit{Multi-Agent methods} \\
    \rowcolor{multi_agent!50}
    \texttt{OWL-4o}
      & 47.87 & 31.96 & \cellcolor{multi_agent}39.92
      & 49.94 & 74.38 & \cellcolor{multi_agent}62.16
      & \cellcolor{multi_agent}51.04
      & \cellcolor{multi_agent}78.69  & \cellcolor{multi_agent}122.91 & \cellcolor{multi_agent}100.80
      \\[1ex]
    \rowcolor{multi_agent!50}
    \texttt{PPTAgent-4o}
      & 39.63 & 11.99 & \cellcolor{multi_agent}25.81
      & 36.22 & 37.15 & \cellcolor{multi_agent}36.68
      & \cellcolor{multi_agent}31.25
      & \cellcolor{multi_agent}51.62  & \cellcolor{multi_agent}73.37  & \cellcolor{multi_agent}62.49
      \\[1ex]
    \midrule
    \textit{\ouralg\ variants} \\[0.3ex]
    \rowcolor{poster_agent!50}
    \texttt{\ouralg-4o}
      & \textbf{52.95} & 49.17                & \cellcolor{poster_agent}51.06
      & \underline{52.29} & 78.42             & \cellcolor{poster_agent}\textbf{65.35}
      & \cellcolor{poster_agent}58.21
      & \cellcolor{poster_agent}\underline{101.87} & \cellcolor{poster_agent}\textbf{130.39} & \cellcolor{poster_agent}\textbf{116.13}
      \\[1ex]
    \rowcolor{poster_agent!50}
    \texttt{\ouralg-Qwen}
      & 51.81 & 48.79                & \cellcolor{poster_agent}50.30
      & \textbf{52.57} & 76.66             & \cellcolor{poster_agent}64.62
      & \cellcolor{poster_agent}57.46
      & \cellcolor{poster_agent}100.35 & \cellcolor{poster_agent}\underline{128.94} & \cellcolor{poster_agent}\underline{114.65}
      \\
    \bottomrule
  \end{tabular}%
  }
  \caption{\textbf{\kqa\ Evaluation on \ourbench{} based on 6 different Readers}, including open-source and closed-source VLMs. 
  Both \textit{Raw Accuracy} and \textit{Density-Augmented Score} are included for \textit{Verbatim} and \textit{Interpretive} settings. 
  Oracle methods together (Paper or author-designed poster) serve as upper bounds empirically. 
  }\vspace{-1em}
  \label{tab:qa_table}
\end{table*}

\setlength{\intextsep}{1pt}      
\setlength{\columnsep}{4pt}      
\begin{wrapfigure}{r}{0.35\columnwidth}
  \vspace{-12pt}           
  \flushright
  \includegraphics[ width=\linewidth]{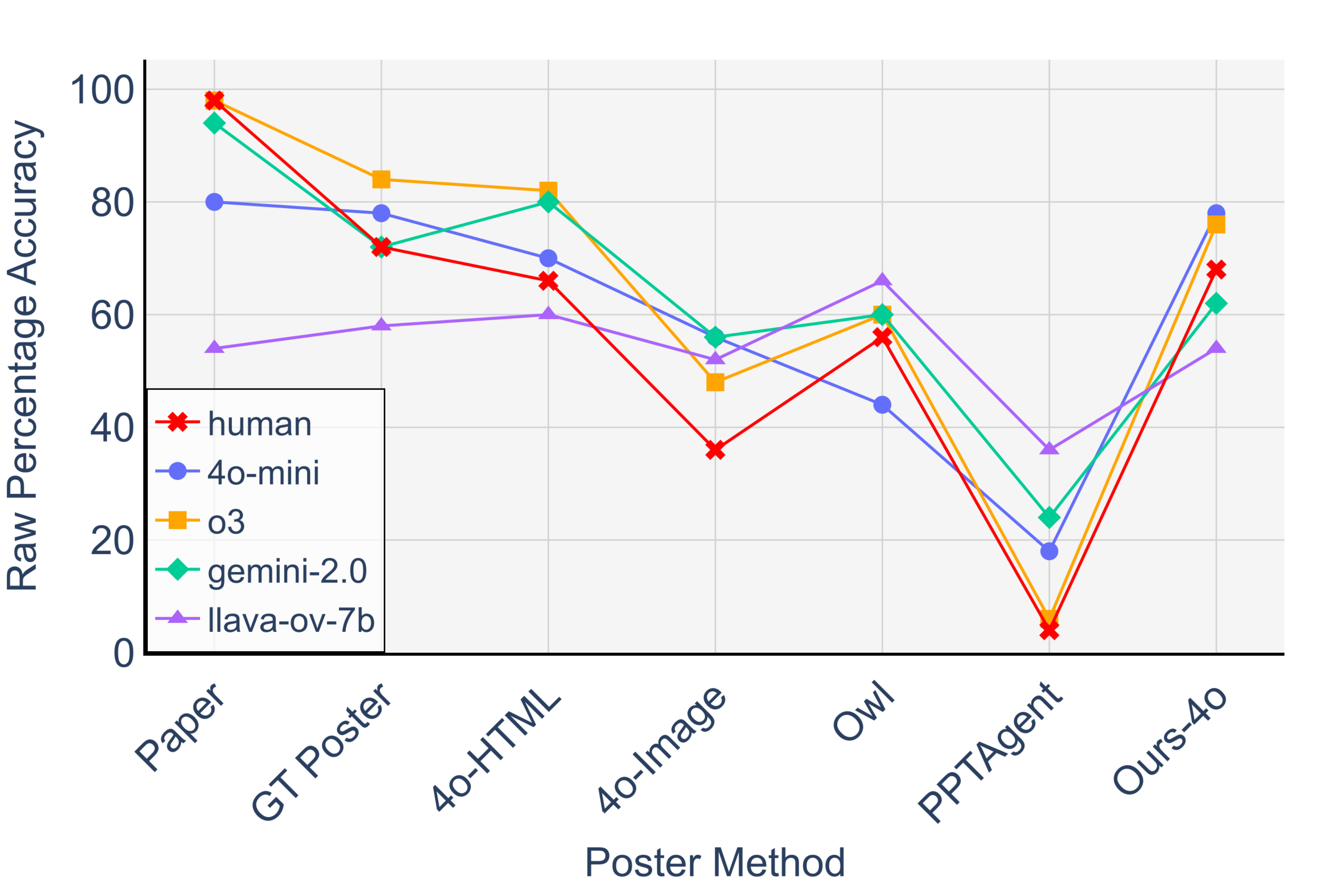}
  \caption{\textbf{\kqa's Avg scores across different types of posters} (x-axis) for readers (colored lines) on \textit{human evaluation} subset.}
  \label{fig:human_vs_model_b}
  \vspace{-9pt}           
\end{wrapfigure}
\noindent \textbf{PaperQuiz readers comparison.} 
In Fig.~\ref{fig:qa_across_models}, we compare the \kqa\ scores of different readers on four baseline posters. On GT and \ouralg's posters, we observe that \textit{as model reasoning capabilities improve, their ability to interpret structured content also increases}, leading to higher QA accuracy. In contrast, this trend is not evident for \texttt{4o-Image} and \texttt{Paper}, suggesting that \textit{more capable models benefit more from poster layouts and condensed information than from information-dense papers}, thereby improving their comprehension and response quality.

\noindent \textbf{Human evaluation.} To assess our method with human judgment, we recruited a PhD student to complete the \kqa\ on $5$ randomly selected papers from the \ourbench\ dataset, covering $4$ baselines, $2$ ground-truth variants, and $2$ \texttt{\ouralg} variants, following the setup in Section~\ref{sec:baseline_setting}. Details of the human evaluation protocol are provided in Appendix~\ref{append:human_eval}. Figure~\ref{fig:human_vs_model_b} demonstrates the average \kqa\ scores across different types of posters (x-axis) for each reader (colored lines). \kqa\ scores across different posters exhibit \textbf{good consistency across both human and VLMs evaluations}. This alignment supports the use of reader models as effective proxies to simulate human judgment.

\setlength{\intextsep}{1pt}      
\setlength{\columnsep}{4pt}      
\begin{wrapfigure}{l}{0.4\columnwidth}
  \vspace{-12pt}           
  \flushright
  \includegraphics[ width=\linewidth]{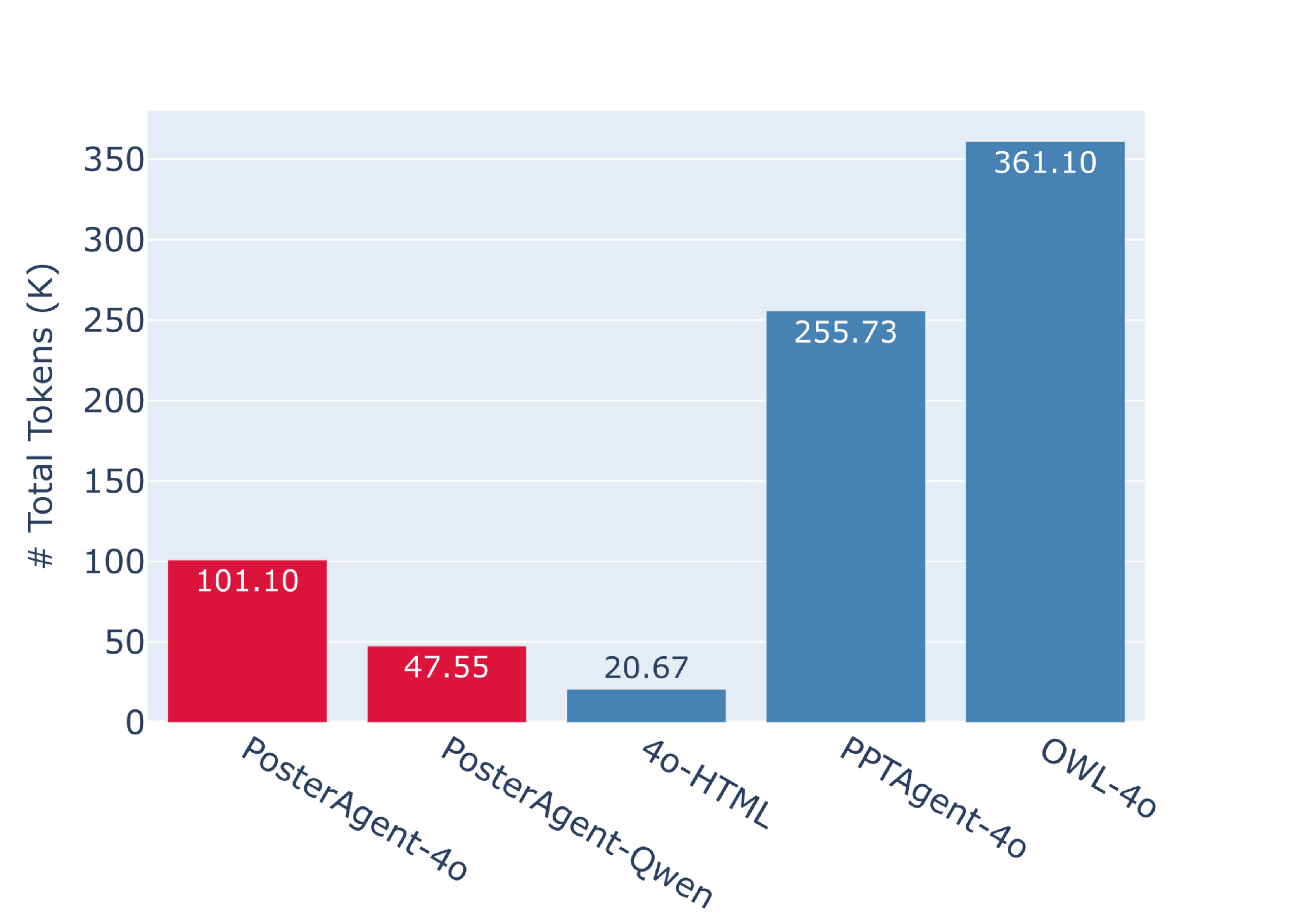}
  \caption{Average token consumptions for different methods. Details are provided in Appendix \ref{app:efficiency_analysis}.}
  \label{fig:token_consumption}         
\end{wrapfigure}

\vspace{-8pt}
\subsection{Qualitative Analysis}
\begin{figure}[!t]
  \centering
  \includegraphics[width=\textwidth]{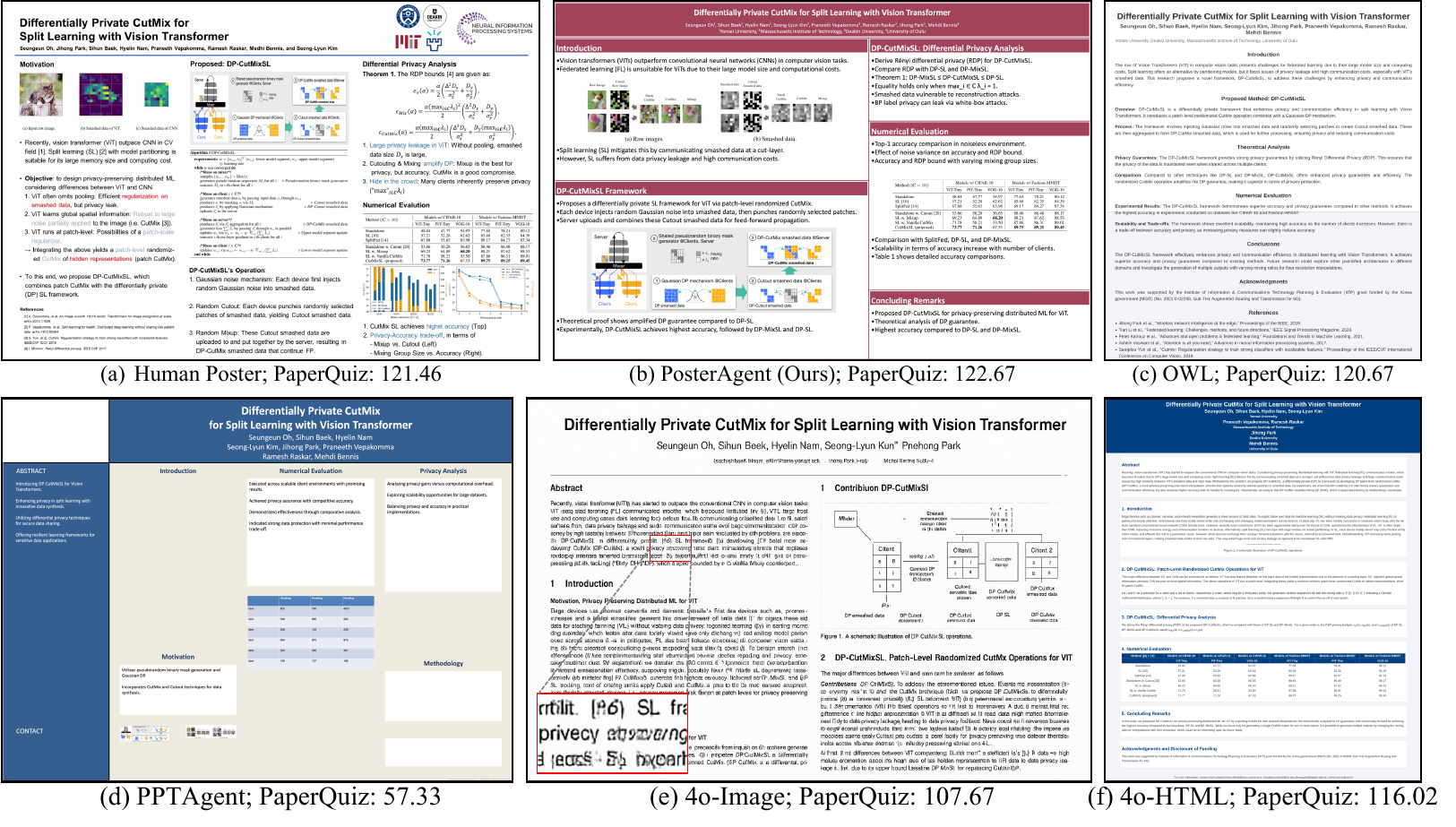}
  \caption{\textbf{Illustration of poster variants for the \href{https://arxiv.org/pdf/2210.15986}{paper} generated by different methods}, including (a) Author designed, (b) Our PosterAgent, multi-agent methods (c) OWL~\cite{owl2025} and (d) PPTAgent~\cite{pptagent}, pixel generative method (e) \texttt{4o-Image} and website generative method (f) \texttt{4o-HTML}.
  We provide the \kqa's augmented score for each method.
  }\vspace{-1.5em}
  \label{fig:example}
\end{figure}

In Figure~\ref{fig:example}, we present a quantitative comparison across different poster baselines for a paper~\cite{diffprivate}. 
GPT-4o's pixel-based generation produces visually acceptable layouts at first glance, but closer inspection (\textcolor{red}{\textit{zoom-in region}}) reveals impaired text rendering, leading to poor readability of fine-grained details. \texttt{4o-HTML} and \texttt{OWL} generate blog-like, text-dense posters that suffer from low visual readability. \texttt{PPTAgent} struggles with layout control, often resulting in missing panels. In contrast, our \ouralg\ generates structurally coherent and readable posters, achieving the highest scores while using significantly fewer words than (c) and (f). However, there is still room for improvements compared to human-designed versions.
\newline
\vspace{-1em}

\subsection{Efficiency Analysis}
Figure~\ref{fig:token_consumption} presents the average token cost per poster across different methods.
Our \texttt{\ouralg} achieves great token efficiency, using only $101.1K$ (4o-based) and $47.6K$ (Qwen-based) tokens—reducing cost by $60\%$–$87\%$ compared to \texttt{OWL-4o}~\cite{owl2025}. This translates to just $\$0.55$ for 4o and $\$0.0045$ for Qwen per poster, highlighting its effectiveness.
Additionally, through parallelization of panel generation, we further reduced runtime by 40.7\%, making \texttt{\ouralg-4o-Parallel} even more competitive in speed (see Append.~\ref{app:efficiency_analysis} for token details and Append.~\ref{app:runtime_analysis} for runtime breakdown).
\section{Conclusions}
We present a new benchmark, \ourbench{}, for poster generation from academic papers, and we highlight the challenges and limitations of current generative models or agents in handling long-context, layout-sensitive tasks. 
Our proposed solution, the \ouralg\ framework, leverages structured parsing, hierarchical planning, and visual feedback to enhance generation quality significantly. \ouralg\ not only narrows the performance gap with human-designed posters but also establishes a new efficiency standard, offering a practical and scalable approach to scientific communication.

\section{Acknowledgements.}
This work was supported by the UKRI Turing AI Fellowship (EP/W002981/1); and by NSERC through a Discovery Grant, an Alliance Grant, and the Canada CIFAR AI Chairs program. Resources were provided, in part, by the Province of Ontario, the Government of Canada through CIFAR, and companies sponsoring the Vector Institute.

\bibliographystyle{plain}
\bibliography{ref}
\clearpage
\appendix
\begin{center}
	\LARGE \bf {Appendix}
\end{center}

\etocdepthtag.toc{mtappendix}
\etocsettagdepth{mtchapter}{none}
\etocsettagdepth{mtappendix}{subsubsection}
\tableofcontents
\newpage

\section{Limitations and Future Work} \label{app:limit}

We spot a limitation in the current design: the sequential execution of panel refinements constitutes the primary efficiency bottleneck. Each panel’s generate–revise cycle is structurally independent and could be parallelized, yet our implementation processes them serially to preserve modularity and output quality. As a result, end‑to‑end poster creation takes approximately 4.5 minutes per document—acceptable for isolated use but restrictive for large‑scale or interactive workflows. Introducing panel‑level parallelism is a clear avenue for future work, with the potential to dramatically reduce runtime and improve scalability in batch generation and real‑time editing contexts.

\textbf{Future works.} 
(i) a well-considered poster should integrate external knowledge beyond paper such as community feedback—such as OpenReview comments and social media reactions—and leverage external assets like institutional icons and conference logos; 
and (ii) an improved workflow would involve human–AI collaboration, where the agent produces an initial draft, solicits user feedback, and iteratively refines its output to meet requirements. We leave these explorations in future.

\section{Example Visualization}

We present representative examples from our \ourbench{} dataset, which comprises 100 pairs of full-length research papers and their corresponding author‐designed posters. For each selected paper, we show (a) the original poster created by the authors—designed to convey the paper’s abstract, methodology, results, and key visuals in a single coherent layout—and (b) the poster automatically generated by our \ouralg{} framework, demonstrating its ability to extract, summarize, and arrange multimodal content into a visually balanced single‐page design. These examples span a range of subfields (reinforcement learning, anomaly detection, neuroscience) and illustrate how \ouralg{} handles diverse layouts, content compression ratios, and figure‐to‐text integration.

\begin{figure*}[!h]
  \centering
  \begin{tabular}{cc}
    \begin{subfigure}{0.48\linewidth}
      \centering
      \includegraphics[width=\linewidth]{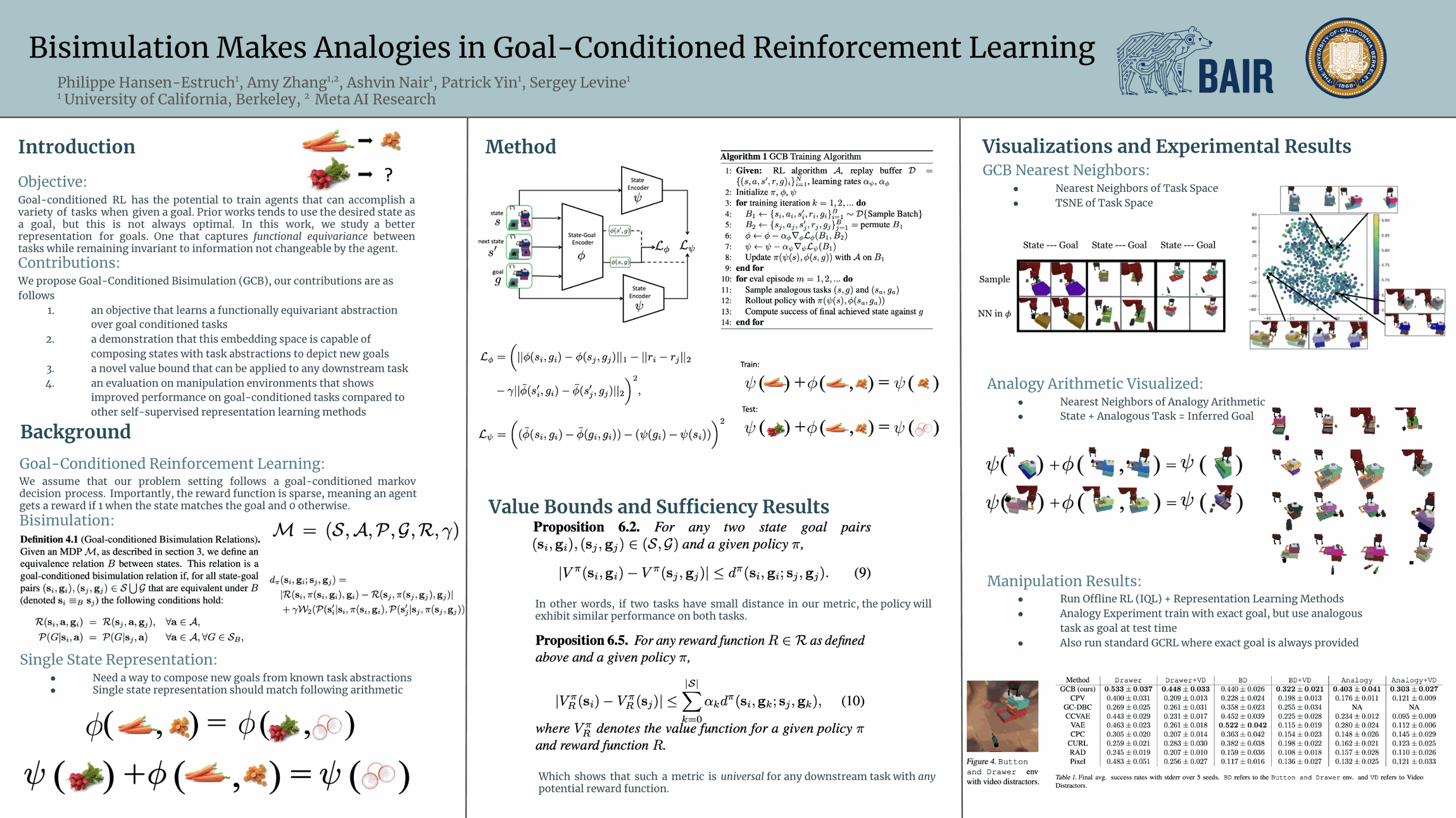}
      \subcaption{Author-designed poster.}
    \end{subfigure}
    &
    \begin{subfigure}{0.48\linewidth}
      \centering
      \includegraphics[width=\linewidth]{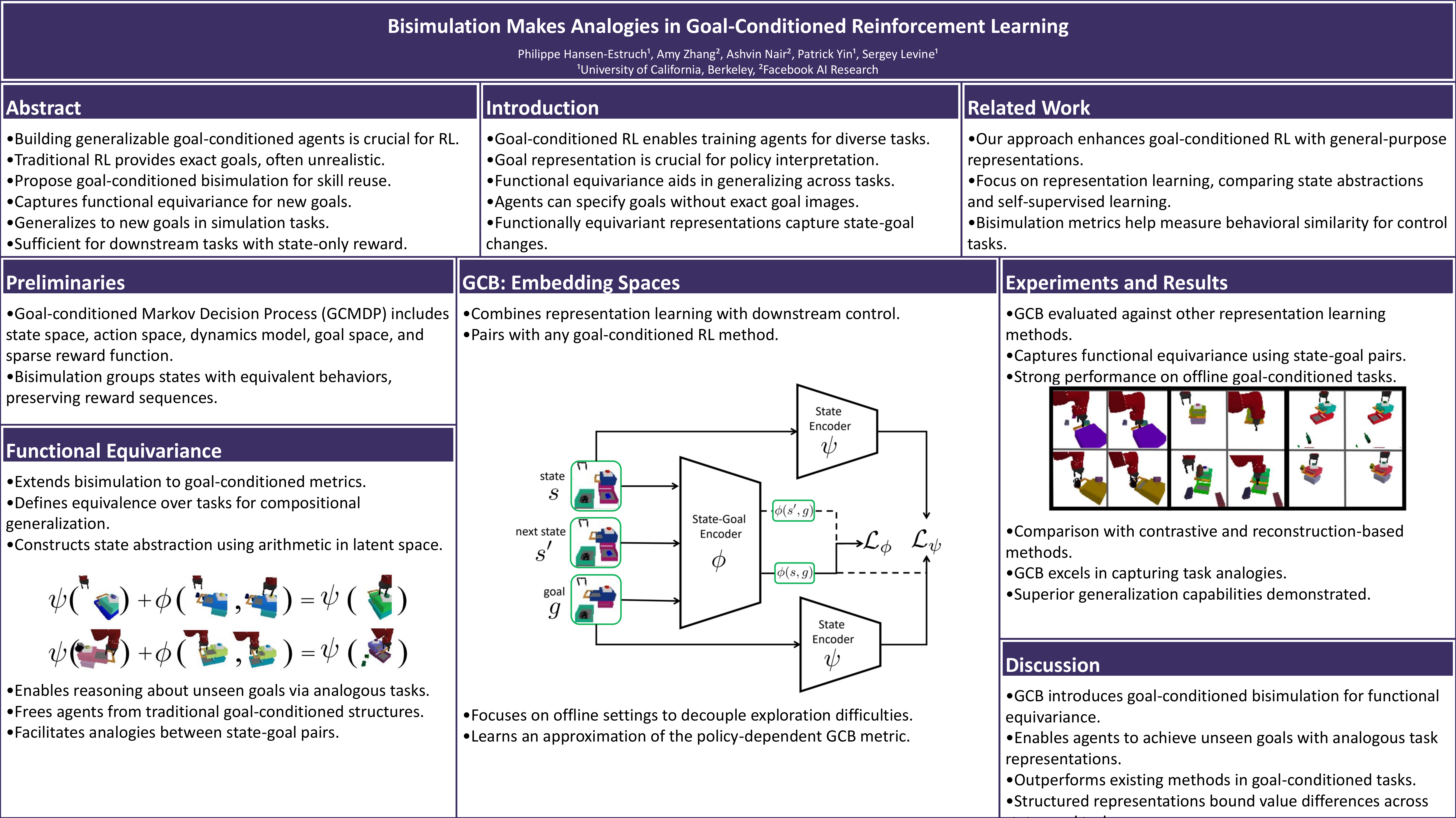}
      \subcaption{\ouralg{}-generated poster.}
    \end{subfigure}
  \end{tabular}
  \vspace{-0.8em}
  \caption{Posters for \href{https://arxiv.org/pdf/2204.13060}{{Bisimulation Makes Analogies in Goal-Conditioned Reinforcement Learning}}.}
  \label{fig:bisimulation_sample}
\end{figure*}

\begin{figure*}[!h]
  \centering
  \begin{tabular}{cc}
    \begin{subfigure}{0.48\linewidth}
      \centering
      \includegraphics[width=\linewidth]{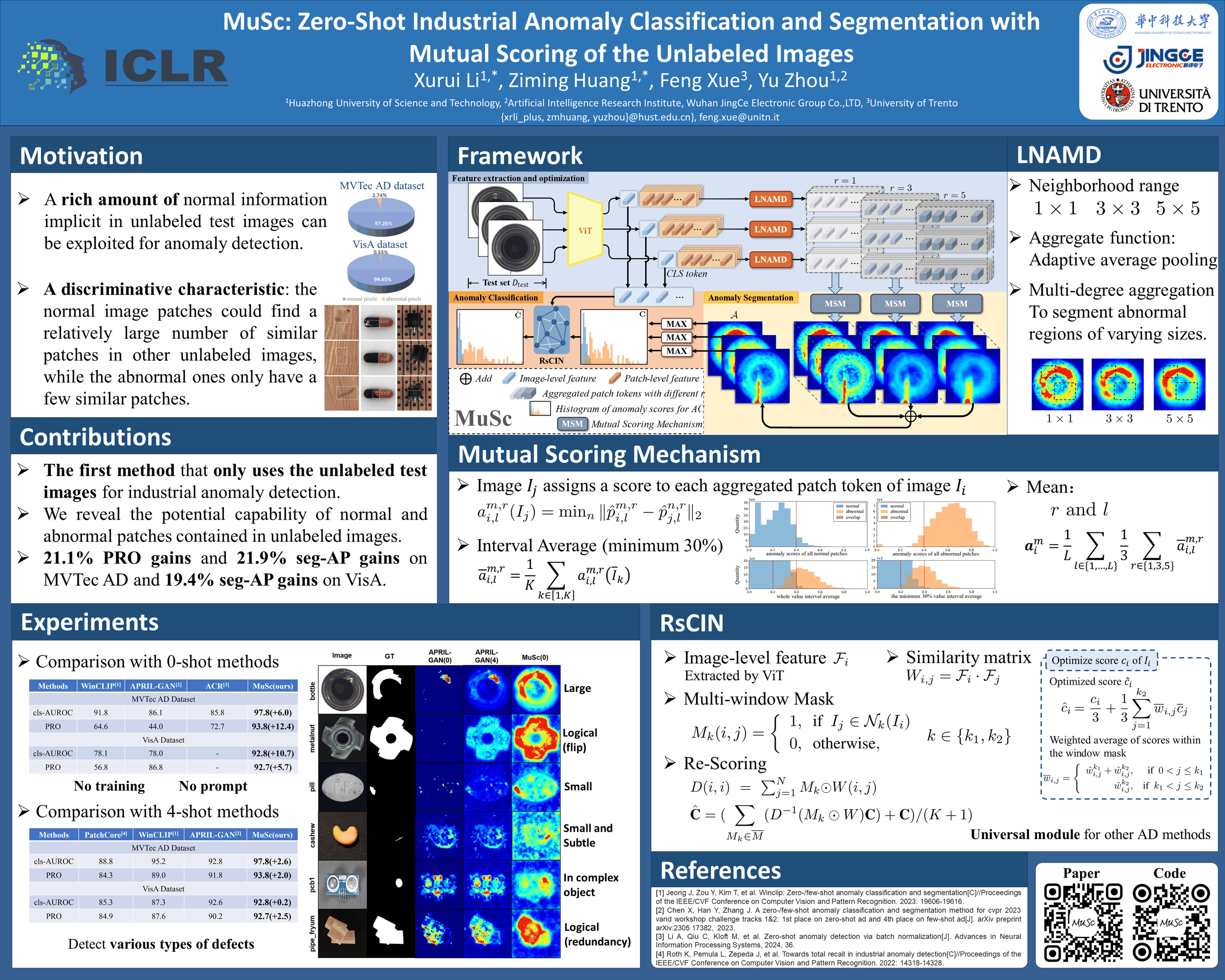}
      \subcaption{Author-designed poster.}
    \end{subfigure}
    &
    \begin{subfigure}{0.48\linewidth}
      \centering
      \includegraphics[width=\linewidth]{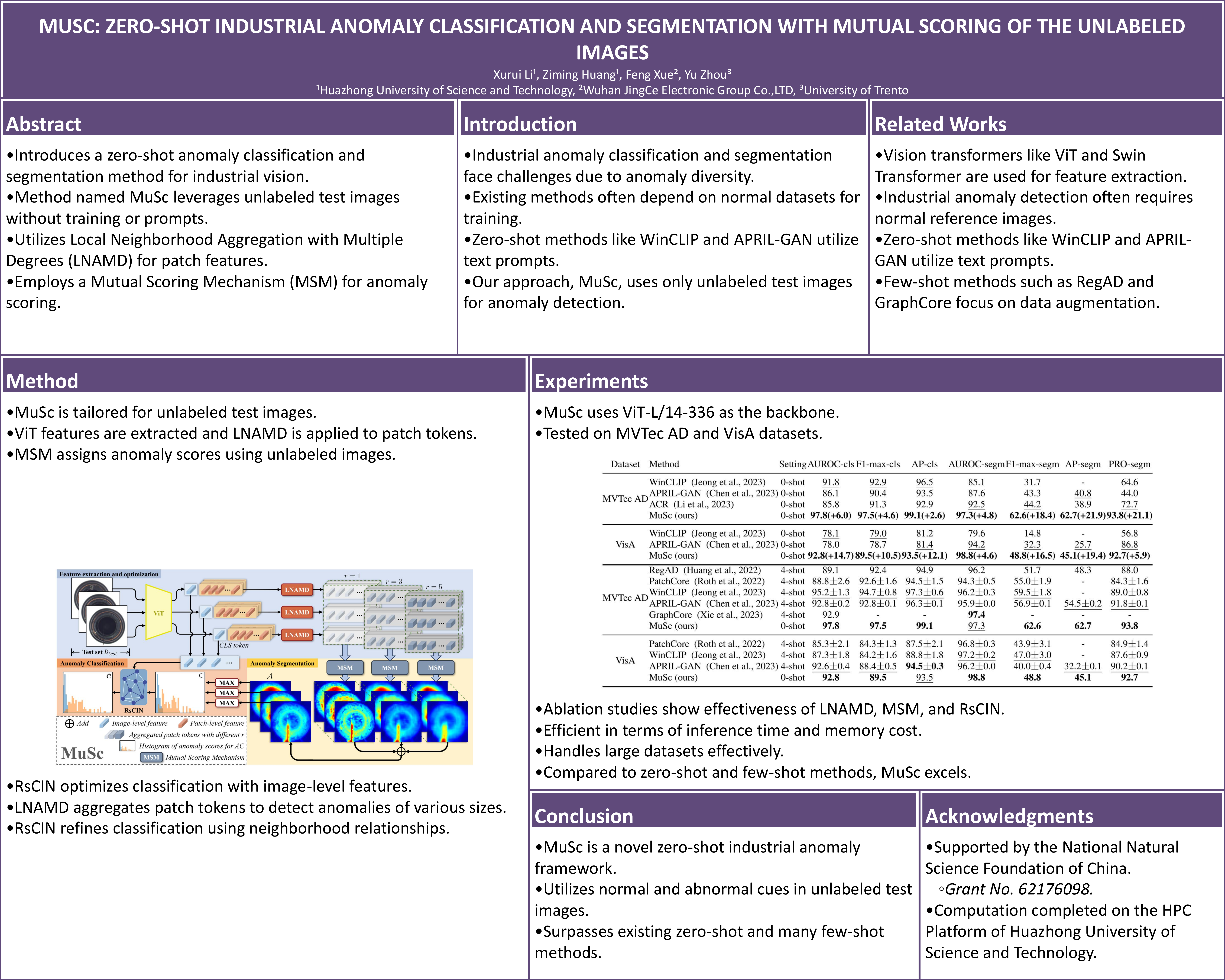}
      \subcaption{\ouralg{}-generated poster.}
    \end{subfigure}
  \end{tabular}
  \vspace{-1em}  
  \caption{Posters for \href{https://arxiv.org/pdf/2401.16753}{{MuSc: Zero-Shot Industrial Anomaly Classification and Segmentation with Mutual Scoring of the Unlabeled Images}}.}
  \label{fig:musc_sample}
\end{figure*}

\begin{figure*}[!h]
  \centering
  \vspace{2em}  
  \begin{tabular}{cc}
    \begin{subfigure}{0.48\linewidth}
      \centering
      \includegraphics[width=\linewidth]{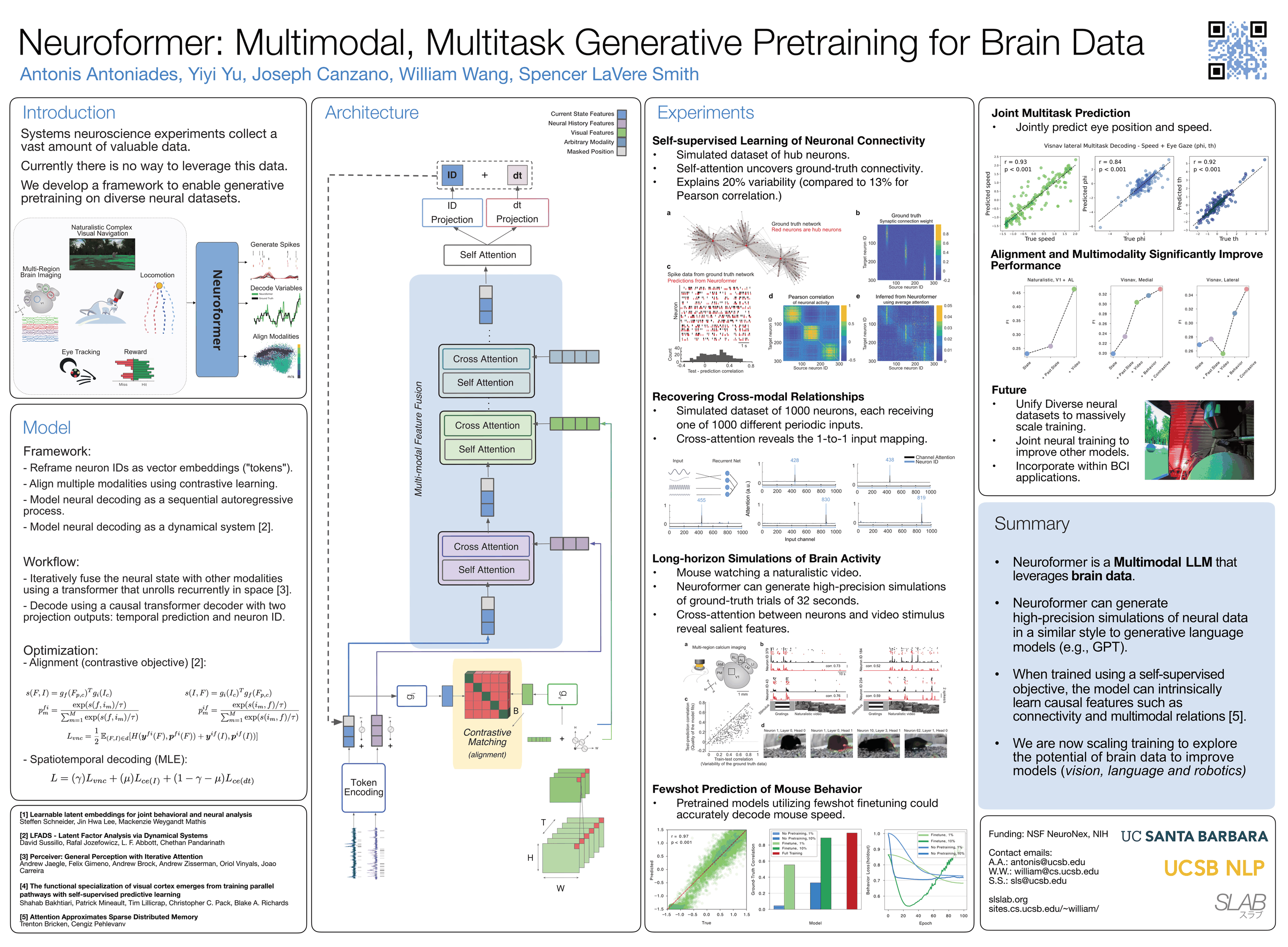}
      \subcaption{Author-designed poster.}
    \end{subfigure}
    &
    \begin{subfigure}{0.48\linewidth}
      \centering
      \includegraphics[width=\linewidth]{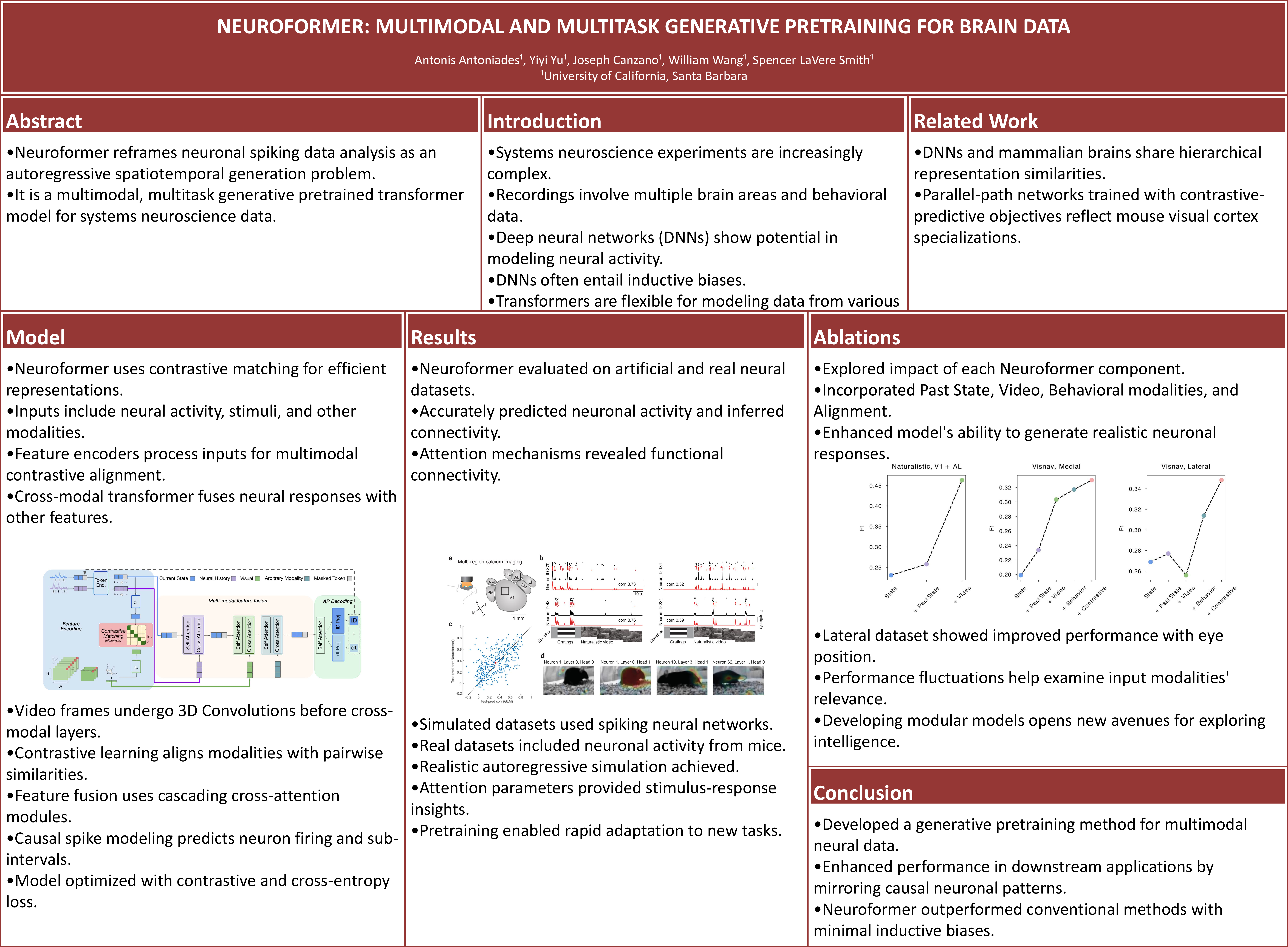}
      \subcaption{\ouralg{}-generated poster.}
    \end{subfigure}
  \end{tabular}
  \vspace{-1em}  
  \caption{Posters for \href{https://arxiv.org/pdf/2311.00136}{{Neuroformer: Multimodal and Multitask Generative Pretraining for Brain Data}}.}
  \label{fig:neuroformer_sample}
\end{figure*}

\begin{figure*}[!h]
  \centering
  \begin{tabular}{cc}
    \begin{subfigure}{0.48\linewidth}
      \centering
      \includegraphics[width=\linewidth]{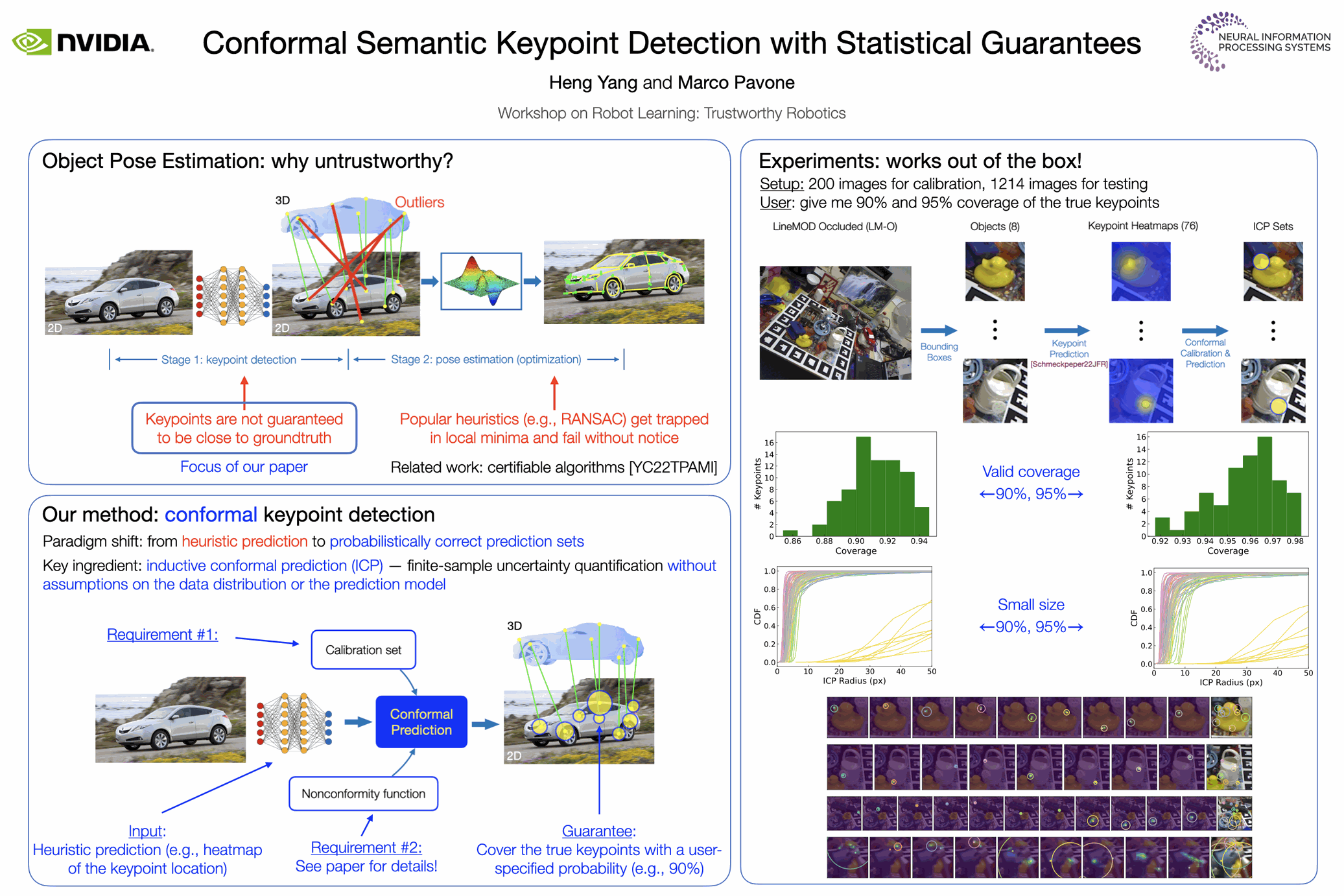}
      \subcaption{Author-designed poster.}
    \end{subfigure}
    &
    \begin{subfigure}{0.48\linewidth}
      \centering
      \includegraphics[width=\linewidth]{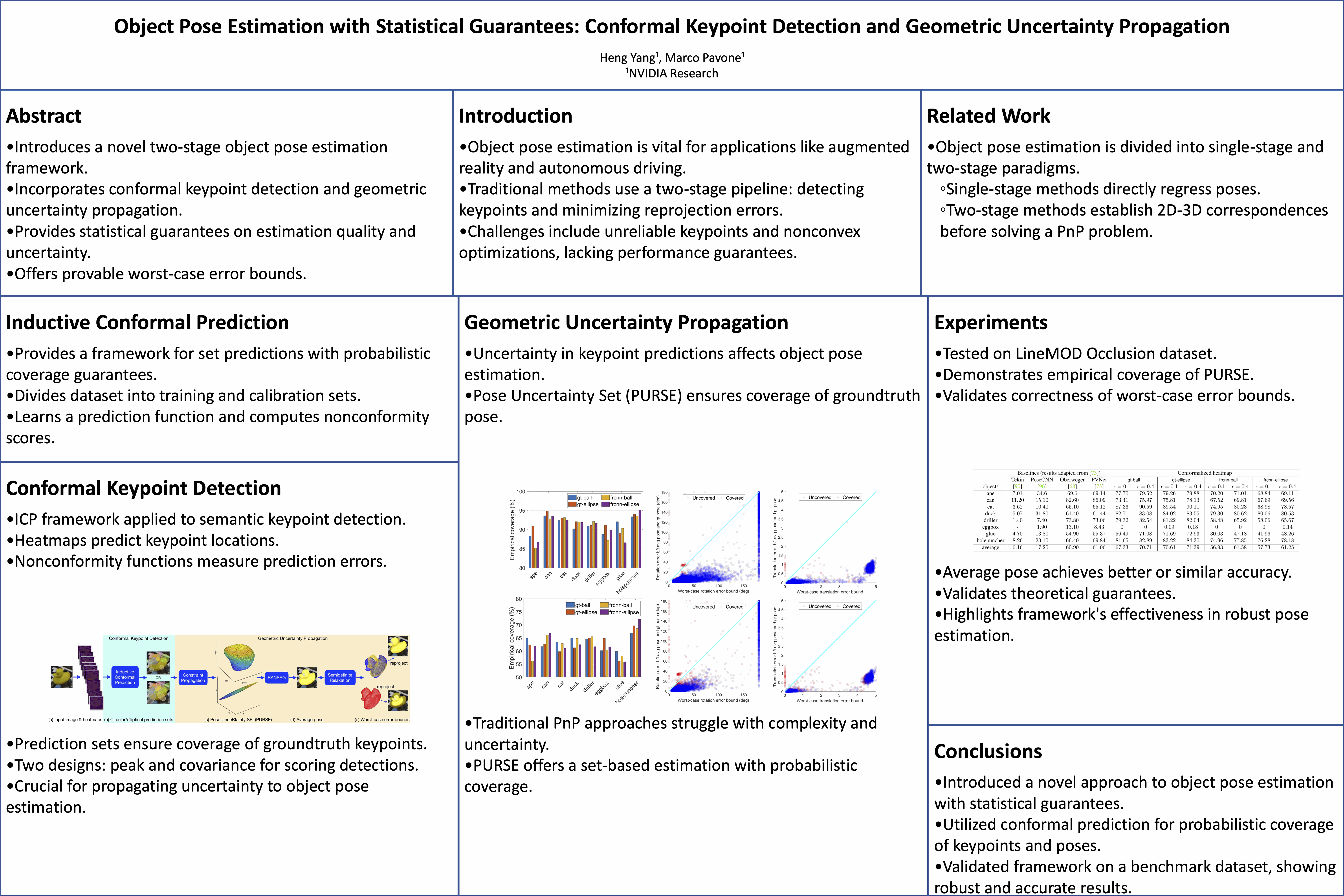}
      \subcaption{\ouralg{}-generated poster.}
    \end{subfigure}
  \end{tabular}
  \vspace{-1em}  
  \caption{Posters for \href{https://arxiv.org/pdf/2303.12246}{{Conformal Semantic Keypoint Detection with Statistical Guarantees}}.}
  \label{fig:conformal_sample}
\end{figure*}

\begin{figure*}[!h]
  \centering
  \begin{tabular}{cc}
    \begin{subfigure}{0.59\linewidth}
      \centering
      \includegraphics[width=\linewidth]{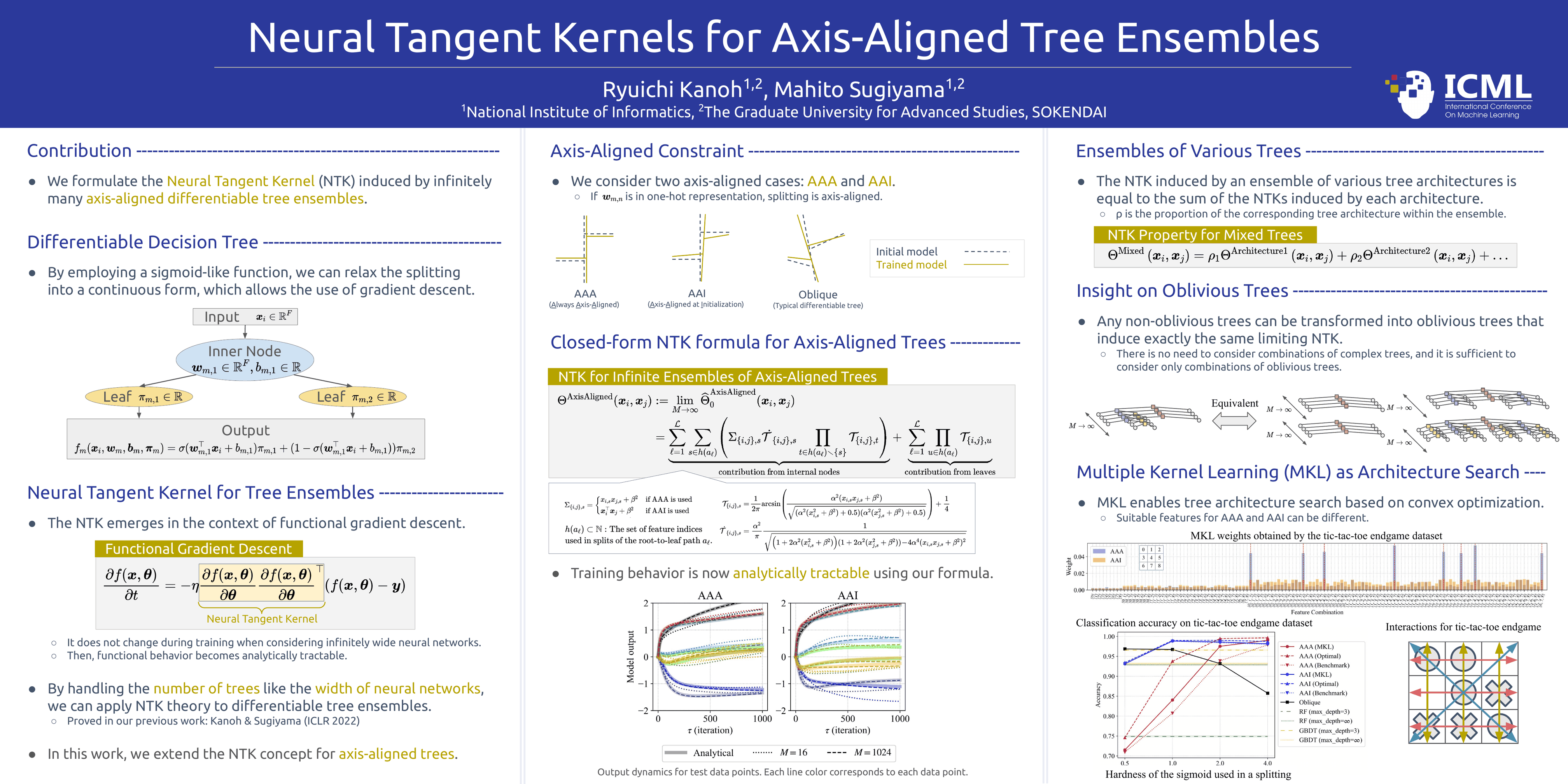}
      \subcaption{Author-designed poster.}
    \end{subfigure}
    &
    \begin{subfigure}{0.395\linewidth}
      \centering
      \includegraphics[width=\linewidth]{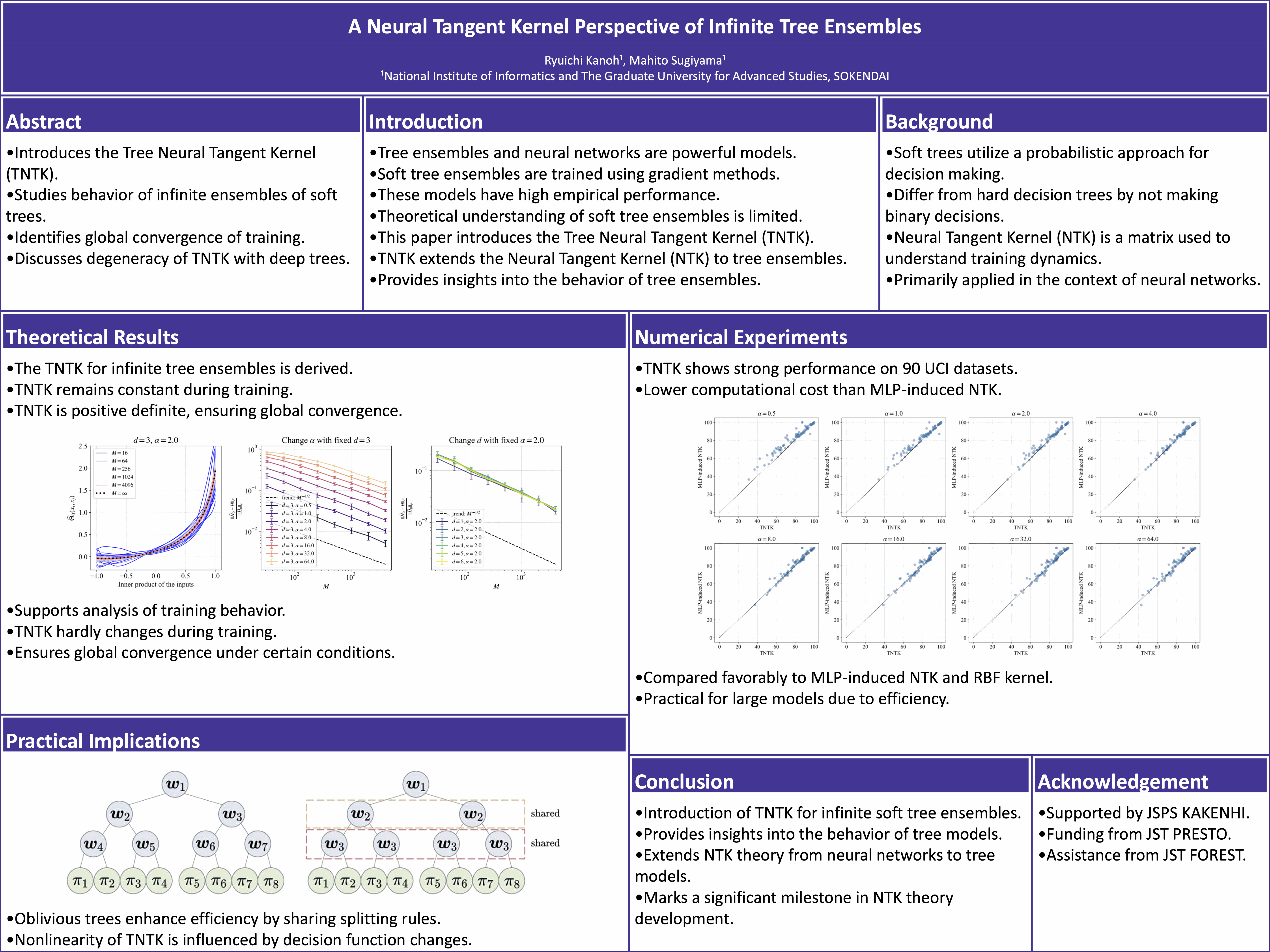}
      \subcaption{\ouralg{}-generated poster.}
    \end{subfigure}
  \end{tabular}
  \vspace{-1em}  
  \caption{Posters for \href{https://arxiv.org/pdf/2109.04983}{{Neural Tangent Kernels for Axis-Aligned Tree Ensembles}}.}
  \label{fig:neural_tangent}
\end{figure*}

\begin{figure*}[!h]
  \centering
  \begin{tabular}{cc}
    \begin{subfigure}{0.48\linewidth}
      \centering
      \includegraphics[width=\linewidth]{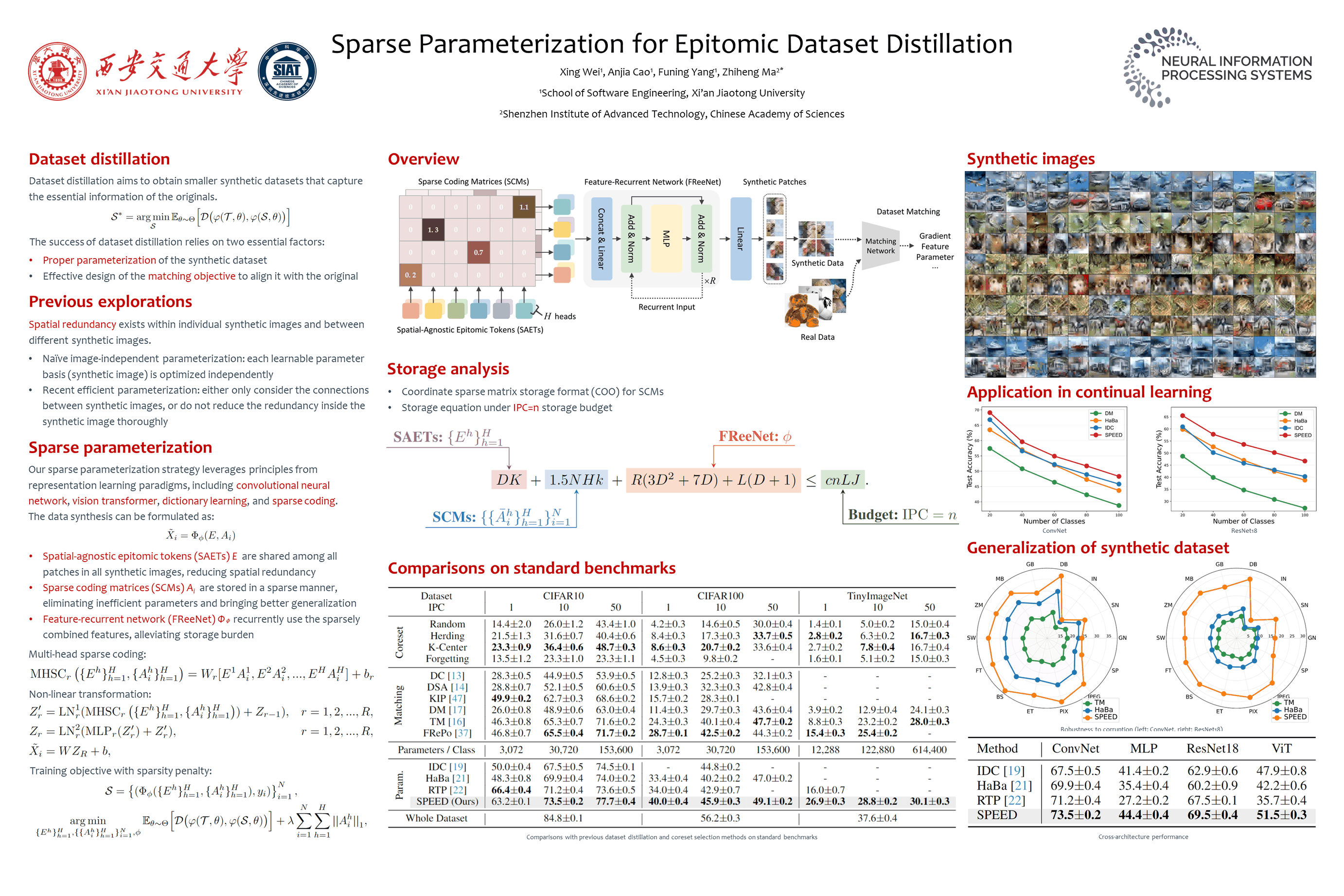}
      \subcaption{Author-designed poster.}
    \end{subfigure}
    &
    \begin{subfigure}{0.48\linewidth}
      \centering
      \includegraphics[width=\linewidth]{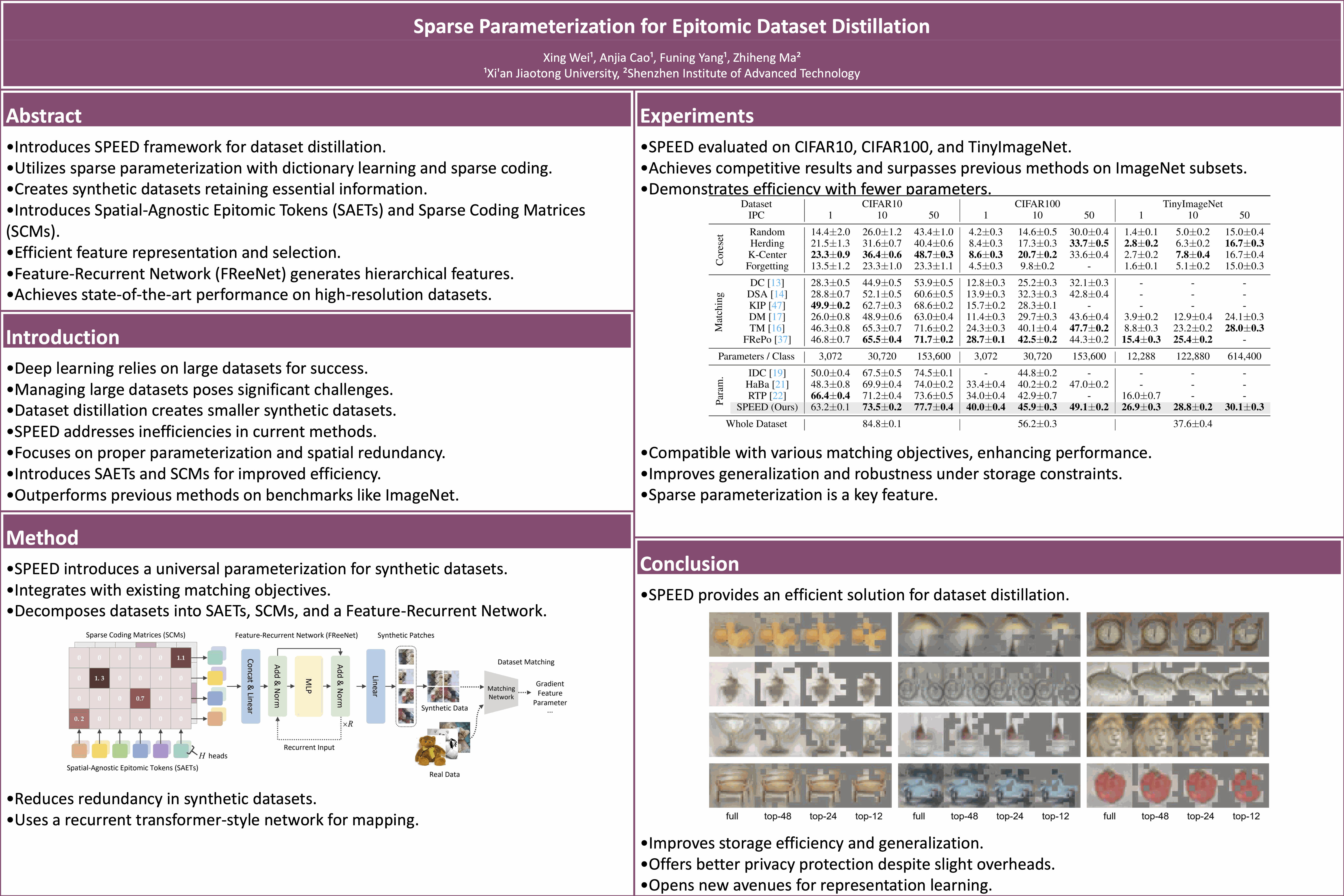}
      \subcaption{\ouralg{}-generated poster.}
    \end{subfigure}
  \end{tabular}
  \vspace{-1em}  
  \caption{Posters for \href{https://papers.nips.cc/paper_files/paper/2023/file/9e8889198d16fb79926e71adbe38cae4-Paper-Conference.pdf}{{Sparse Parameterization for Epitomic Dataset Distillation}}.}
  \label{fig:sparse_parameterization}
\end{figure*}

\begin{figure*}[!h]
  \centering
  \begin{tabular}{cc}
    \begin{subfigure}{0.6\linewidth}
      \centering
      \includegraphics[width=\linewidth]{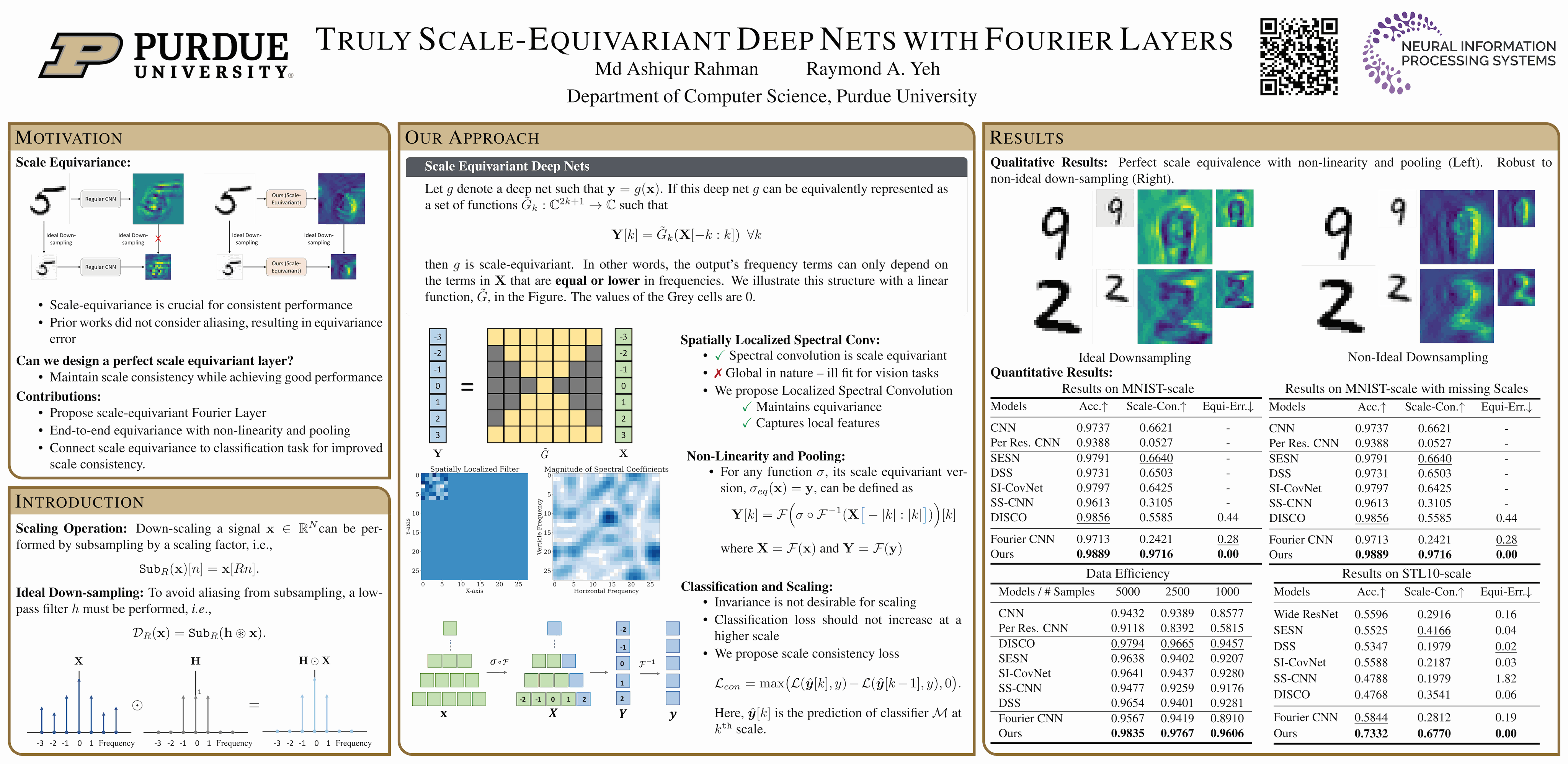}
      \subcaption{Author-designed poster.}
    \end{subfigure}
    &
    \begin{subfigure}{0.38\linewidth}
      \centering
      \includegraphics[width=\linewidth]{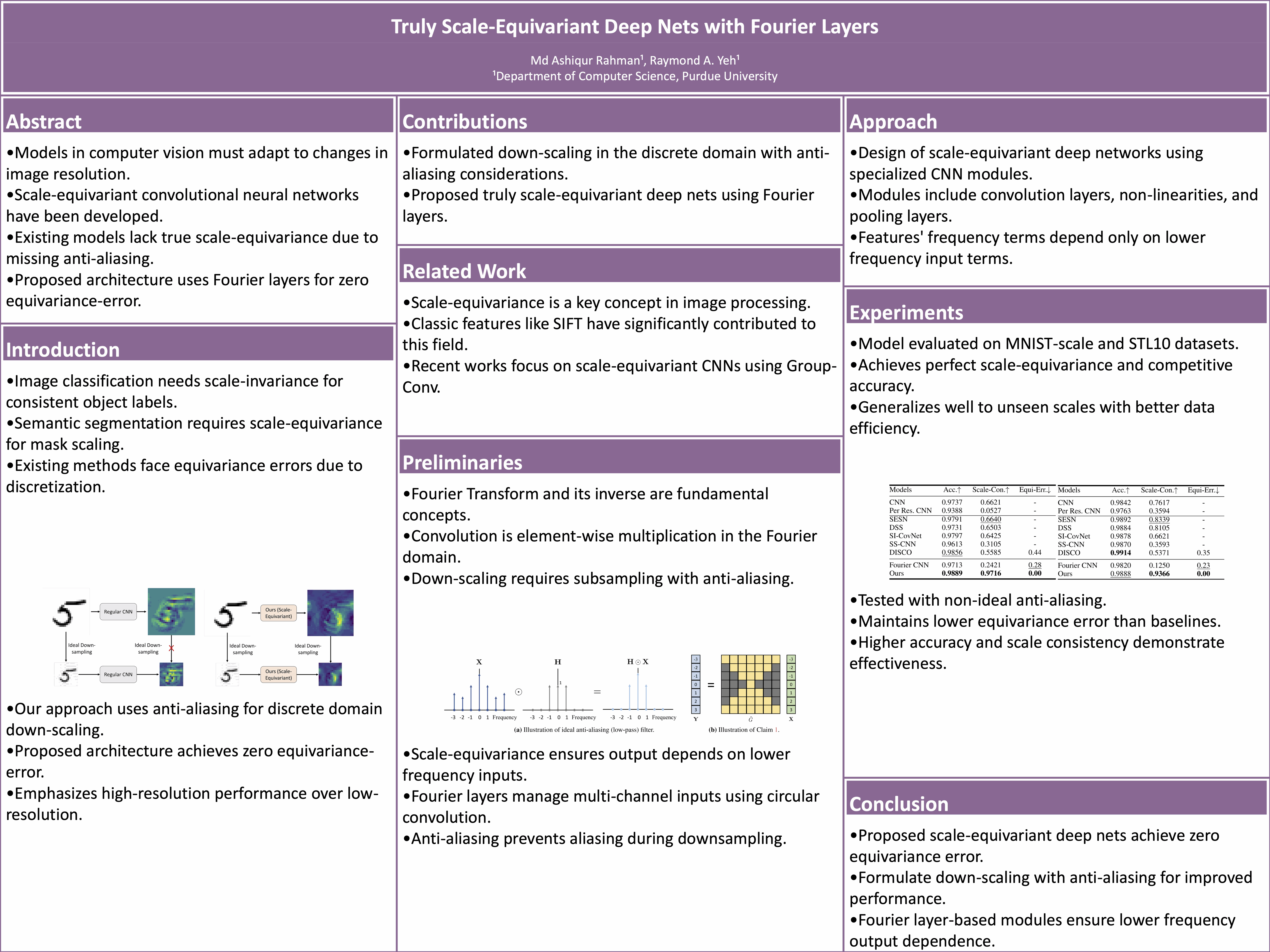}
      \subcaption{\ouralg{}-generated poster.}
    \end{subfigure}
  \end{tabular}
  \vspace{-1em}  
  \caption{Posters for \href{https://proceedings.neurips.cc/paper_files/paper/2023/file/1343edb2739a61a6e20bd8764e814b50-Paper-Conference.pdf}{{Truly Scale-Equivariant Deep Nets with Fourier Layers}}.}
  \label{fig:truly}
\end{figure*}

\begin{figure*}[!h]
  \centering
  \begin{tabular}{cc}
    \begin{subfigure}{0.48\linewidth}
      \centering
      \includegraphics[width=\linewidth]{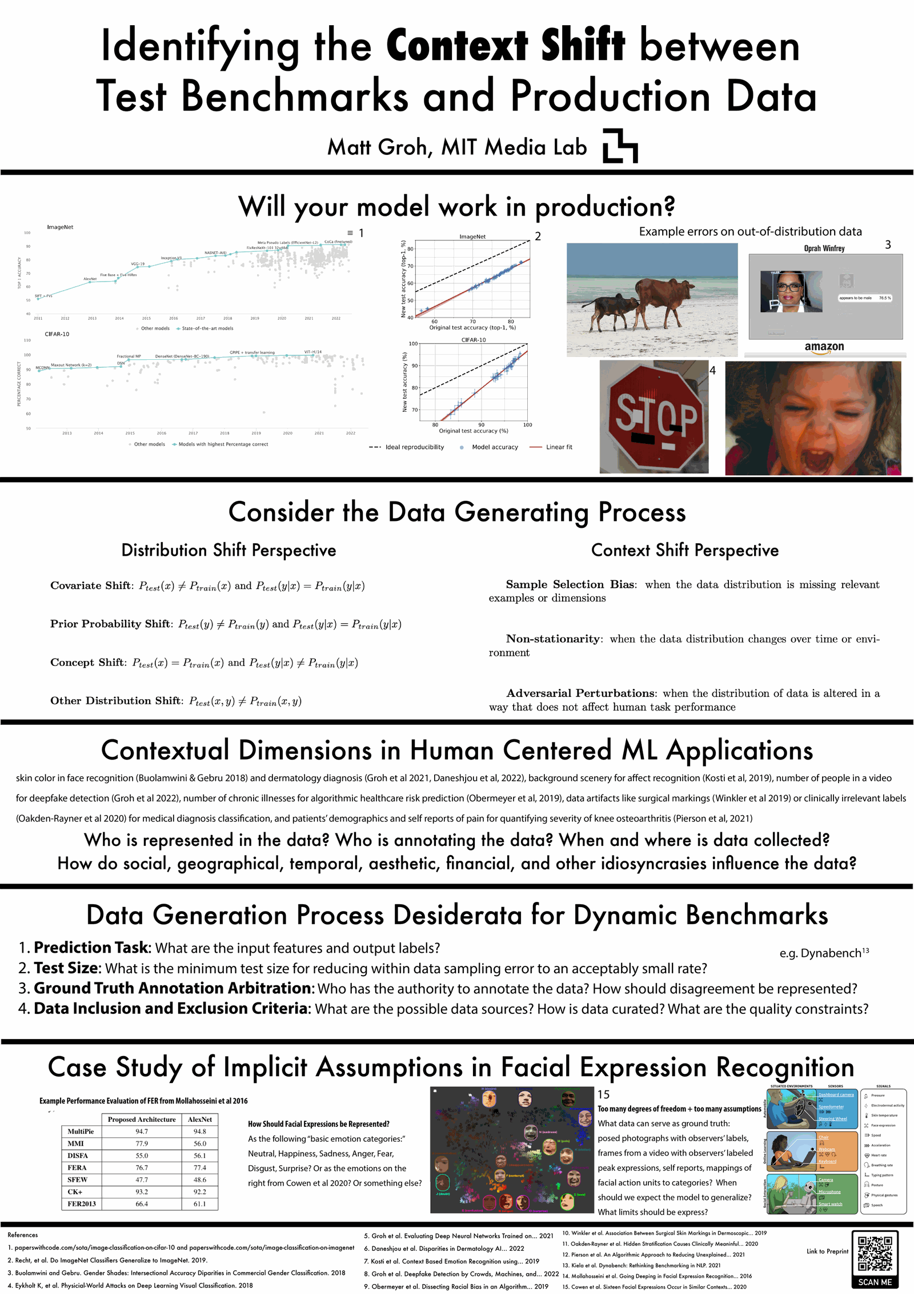}
      \subcaption{Author-designed poster.}
    \end{subfigure}
    &
    \begin{subfigure}{0.48\linewidth}
      \centering
      \includegraphics[width=\linewidth]{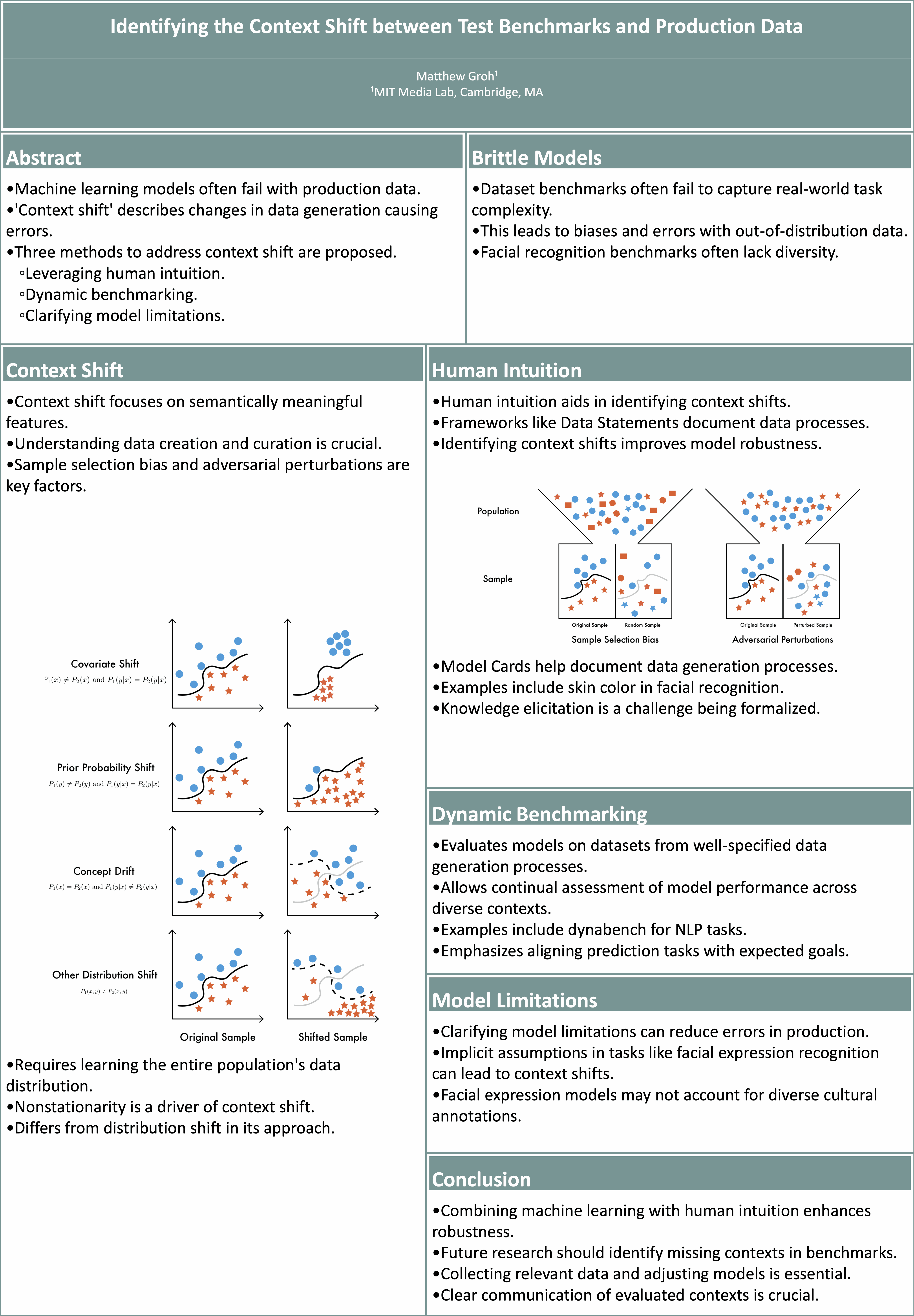}
      \subcaption{\ouralg{}-generated poster.}
    \end{subfigure}
  \end{tabular}
  \vspace{-1em}  
  \caption{Posters for \href{https://arxiv.org/pdf/2207.01059}{{Identifying the Context Shift between Test Benchmarks and Production Data}}.}
  \label{fig:identifying_context}
\end{figure*}

\section{Ablation Study}
We conduct ablation studies to evaluate three key design choices in \ouralg{}: (1) the binary-tree layout strategy for layout planning; (2) the inclusion of a commenter module as a visual critic; and (3) the use of in-context examples to enhance the visual perception capabilities of the commenter.

We define the following variants:
\begin{itemize}
	\item \textit{Direct}: replacing the binary-tree layout with direct layout generation by an LLM;
	\item \textit{Tree}: using the binary-tree layout strategy but removing the commenter module;
	\item \textit{Tree + Commenter}: including the commenter module but without in-context examples;
	\item \textit{Tree + Commenter + IC}: the full system, with both the commenter and in-context examples.
\end{itemize}

\begin{figure}[htbp]
  \centering
  \begin{subfigure}[b]{0.45\textwidth}
    \includegraphics[width=\linewidth]{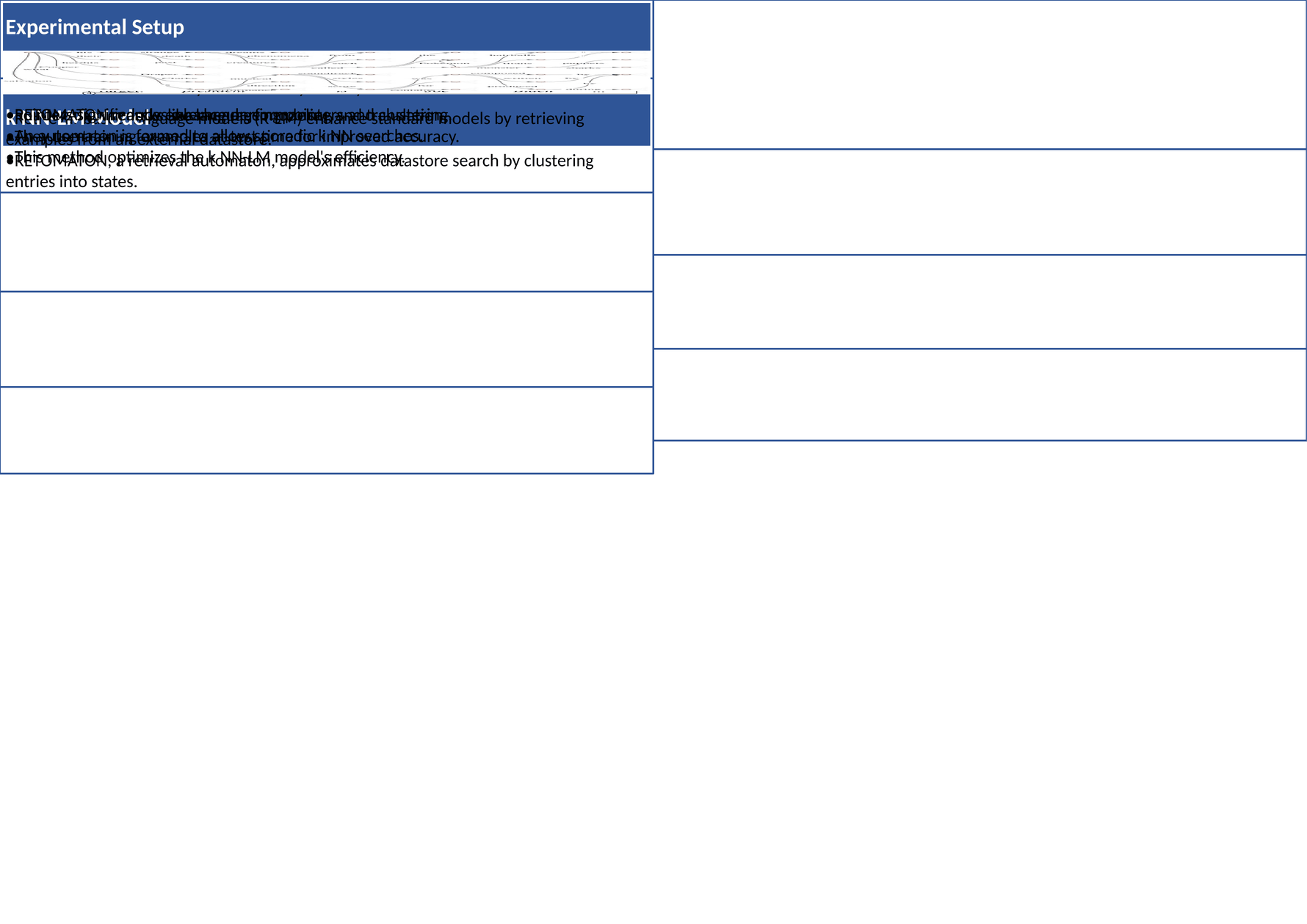}
    \caption{Direct.}
  \label{fig:ablation1_a}
  \end{subfigure}
  \hfill
  \begin{subfigure}[b]{0.45\textwidth}
    \includegraphics[width=\linewidth]{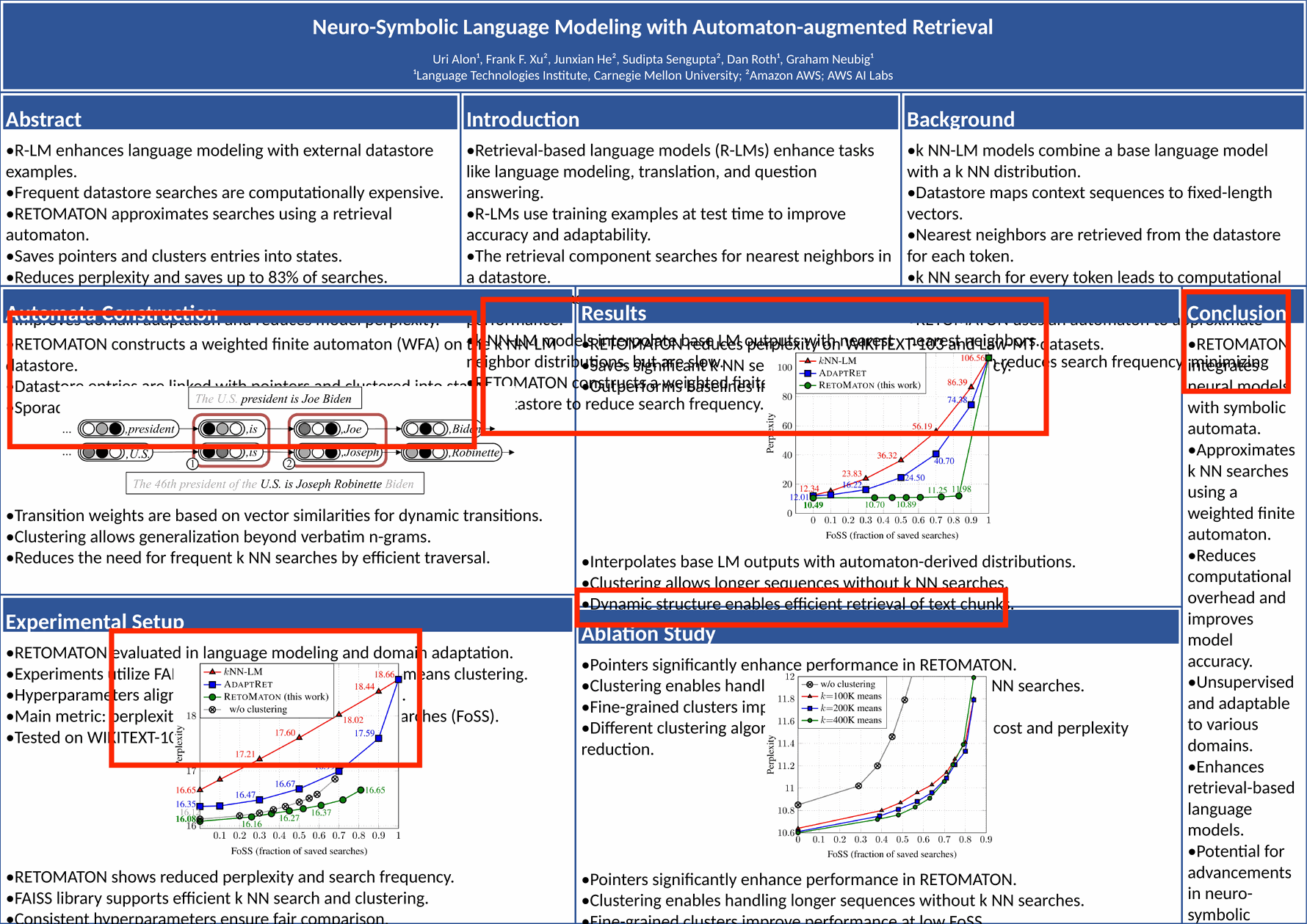}
    \caption{Tree.}
    \label{fig:ablation1_b}
  \end{subfigure}

  \vspace{1em}

  \begin{subfigure}[b]{0.45\textwidth}
    \includegraphics[width=\linewidth]{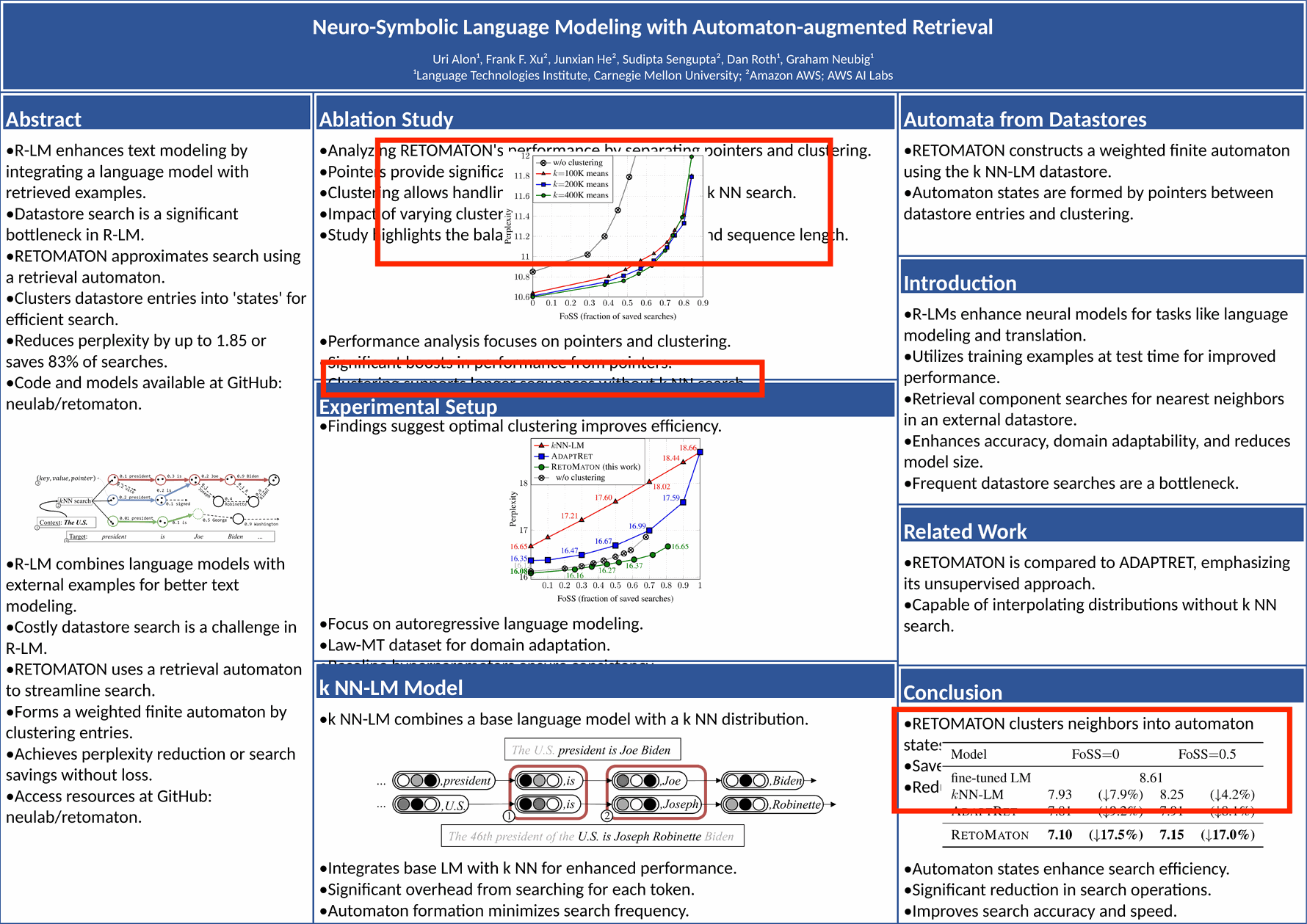}
    \caption{Tree + Commenter.}
  \label{fig:ablation1_c}
  \end{subfigure}
  \hfill
  \begin{subfigure}[b]{0.45\textwidth}
    \includegraphics[width=\linewidth]{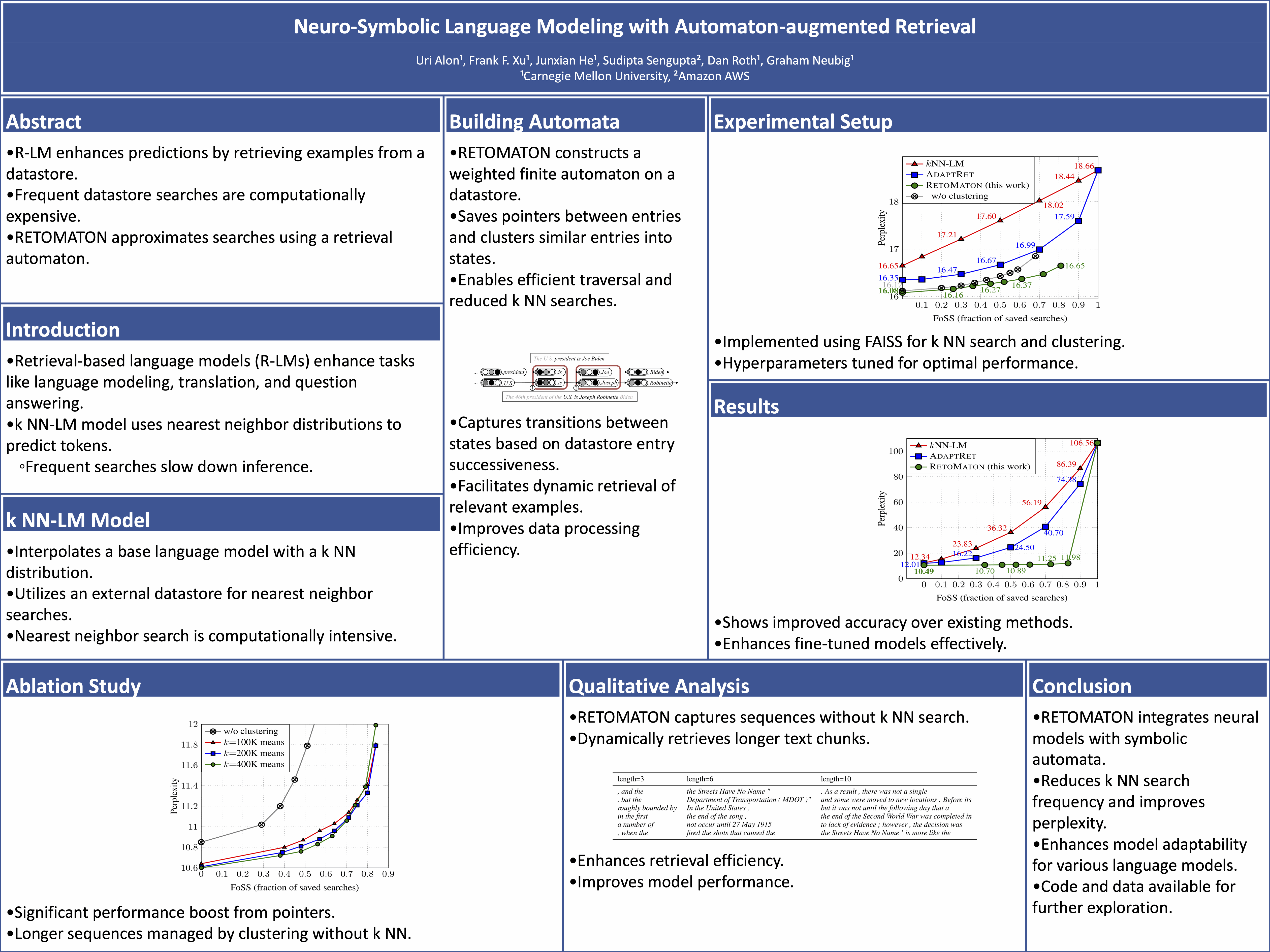}
    \caption{Tree + Commenter + IC.}
    \label{fig:ablation1_d}
  \end{subfigure}

  \caption{Ablation study on \href{https://arxiv.org/pdf/2201.12431}{Neuro-Symbolic Language Modeling with Automaton-augmented Retrieval}. Text overflow areas are highlighted with \textcolor{red}{red} bounding boxes.}
  \label{fig:ablation1}
\end{figure}

\begin{figure}[htbp]
  \centering
  \begin{subfigure}[b]{0.45\textwidth}
    \includegraphics[width=\linewidth]{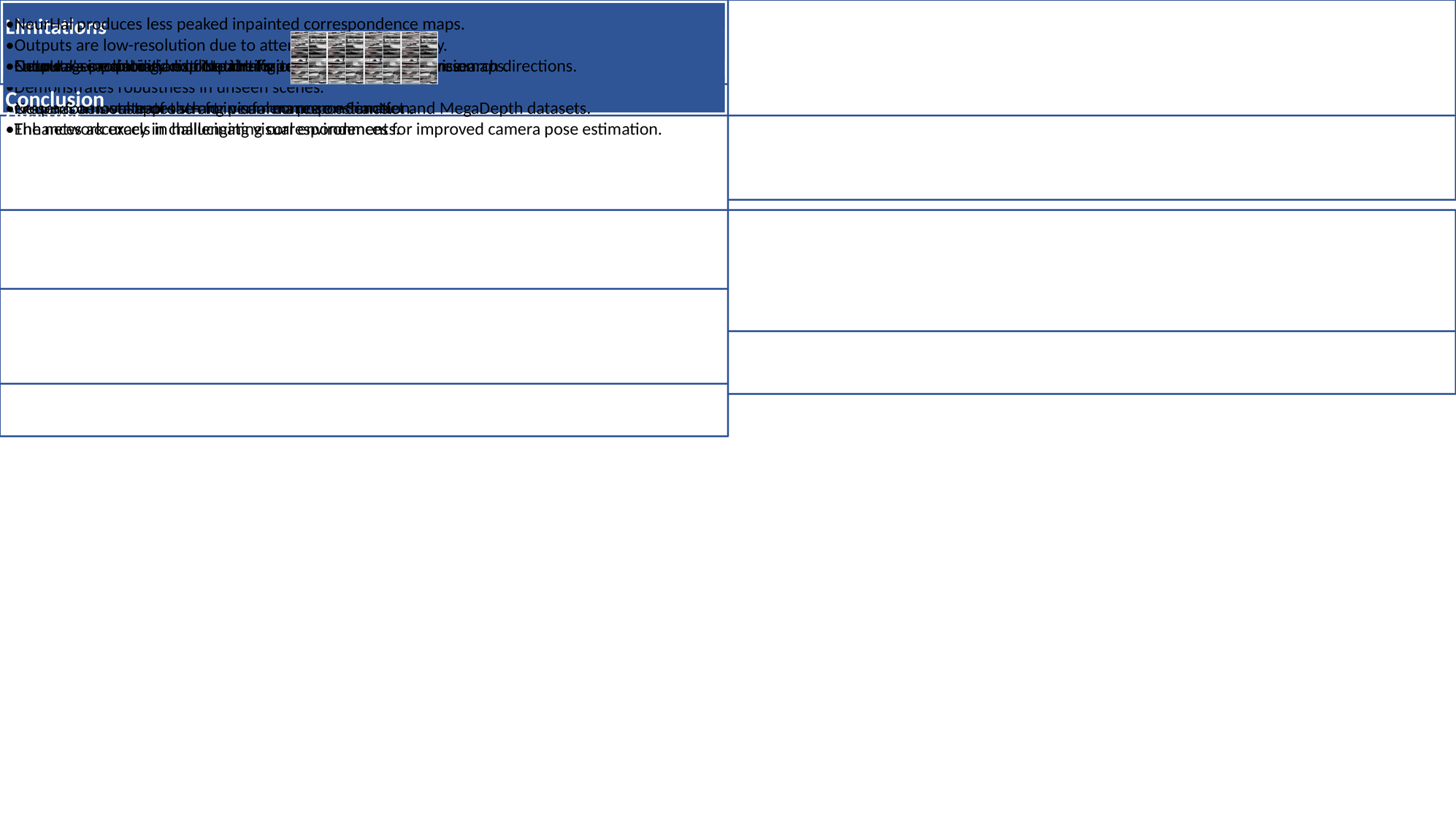}
    \caption{Direct.}
  \label{fig:ablation2_a}
  \end{subfigure}
  \hfill
  \begin{subfigure}[b]{0.45\textwidth}
    \includegraphics[width=\linewidth]{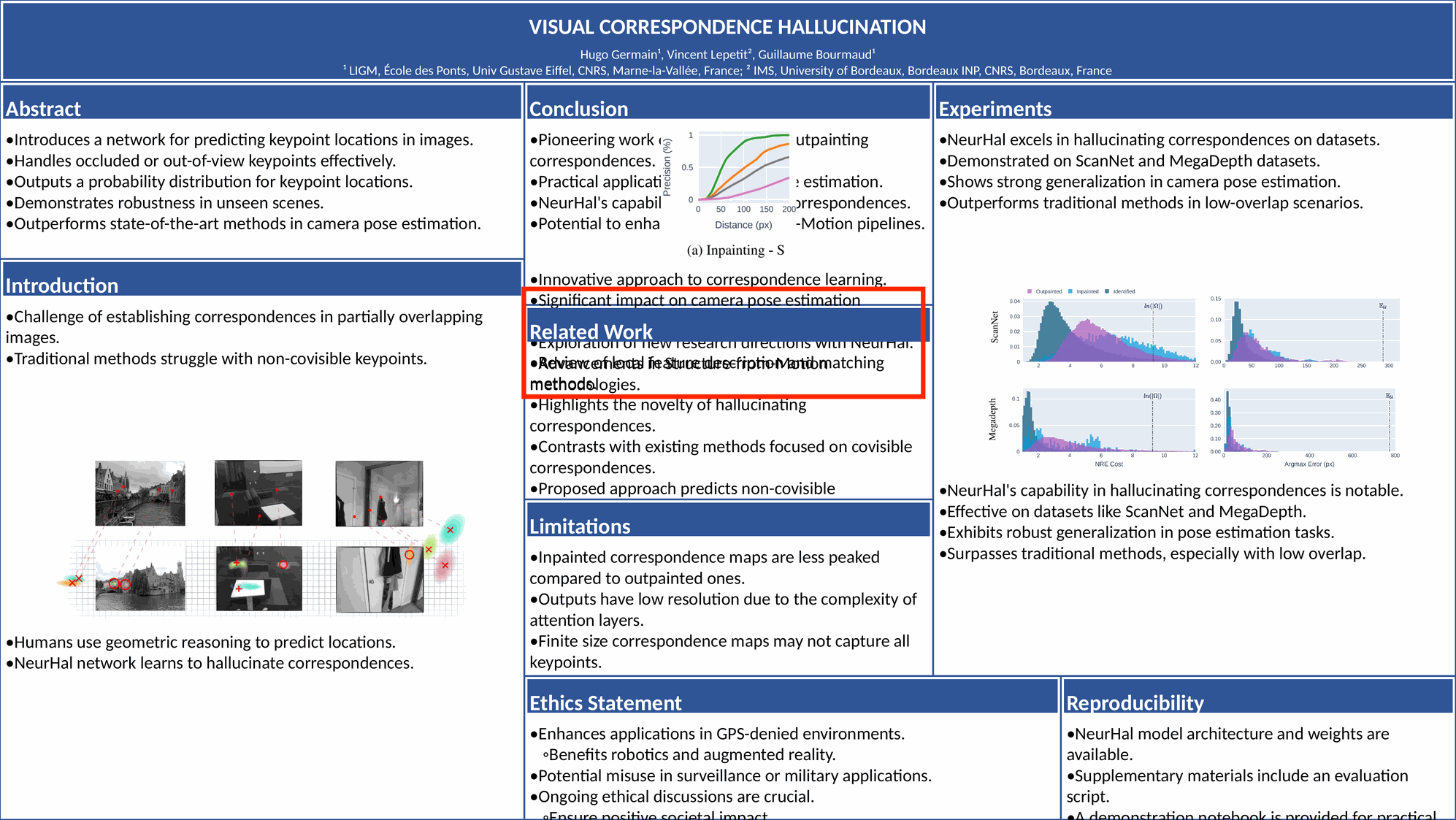}
    \caption{Tree.}
    \label{fig:ablation2_b}
  \end{subfigure}

  \vspace{1em}

  \begin{subfigure}[b]{0.45\textwidth}
    \includegraphics[width=\linewidth]{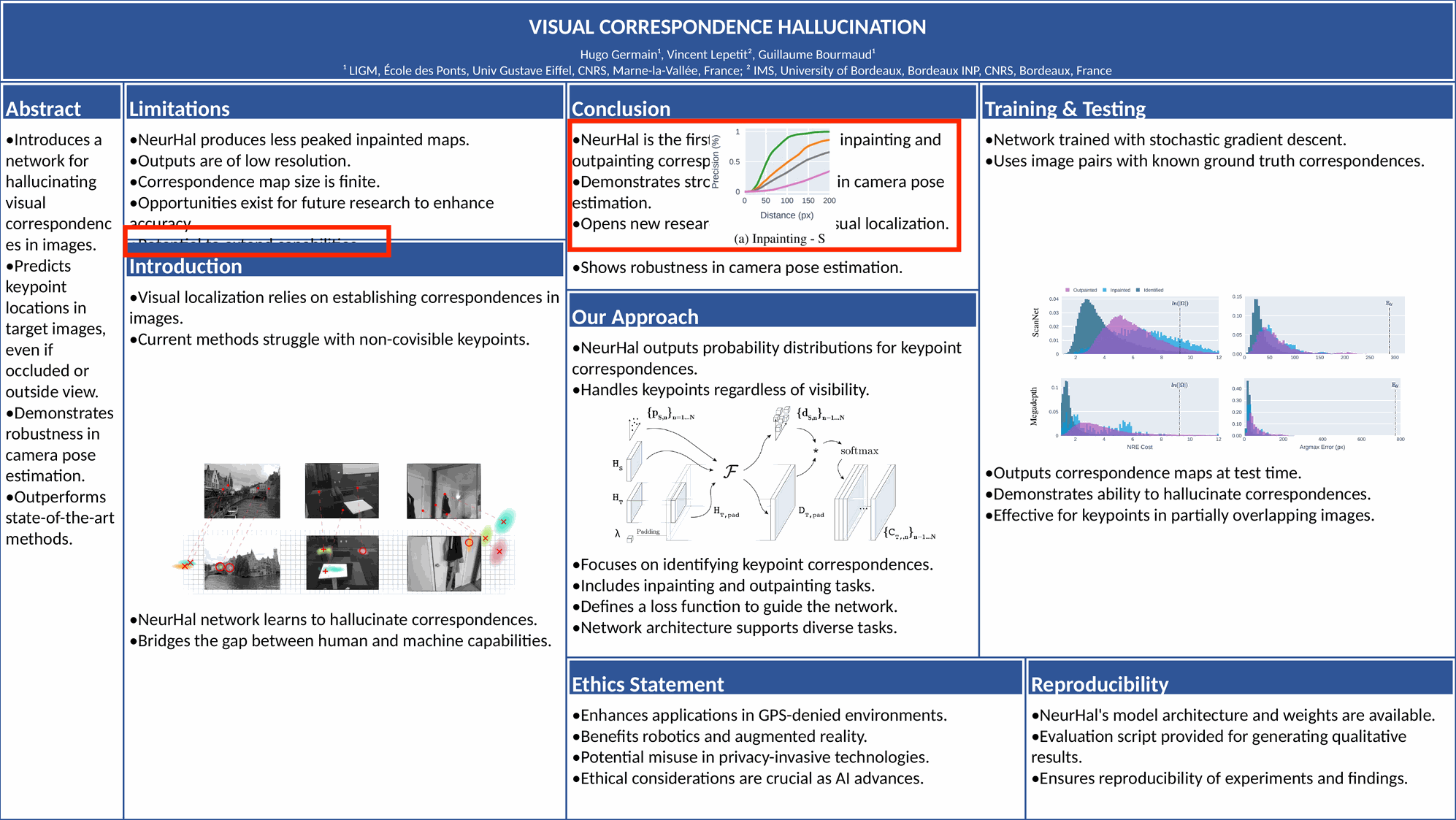}
    \caption{Tree + Commenter.}
    \label{fig:ablation2_c}
  \end{subfigure}
  \hfill
  \begin{subfigure}[b]{0.45\textwidth}
    \includegraphics[width=\linewidth]{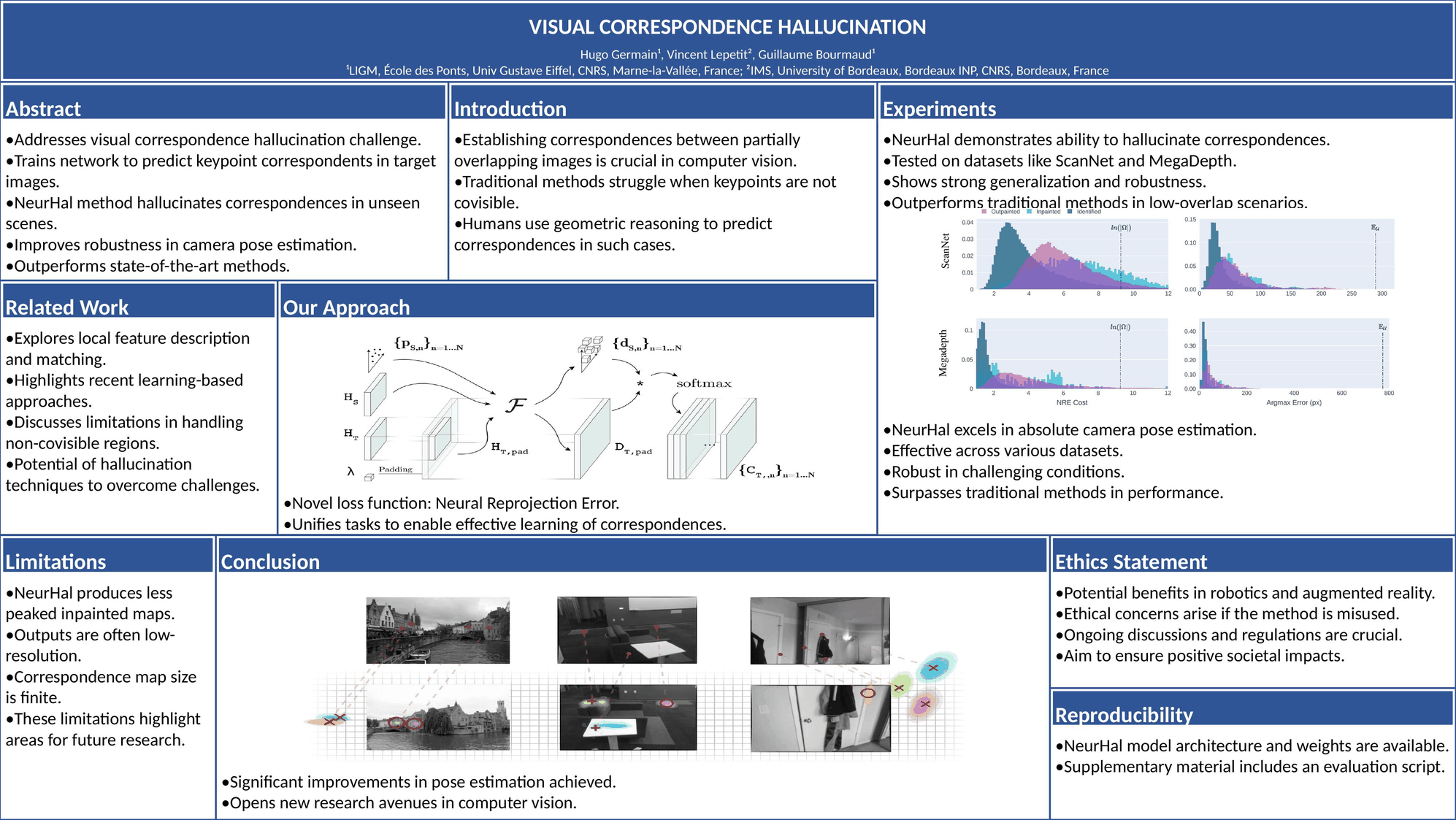}
    \caption{Tree + Commenter + IC.}
    \label{fig:ablation2_d}
  \end{subfigure}

  \caption{Ablation study on \href{https://arxiv.org/pdf/2106.09711}{Visual Correspondence Hallucination}. Text overflow areas are highlighted with \textcolor{red}{red} bounding boxes.}
  \label{fig:ablation2}
\end{figure}

\begin{figure}[htbp]
  \centering
  \begin{subfigure}[b]{0.45\textwidth}
    \includegraphics[width=\linewidth]{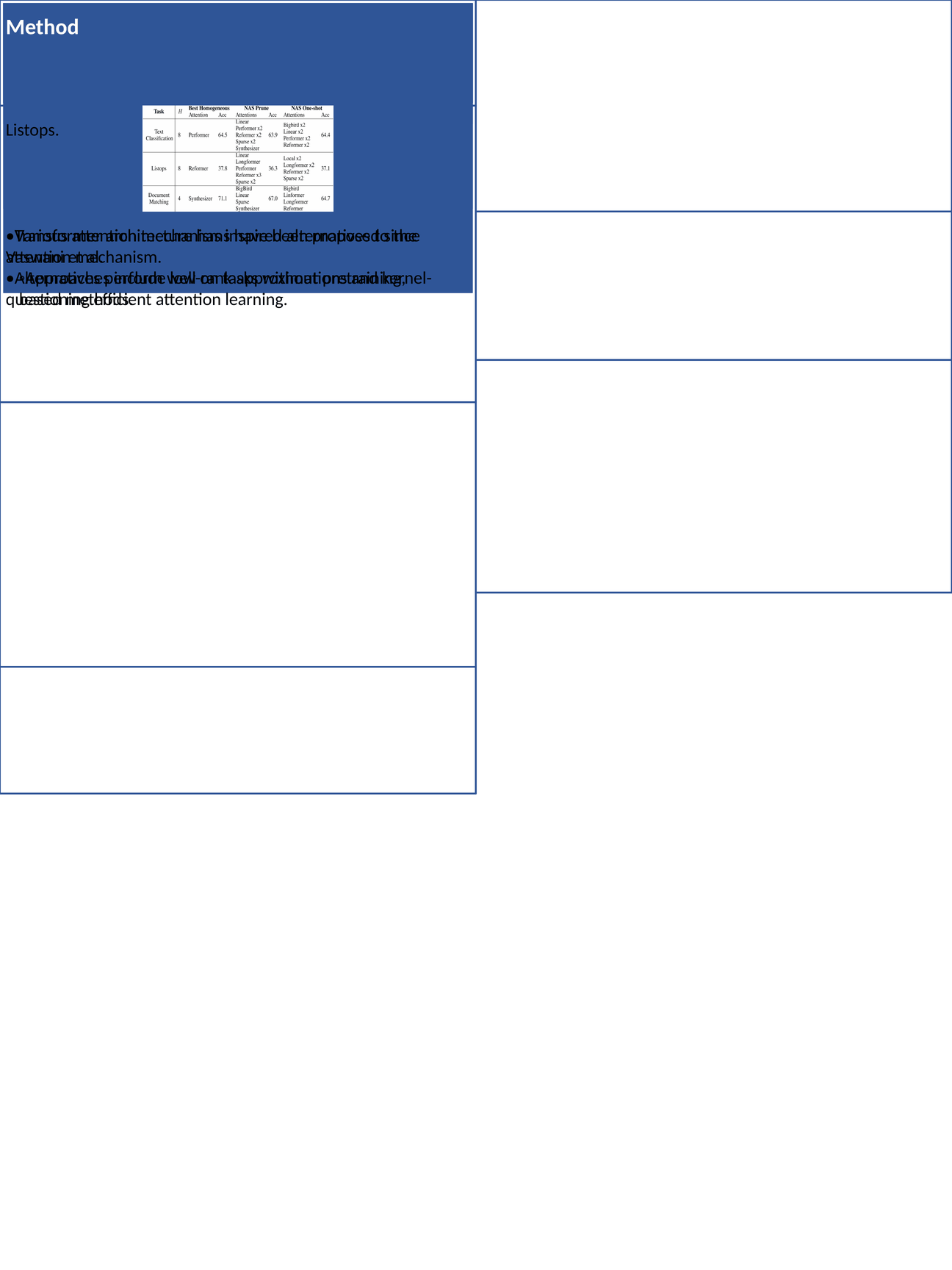}
    \caption{Direct.}
    \label{fig:ablation3_a}
  \end{subfigure}
  \hfill
  \begin{subfigure}[b]{0.45\textwidth}
    \includegraphics[width=\linewidth]{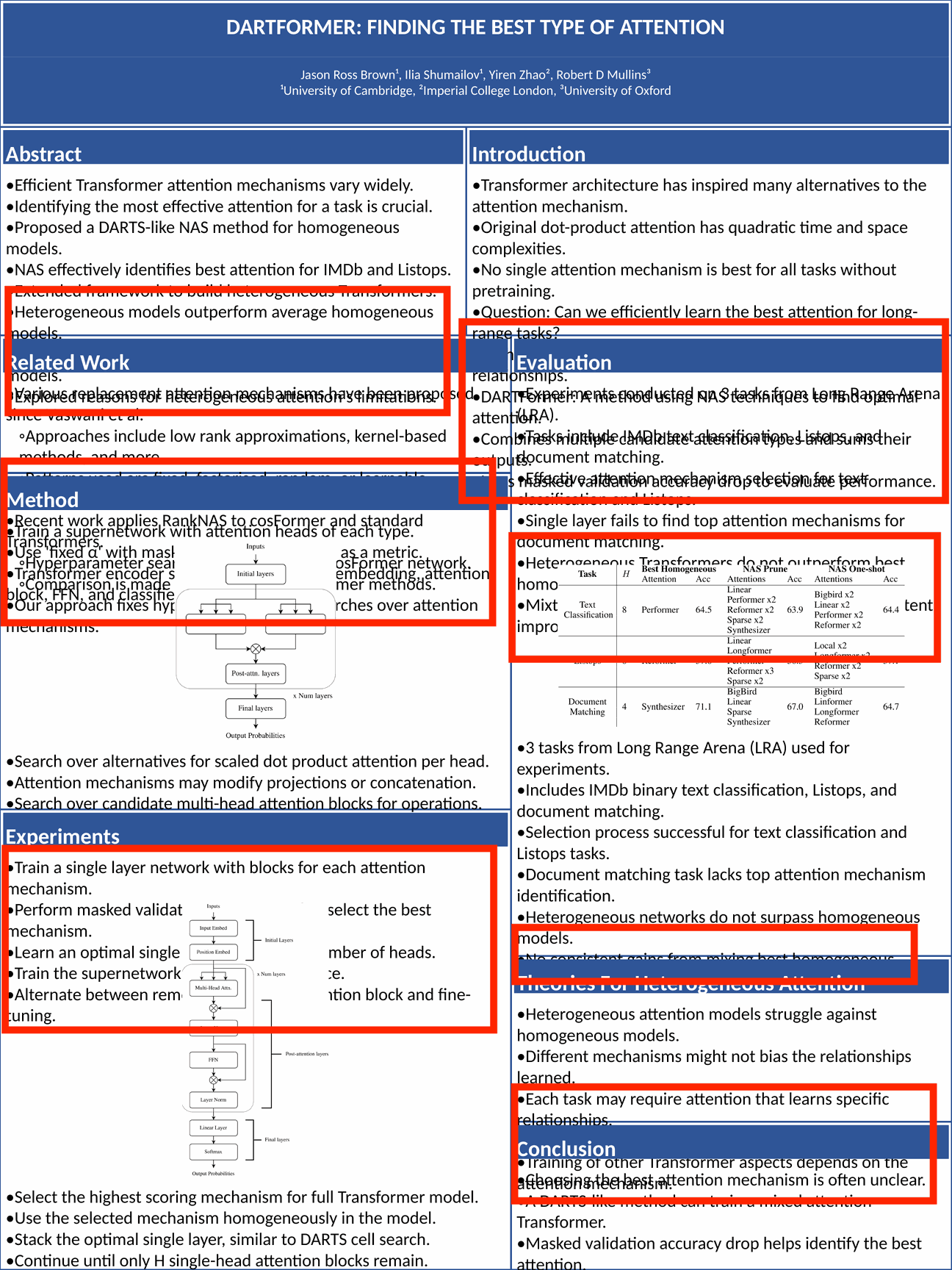}
    \caption{Tree.}
    \label{fig:ablation3_b}
  \end{subfigure}

  \vspace{1em}

  \begin{subfigure}[b]{0.45\textwidth}
    \includegraphics[width=\linewidth]{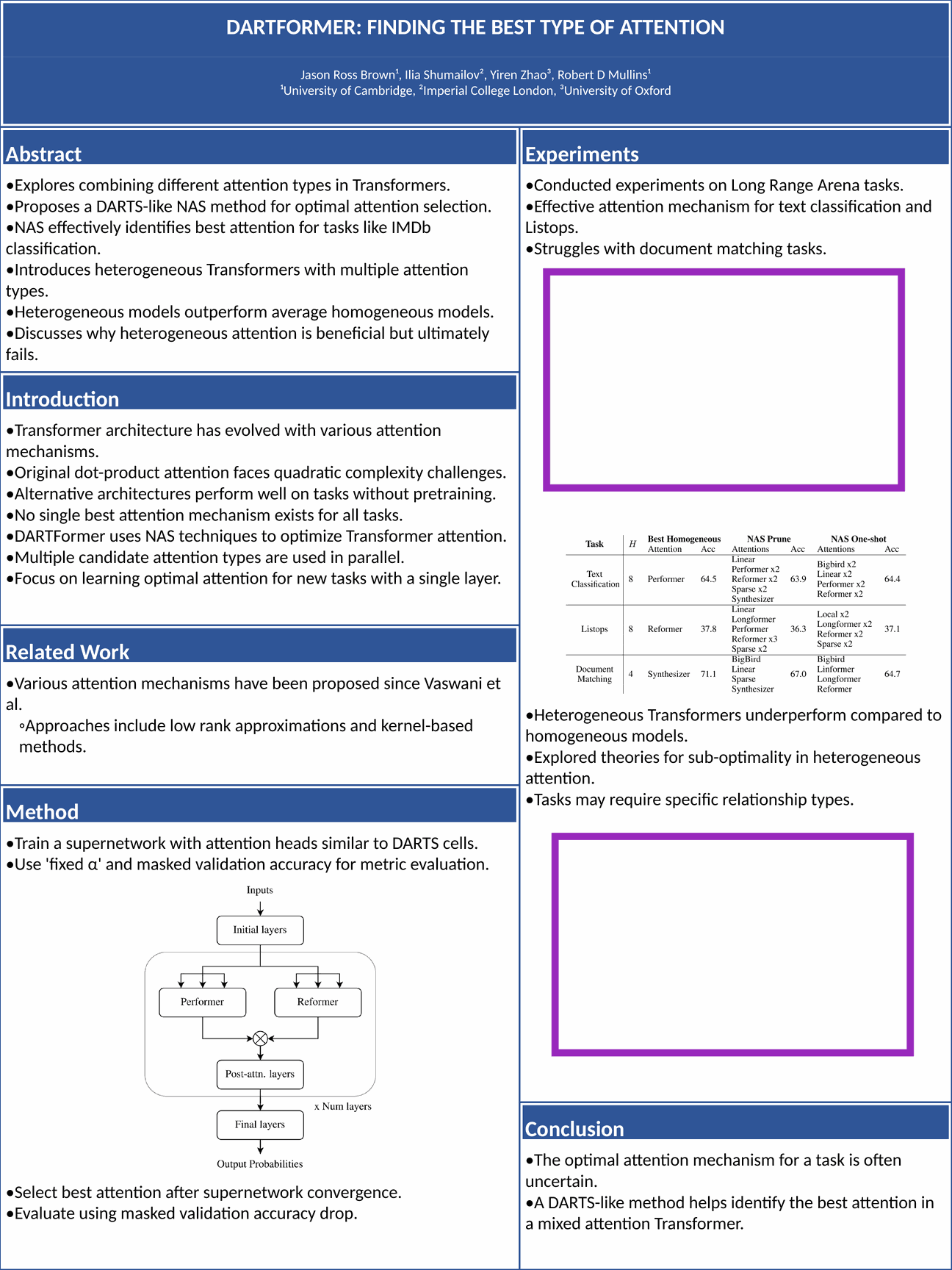}
    \caption{Tree + Commenter.}
    \label{fig:ablation3_c}
  \end{subfigure}
  \hfill
  \begin{subfigure}[b]{0.45\textwidth}
    \includegraphics[width=\linewidth]{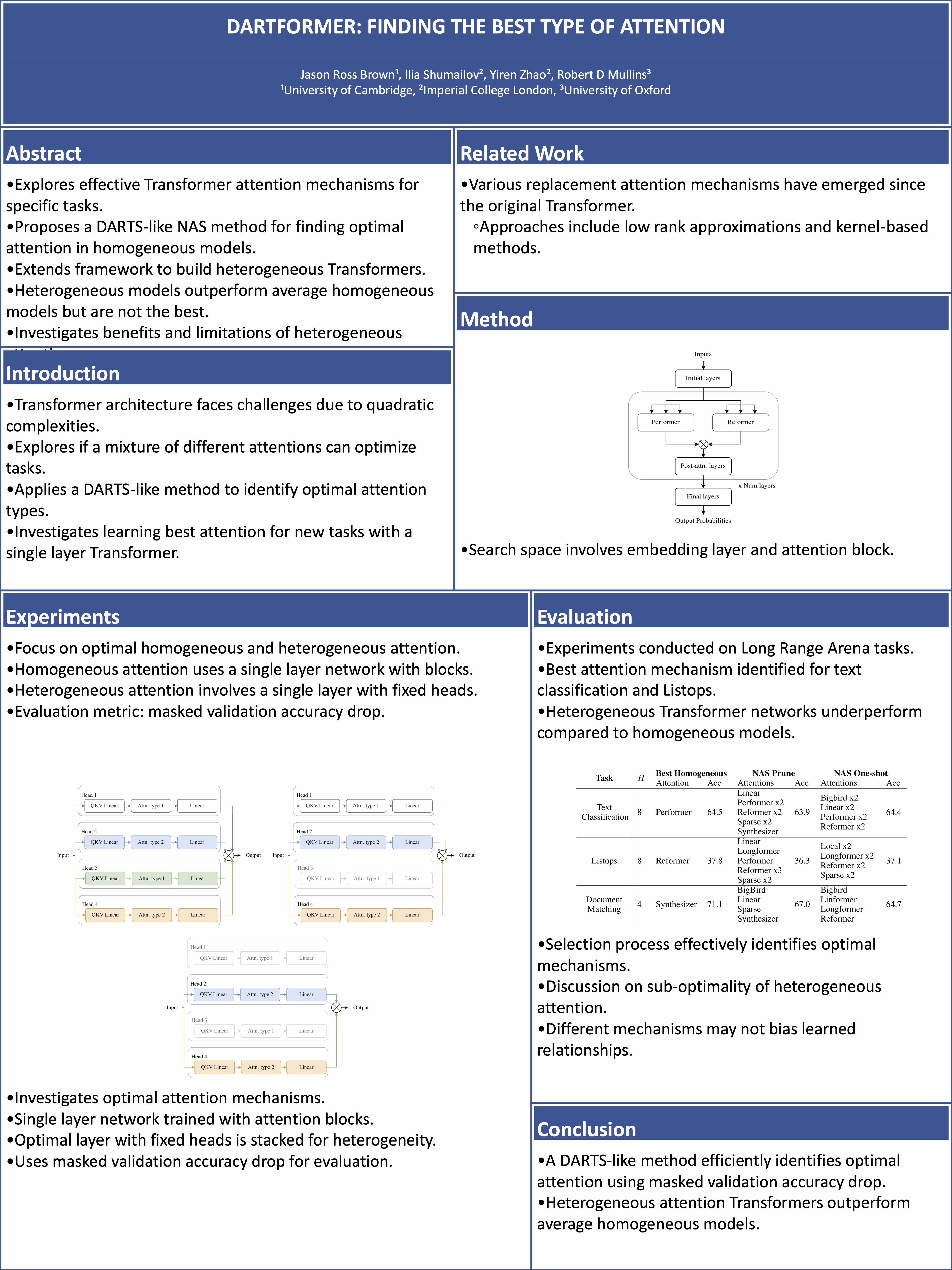}
    \caption{Tree + Commenter + IC.}
    \label{fig:ablation3_d}
  \end{subfigure}

  \caption{Ablation study on \href{https://arxiv.org/pdf/2210.00641}{DARTFormer: Finding The Best Type Of Attention}. Text overflow areas are highlighted with \textcolor{red}{red} bounding boxes, large blank regions are highlighted with \textcolor{purple}{purple} bounding boxes.}
  \label{fig:ablation3}
\end{figure}

\begin{figure}[htbp]
  \centering
  \begin{subfigure}[b]{0.45\textwidth}
    \includegraphics[width=\linewidth]{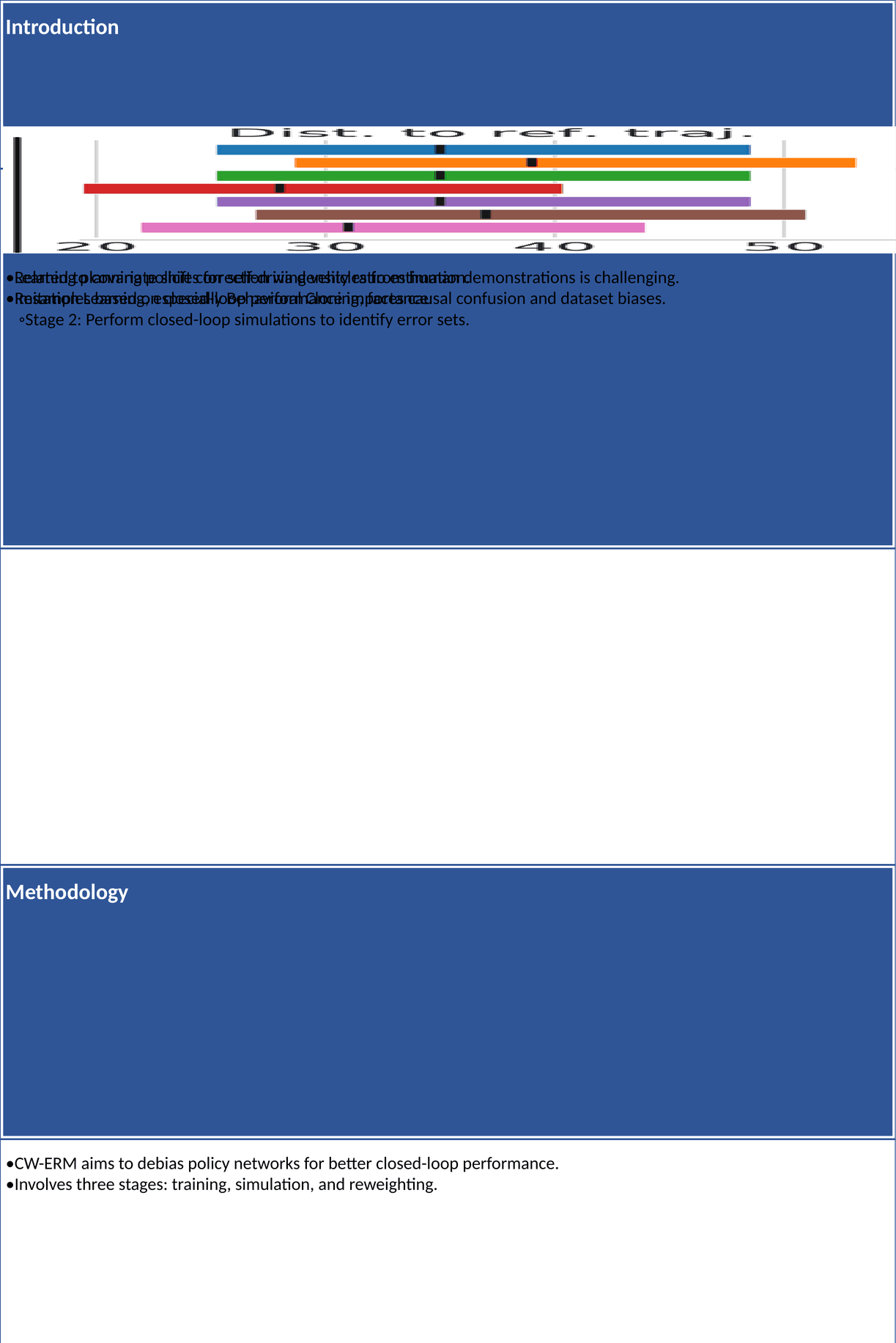}
    \caption{Direct.}
    \label{fig:ablation4_a}
  \end{subfigure}
  \hfill
  \begin{subfigure}[b]{0.45\textwidth}
    \includegraphics[width=\linewidth]{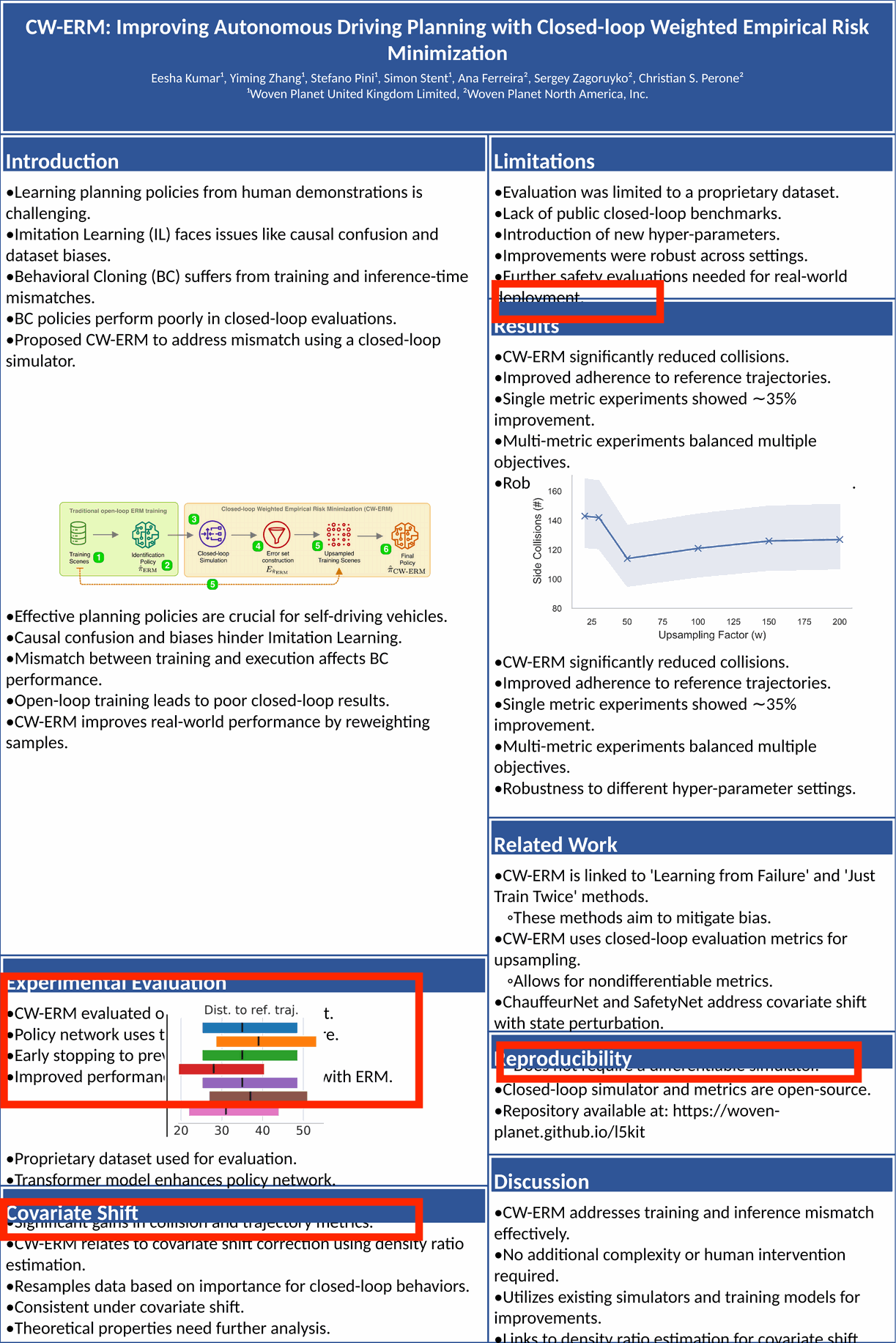}
    \caption{Tree.}
    \label{fig:ablation4_b}
  \end{subfigure}

  \vspace{1em}

  \begin{subfigure}[b]{0.45\textwidth}
    \includegraphics[width=\linewidth]{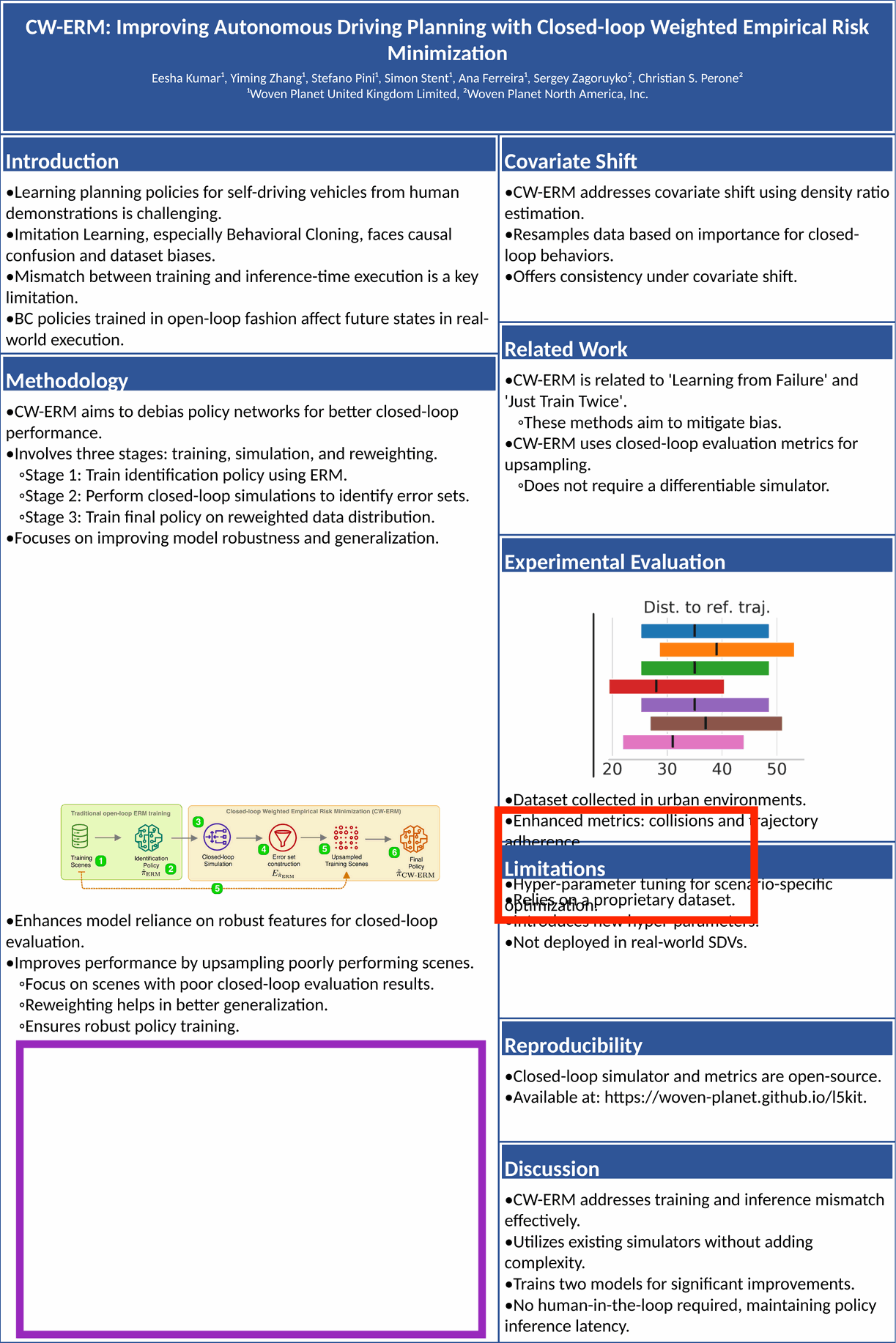}
    \caption{Tree + Commenter.}
    \label{fig:ablation4_c}
  \end{subfigure}
  \hfill
  \begin{subfigure}[b]{0.45\textwidth}
    \includegraphics[width=\linewidth]{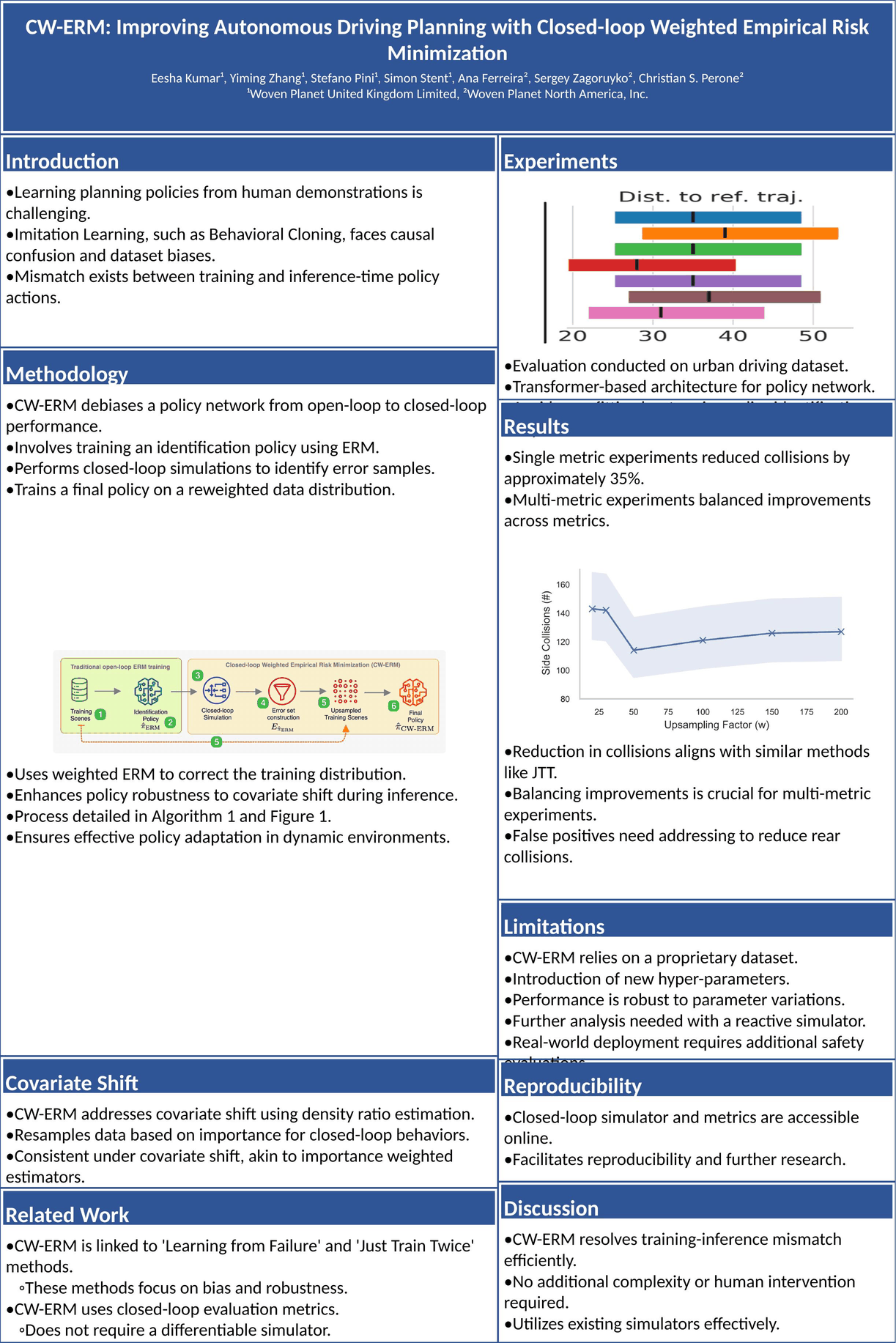}
    \caption{Tree + Commenter + IC.}
    \label{fig:ablation4_d}
  \end{subfigure}

  \caption{Ablation study on \href{https://arxiv.org/pdf/2210.02174}{CW-ERM: Improving Autonomous Driving Planning with Closed-loop Weighted Empirical Risk Minimization}. Text overflow areas are highlighted with \textcolor{red}{red} bounding boxes, and large blank regions are highlighted with \textcolor{purple}{purple} bounding boxes.}
  \label{fig:ablation4}
\end{figure}

\begin{figure}[htbp]
  \centering
  \begin{subfigure}[b]{0.45\textwidth}
    \includegraphics[width=\linewidth]{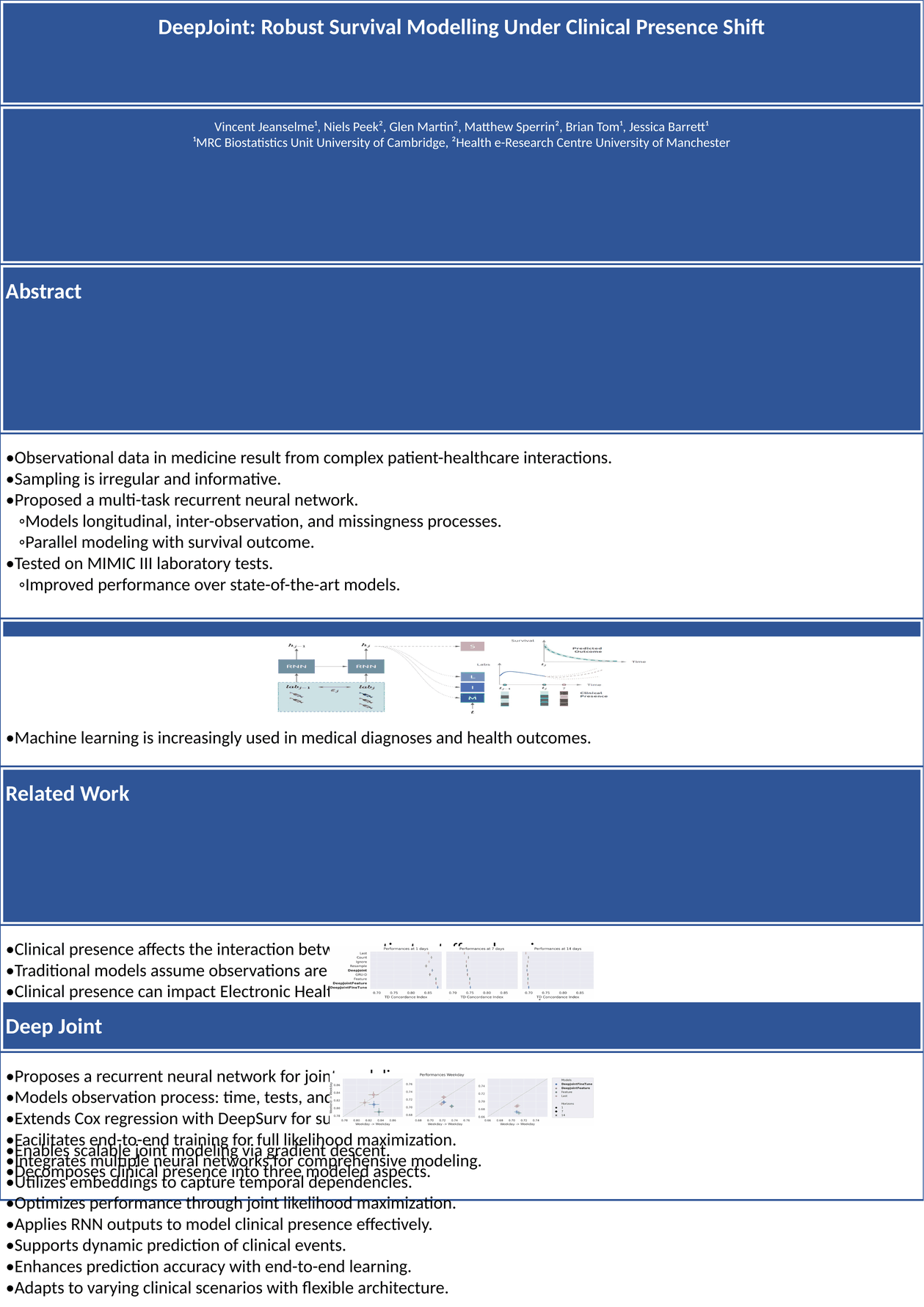}
    \caption{Direct.}
    \label{fig:ablation5_a}
  \end{subfigure}
  \hfill
  \begin{subfigure}[b]{0.45\textwidth}
    \includegraphics[width=\linewidth]{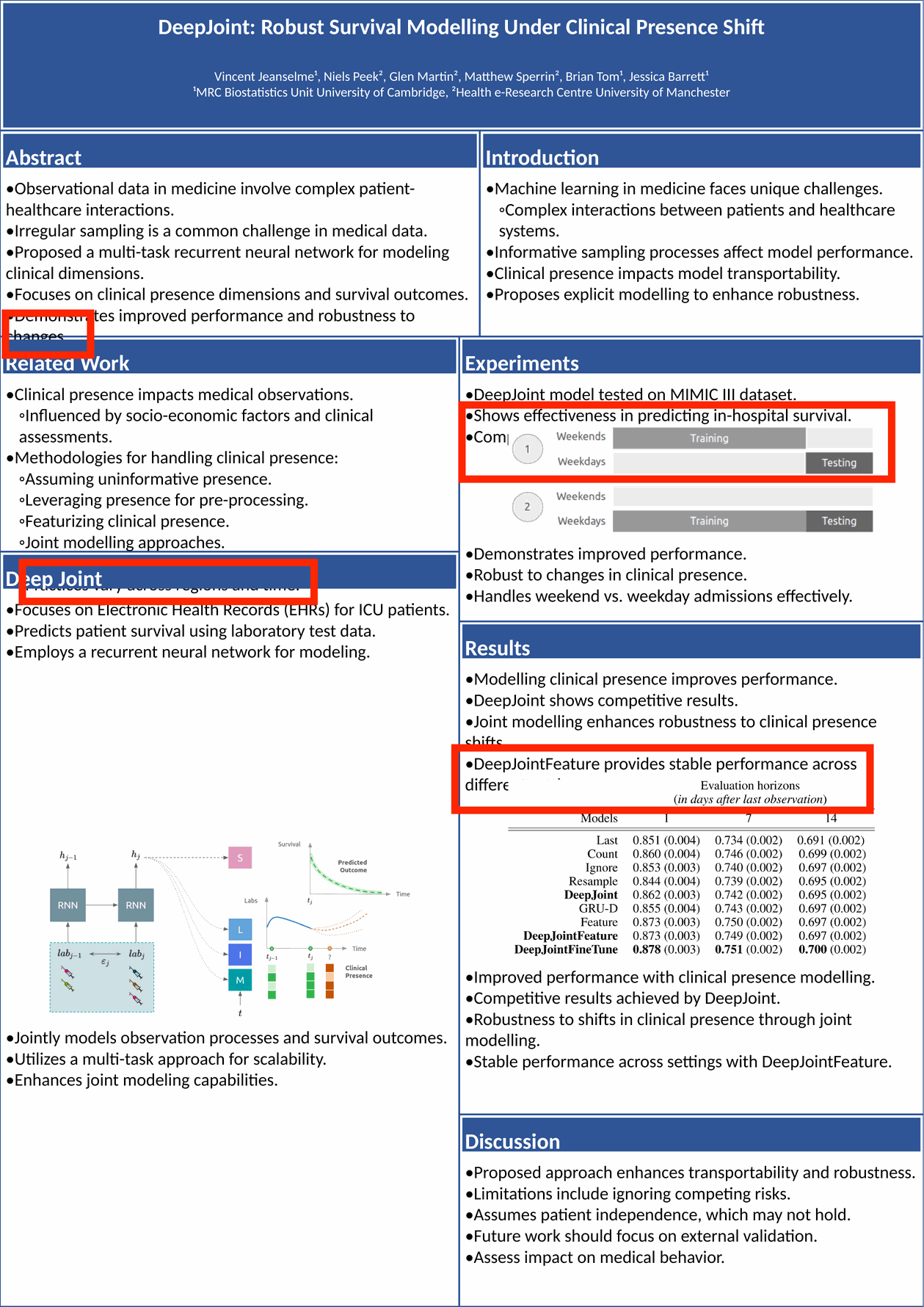}
    \caption{Tree.}
    \label{fig:ablation5_b}
  \end{subfigure}

  \vspace{1em}

  \begin{subfigure}[b]{0.45\textwidth}
    \includegraphics[width=\linewidth]{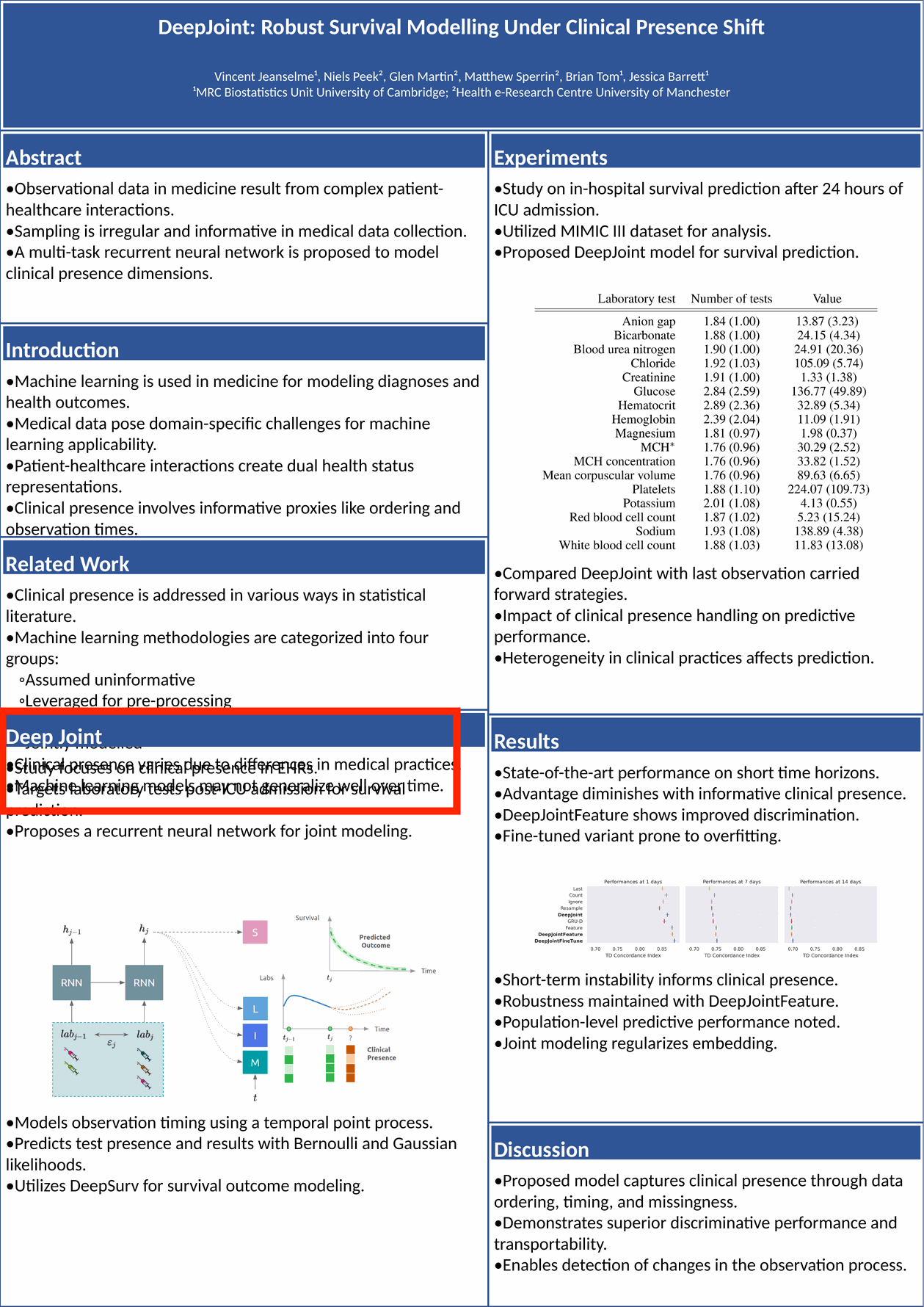}
    \caption{Tree + Commenter.}
    \label{fig:ablation5_c}
  \end{subfigure}
  \hfill
  \begin{subfigure}[b]{0.45\textwidth}
    \includegraphics[width=\linewidth]{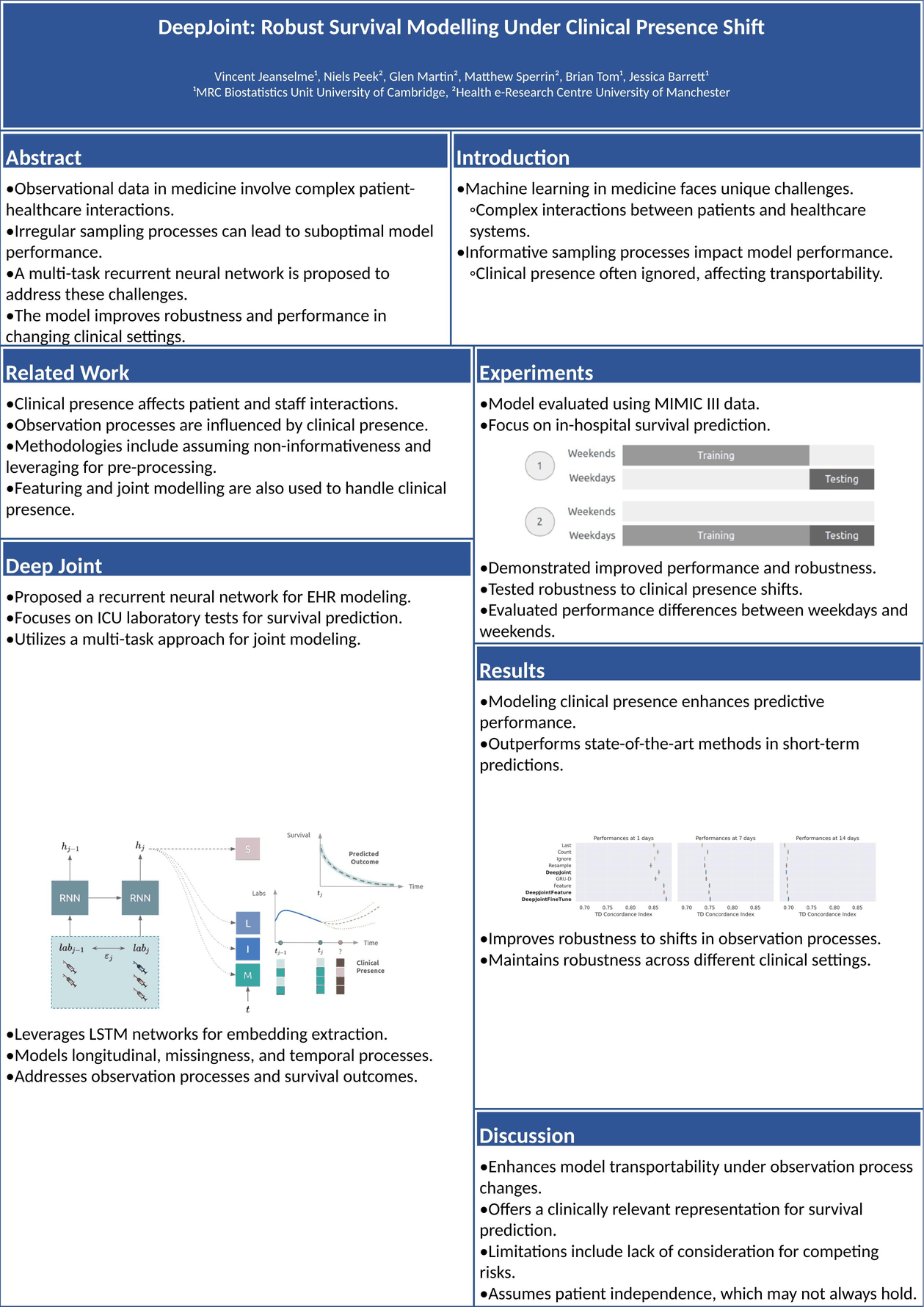}
    \caption{Tree + Commenter + IC.}
    \label{fig:ablation5_d}
  \end{subfigure}

  \caption{Ablation study on \href{https://arxiv.org/pdf/2205.13481}{DeepJoint: Robust Survival Modelling Under Clinical Presence Shift}. Text overflow areas are highlighted with \textcolor{red}{red} bounding boxes.}
  \label{fig:ablation5}
\end{figure}

All ablation variants are implemented using \texttt{\ouralg{}-4o}, keeping all other components unchanged to isolate the effect of each factor. We visualize and compare results across five randomly selected papers from \ourbench{}, as shown in Figures~\ref{fig:ablation1} to~\ref{fig:ablation5}.

When prompting the LLM to directly generate poster layouts (\textit{Direct}), the results are often structurally compromised (e.g., Figures~\ref{fig:ablation1_a}–\ref{fig:ablation3_a}), or resemble blog-style layouts that lack visual hierarchy and appeal (Figures~\ref{fig:ablation4_a},\ref{fig:ablation5_a}). Fine-grained layout components, such as text boxes and figures, are especially challenging to synthesize in this setting: for instance, Figures\ref{fig:ablation1_a}–\ref{fig:ablation4_a} exhibit missing text boxes that leave noticeable blank areas, and Figure~\ref{fig:ablation4_a} fails to preserve the correct aspect ratio of figures.

The \textit{Tree} variant, which omits the commenter module, leads to severe layout defects across all test cases (Figures~\ref{fig:ablation1_b}–\ref{fig:ablation5_b}), primarily manifesting as text overflow—where content spills outside its designated textbox or section panel—resulting in overlaps with other text or visual elements.

Using \textit{Tree + Commenter}, which includes the commenter but without in-context examples, yields improved results compared to the variant without the commenter, but still exhibits noticeable issues. As shown in Figures~\ref{fig:ablation1_c},\ref{fig:ablation2_c},\ref{fig:ablation4_c}, and~\ref{fig:ablation5_c}, some degree of text overflow remains. Furthermore, Figures~\ref{fig:ablation3_c} and~\ref{fig:ablation4_c} highlight substantial unused white space that the commenter fails to flag in the absence of in-context guidance.

Finally, the full \textit{Tree+Commenter+IC} system achieves the best results, as detailed throughout the main paper and demonstrated in Fig. \ref{fig:ablation1_d},\ref{fig:ablation2_d},\ref{fig:ablation3_d},\ref{fig:ablation4_d},\ref{fig:ablation5_d}.

\FloatBarrier

\section{Abbreviations}
We provide a reference for the abbreviations of models used in this paper in Tab.~\ref{tab:abbrev_fullname}.

\begin{table}[!h]
  \centering
  \begin{tabular}{ll}
    \toprule
    \textbf{Abbreviation} & \textbf{Full Name} \\
    \midrule
    llava-ov-7b     & \texttt{LLaVA-OneVision-Qwen2-7b-ov-hf}~\citep{li2024llava} \\
phi4    & \texttt{Phi-4-multimodal-instruct}~\citep{abouelenin2025phi} \\
    gemini-2.0    & \texttt{Gemini-2.0-Flash} \\
    llama4-17b    & 
\texttt{Llama-4-Scout-17B-16E-Instruct} \\
4o-mini & \texttt{GPT-4o-mini} \\

    \bottomrule
  \end{tabular}
  \caption{List of abbreviations and their full names.}
  \label{tab:abbrev_fullname}
\end{table}

\section{More Analysis}
\subsection{Efficiency Analysis}\label{app:efficiency_analysis} In Tab.~\ref{tab:efficiency}, we evaluate the efficiency of \texttt{\ouralg}\ against both direct generation and multi-agent baselines. While \texttt{4o-Image} achieves the highest efficiency by avoiding multi-turn reasoning, it lacks layout-awareness. \ouralg-Qwen-2.5-7B strikes a strong balance, significantly reducing token usage and runtime (47.6K, 192.0s) compared to PPTAgent (255.7K, 230.7s), while maintaining output quality. This highlights the challenge, as well as the efficiency issue
 of \ourbench.

\begin{table*}[t]
  \centering
  \setlength{\tabcolsep}{1.8pt}
  \resizebox{\textwidth}{!}{%
  \begin{tabular}{l c c c c c c c c c c c}
    \toprule
    \textbf{Model}
      & \textbf{in\_t ($K$)~$\downarrow$} 
      & \textbf{out\_t ($K$)~$\downarrow$} 
      & \textbf{in\_v ($K$)~$\downarrow$} 
      & \textbf{out\_v ($K$)~$\downarrow$} 
      & \textbf{total\_t ($K$)~$\downarrow$} 
      & \textbf{total\_v ($K$)~$\downarrow$} 
      & \textbf{Input Tokens ($K$)~$\downarrow$}
      & \textbf{Output Tokens ($K$)~$\downarrow$}
      & \textbf{Total Tokens ($K$)~$\downarrow$}
      & \textbf{Time (s)~$\downarrow$} 
      & \textbf{Cost (\$)~$\downarrow$}\\[0.5ex]
    \midrule

    \textit{{End-to-end methods}} \\

    \rowcolor{direct_gen!50}
    \texttt{4o-HTML}             
      & \textbf{18.53} & \textbf{2.15} & \textbf{0} & \textbf{0} & \textbf{20.67} & \textbf{0}
      & \textbf{18.53} & \textbf{2.15} & \textbf{20.67} & \textbf{62.26} & \underline{0.14} \\[1ex]

    \midrule
    \textit{{Multi-Agent methods}} \\

    \rowcolor{multi_agent!50}
    \texttt{OWL-4o}                 
      & 356.48 & 4.62 & \textbf{0} & \textbf{0} & 361.00 & \textbf{0}
      & 356.48 & 4.62 & 361.10 & 124.29 & 1.87 \\[1ex]

    \rowcolor{multi_agent!50}
    \texttt{PPTAgent-4o}            
      & 202.46 & 33.42 & 18.98 & 0.87 & 235.88 & 19.85
      & 221.43 & 34.29 & 255.73 & 230.70 & 1.79 \\[1ex]

    \midrule
    \textit{{\ouralg\ variants}} \\[0.3ex]

    \rowcolor{poster_agent!50}
    \texttt{\ouralg-4o}            
      & \underline{28.85} & \underline{2.95} & 69.25 & \underline{0.05} & \underline{31.80} & 69.30
      & \underline{98.10} & \underline{3.00} & \underline{101.10} & 281.55  & 0.55 \\[1ex]

    \rowcolor{poster_agent!50}
    \texttt{\ouralg-Qwen}    
      & 29.22 & 3.56 & \underline{14.75} & \underline{0.02} & 32.78 & \underline{14.78}
      & 43.97 & 3.58 & 47.55 & \underline{124.29} & \textbf{0.0045} \\

    \bottomrule
  \end{tabular}%
  }
  \caption{\textbf{Efficiency Analysis} in terms of text and vision tokens, and computation times. Prices of GPT-4o are based on OpenAI’s GPT-4o API pricing as of May 22, 2025 (\$5 / MTok for input and \$20 / MTok for output). Prices of Qwen-2.5-7B (\$0.04 / MTok input and \$0.1 / MTok for output) and Qwen-2.5-VL-7B (\$0.2 / MTok for both) are based on the ones offered by OpenRouter on May 26, 2025.  Best scores in each column are \textbf{bolded} and second best are \underline{underlined}.}
  \label{tab:efficiency}
  \vspace{-10pt}
\end{table*}

\subsubsection{Runtime Analysis and Parallelization}\label{app:runtime_analysis}

While \texttt{\ouralg{}-4o} achieves superior quality and token efficiency compared to baselines, its sequential panel-by-panel content generation initially resulted in longer runtime (281.48s on average) compared to \texttt{OWL-4o} (158.97s). To address this efficiency bottleneck, we implemented a parallelized version that generates content for all panels simultaneously, as panels are independent and can be processed concurrently.

Table~\ref{tab:runtime_breakdown} provides a fine-grained breakdown of runtime across six major procedures: (i) PDF parsing, (ii) figure filtering, (iii) outline generation, (iv) layout generation, (v) content generation (panel iteration), and (vi) rendering. The analysis reveals two primary bottlenecks in the original sequential implementation: PDF parsing (81.08s) and content generation (176.69s).

While PDF parsing relies on established off-the-shelf parsers (Docling and Marker) with limited room for optimization, content generation offers significant parallelization opportunities. Our parallelized implementation reduces content generation time from 176.69s to 54.16s—a 69.3\% reduction—bringing the overall runtime to 166.80s. This represents a \textbf{40.7\% improvement} over the sequential version and makes \texttt{\ouralg{}-4o-Parallel} highly competitive with \texttt{OWL-4o} (166.80s vs. 158.97s), while maintaining superior output quality across all metrics. The small increase in other procedures (Parser, Filter, Outline, Layout) is due to measurement variance and system load, as these steps remain unchanged between versions.

\begin{table*}[t]
  \centering
  \setlength{\tabcolsep}{3pt}
  \resizebox{0.95\textwidth}{!}{%
  \begin{tabular}{l c c c c c c c}
    \toprule
    \textbf{Model}
      & \textbf{(i) Parser (s)~$\downarrow$}
      & \textbf{(ii) Filter (s)~$\downarrow$}
      & \textbf{(iii) Outline (s)~$\downarrow$}
      & \textbf{(iv) Layout (s)~$\downarrow$}
      & \textbf{(v) Content (s)~$\downarrow$}
      & \textbf{(vi) Render (s)~$\downarrow$}
      & \textbf{Total (s)~$\downarrow$} \\[0.5ex]
    \midrule
    \rowcolor{multi_agent!50}
    \texttt{OWL-4o} (reference)
      & \multicolumn{6}{c}{\cellcolor{multi_agent}\textit{(no fine-grained breakdown available)}}
      & \cellcolor{multi_agent}158.97 \\[1ex]

    \rowcolor{poster_agent!50}
    \texttt{\ouralg-4o} (sequential)
      & 81.08 & 17.42 & 3.47 & 0.15 & 176.69 & 2.67
      & \cellcolor{poster_agent}281.48 \\[1ex]

    \rowcolor{poster_agent!50}
    \texttt{\ouralg-4o-Parallel}
      & 87.45 & 18.29 & 4.09 & 0.17 & 54.16 & 2.65
      & \cellcolor{poster_agent}166.80 \textbf{($\downarrow$40.7\%)} \\

    \bottomrule
  \end{tabular}%
  }
  \caption{\textbf{Fine-grained runtime breakdown across six major procedures.} Results are averaged over a random subset of 10 papers. The parallelized implementation achieves a 40.7\% reduction in total runtime by concurrently generating content for all panels.}
  \label{tab:runtime_breakdown}
\end{table*}

\subsection{Cost Analysis}\label{app:cost_analysis}

Token consumptions are depicted in Figure \ref{fig:token_consumption} and Table \ref{tab:efficiency}. Using \texttt{GPT-4o} as the backbone for both the LLM and VLM components, the average cost of generating a single paper with \texttt{\ouralg{}-4o} is approximately:
\[
\frac{98.1 \times 1000}{1,000,000} \times 5 + \frac{3 \times 1000}{1,000,000} \times 20 = 0.55\ \text{USD},
\]
based on OpenAI’s GPT-4o API pricing as of May 22, 2025.

Using \texttt{Qwen-2.5-7B} as the backbone for LLM and \texttt{Qwen-2.5-VL-7B} as VLM, the average cost of generating a single paper with \texttt{\ouralg{}-4o} is approximately:
\[
\frac{29.22 \times 1000}{1,000,000} \times 0.04 + \frac{3.56 \times 1000}{1,000,000} \times 0.1 + \frac{14.78 \times 1000}{1,000,000} \times 0.2  = 0.0045\ \text{USD},
\]
based on OpenRouter's API pricing as of May 26, 2025.

\subsection{Impact of Backbone Choices}\label{append:cross_backbone}

Table~\ref{tab:poster_agent_variants} compares four \texttt{\ouralg{}} variants obtained by crossing two language models (LMs)—\texttt{GPT-4o} and \texttt{Qwen-2.5-7B}—with the same two models used as vision--language backbones (VLMs).

\textbf{Overall robustness.} All configurations perform similarly. The \kqa{} metric spans only $114.09$ (\texttt{Qwen--4o}) to $118.25$ (\texttt{4o--Qwen}), a spread approximately $4$, indicating that \texttt{\ouralg{}} is largely insensitive to the specific backbone combination.

\textbf{Open-source competitiveness.} The fully open-source stack (\texttt{Qwen--Qwen}) achieves a \kqa{} score of $114.65$, trailing the best closed-source variant by merely $3.6$. Strong performance is therefore attainable without proprietary APIs.

\textbf{Stable multimodal quality.} Visual similarity and figure relevance vary by less than $0.01$ across variants, underscoring the stability of our multimodal generation pipeline.

\textbf{LLM vs. VLM trade-off.} Holding the LLM fixed, substituting \texttt{Qwen} for the VLM consistently improves \kqa{} (\texttt{4o-Qwen}: $+2.1$ over \texttt{4o-4o}; \texttt{Qwen-Qwen}: $+0.56$ over \texttt{Qwen-4o}). We attribute this to \texttt{GPT-4o} acting as a stricter layout critic, trimming overflow text and modestly reducing information volume. Conversely, the stricter VLM raises aesthetic scores, yielding higher VLM-as-judge ratings (\texttt{4o-4o}: $3.72$ vs. \texttt{4o-Qwen}: $3.58$). The \texttt{4o-4o} configuration thus offers the best balance between informativeness and visual appeal.

\begin{table*}[t]
  \centering
  \resizebox{\textwidth}{!}{%
  \begin{tabular}{ll *{9}{c}}
    \toprule
    \multirow{3}{*}{\textbf{LLM}}
    & \multirow{3}{*}{\textbf{VLM}}
    & \multicolumn{3}{c}{\textbf{Vis. quality \& Txt. coherence}}
    & \multicolumn{3}{c}{\textbf{VLM-as-Judge}}
    & \multicolumn{3}{c}{\textbf{Density-augmented Score}} \\[1ex]
    & & {Visual Similarity} & {PPL} & {Figure Relevance}
      & {Aesthetic} & {Information} & {Overall}
      & {V-Avg} & {I-Avg} & {Overall} \\[0.5ex]
    \midrule
    \texttt{GPT-4o}      & \texttt{GPT-4o}
      & \underline{0.75} & \textbf{8.31}         & \underline{0.24}
      & \textbf{3.58}    & \textbf{3.86} & \textbf{3.72}
      & \underline{101.87} & \underline{130.39} & \underline{116.13} \\[1ex]
    \texttt{GPT-4o}      & \texttt{Qwen-2.5-7B}
      & \underline{0.75} & 9.25 & \underline{0.24}
      & 3.33             & 3.82          & 3.58
      & \textbf{105.61}    & \textbf{130.88}    & \textbf{118.25}    \\[1ex]
    \texttt{Qwen-2.5-7B} & \texttt{GPT-4o}
      & \textbf{0.76}    & 9.12 & \textbf{0.25}
      & \underline{3.57} & 3.82          & \underline{3.70}
      & 100.09            & 128.09             & 114.09             \\[1ex]
    \texttt{Qwen-2.5-7B} & \texttt{Qwen-2.5-7B}
      & \underline{0.75} & \underline{8.81}         & \underline{0.24}
      & 3.50             & \underline{3.83} & 3.66
      & 100.35            & 128.94             & 114.65             \\
    \bottomrule
  \end{tabular}%
  }
  \caption{\textbf{Ablation studies of our \ouralg\ variants}. Best scores in each column are \textbf{bolded} and second best are \underline{underlined}.}
  \label{tab:poster_agent_variants}
\end{table*}

\subsection{Additional Backbone Evaluations}\label{append:additional_backbones}

To further evaluate the generalizability of \texttt{\ouralg{}}, we conducted experiments with two additional backbones: \texttt{o4-mini} and \texttt{Qwen-2.5-72B}.

Table~\ref{tab:additional_backbones_full} presents the complete evaluation across all metrics for the new backbones, alongside the original \texttt{4o-Image} baseline for reference. Table~\ref{tab:additional_backbones_paperquiz} provides detailed \kqa{} scores broken down by question type (Verbatim vs. Interpretive) and reader model categories (open-source vs. closed-source). Finally, Table~\ref{tab:additional_backbones_summary} offers a concise comparison of the key metrics.

\textbf{Key observations:}
\textit{(i)} All \texttt{\ouralg{}} variants substantially outperform the \texttt{4o-Image} baseline across nearly all metrics, with overall VLM-as-Judge scores ranging from 3.56–3.72 vs. 2.33 (+1.23–1.39 absolute, $\sim$53–60\% relative improvement).
\textit{(ii)} Visual similarity remains high and stable (0.75–0.78) across all backbones, with \texttt{PosterAgent-Qwen-72B} achieving the highest score (0.78).
\textit{(iii)} \texttt{PosterAgent-o4-mini} achieves the highest raw \kqa{} overall score (61.33) and augmented score (121.91), demonstrating that reasoning models can produce highly informative posters.
\textit{(iv)} Information scores cluster at 3.83–3.87 (vs. 1.77 for baseline), and Aesthetic scores at 3.26–3.58 (vs. 2.90), indicating backbone-insensitive improvements in both informativeness and visual quality.

These results confirm that \texttt{\ouralg{}}'s multi-agent design generalizes well across different backbone choices, maintaining strong performance with both reasoning closed-source models and larger open-source alternatives.

\begin{table*}[t]
  \centering
  \setlength{\tabcolsep}{1.8pt}
  \resizebox{\textwidth}{!}{%
  \begin{tabular}{l *{3}{c} *{9}{c}}
    \toprule
    \multirow{5}{*}{\textbf{Model}}
      & \multicolumn{3}{c}{\textbf{Vis. quality \& Txt. coherence}}
      & \multicolumn{9}{c}{\textbf{VLM-as-Judge}} \\[1ex]
    \cmidrule(lr){2-4}\cmidrule(lr){5-13}
      & \multirow{3}{*}{Vis. Sim.~$\uparrow$}
      & \multirow{3}{*}{PPL~$\downarrow$}
      & \multirow{3}{*}{Fig. Rel.~$\uparrow$}
      & \multicolumn{4}{c}{\textbf{Aesthetic score}~$\uparrow$}
      & \multicolumn{4}{c}{\textbf{Information score}~$\uparrow$}
      & \multirow{3}{*}{\textbf{Overall}~$\uparrow$} \\[1ex]
    \cmidrule(lr){5-8}\cmidrule(lr){9-12}
      &  &  &
      & Element & Layout & Engage. & \textbf{Avg.}
      & Clarity & Content & Logic & \textbf{Avg.}
      & \\[0.5ex]
    \midrule
    \textit{{Baseline}} \\
    \rowcolor{direct_gen!50}
    \texttt{4o-Image}
      & \textbf{0.76} & 77.13 & 0.21
      & 2.93 & 3.02 & 2.75 & \cellcolor{direct_gen}2.90
      & 1.05 & 2.04 & 2.22 & \cellcolor{direct_gen}1.77
      & \cellcolor{direct_gen}{2.33} \\[1ex]

    \midrule
    \textit{{\ouralg\ with additional backbones}} \\[0.3ex]

    \rowcolor{poster_agent!50}
    \texttt{\ouralg-4o}
      & 0.75 & \textbf{8.31} & \underline{0.24}
      & \textbf{3.95} & \textbf{3.86} & \textbf{2.93} & \cellcolor{poster_agent}\textbf{3.58}
      & \textbf{4.03} & \underline{3.96} & 3.60 & \cellcolor{poster_agent}\underline{3.86}
      & \cellcolor{poster_agent}{\textbf{3.72}} \\[1ex]

    \rowcolor{poster_agent!50}
    \texttt{\ouralg-o4-mini}
      & \textbf{0.76} & 14.00 & 0.23
      & 3.79 & 3.38 & 2.64 & \cellcolor{poster_agent}3.27
      & \underline{3.98} & \textbf{3.98} & 3.64 & \cellcolor{poster_agent}\textbf{3.87}
      & \cellcolor{poster_agent}{3.57} \\[1ex]

    \rowcolor{poster_agent!50}
    \texttt{\ouralg-Qwen-7B}
      & 0.75 & \underline{8.81} & \underline{0.24}
      & \underline{3.93} & \underline{3.67} & \underline{2.89} & \cellcolor{poster_agent}\underline{3.50}
      & 3.95 & 3.85 & \underline{3.68} & \cellcolor{poster_agent}3.83
      & \cellcolor{poster_agent}{\underline{3.66}} \\[1ex]

    \rowcolor{poster_agent!50}
    \texttt{\ouralg-Qwen-72B}
      & \underline{0.78} & \underline{8.81} & \textbf{0.25}
      & 3.76 & 3.39 & 2.63 & \cellcolor{poster_agent}3.26
      & 3.88 & \underline{3.96} & \textbf{3.74} & \cellcolor{poster_agent}\underline{3.86}
      & \cellcolor{poster_agent}{3.56} \\

    \bottomrule
  \end{tabular}%
  }
  \caption{\textbf{Detailed evaluation with additional backbones.} All \texttt{\ouralg{}} variants substantially outperform the \texttt{4o-Image} baseline. Best scores in each column are \textbf{bolded} and second best are \underline{underlined}.}
  \label{tab:additional_backbones_full}
\end{table*}

\begin{table*}[t]
  \centering
  \setlength{\tabcolsep}{1.8pt}
  \resizebox{\textwidth}{!}{%
  \begin{tabular}{l *{7}{c} *{3}{c}}
    \toprule
      & \multicolumn{7}{c}{\textbf{Raw Accuracy}}
      & \multicolumn{3}{c}{\textbf{Density-Augmented Score}} \\[1ex]
    \cmidrule(lr){2-8}\cmidrule(lr){9-11}
    \textbf{Model}
      & \multicolumn{3}{c}{\textbf{Verbatim}~$\uparrow$}
      & \multicolumn{3}{c}{\textbf{Interpretive}~$\uparrow$}
      & \multirow{2}{*}{\textbf{Overall}~$\uparrow$}
      & \multirow{2}{*}{\textbf{V-Avg}~$\uparrow$} & \multirow{2}{*}{\textbf{I-Avg}~$\uparrow$} & \multirow{2}{*}{\textbf{Overall}~$\uparrow$}
      \\[0.5ex]
      & open-source & closed-source & \textbf{V-Avg}
      & open-source & closed-source & \textbf{I-Avg}
      &
      & & &
      \\[0.5ex]
    \midrule
    \textit{Baseline} \\
    \rowcolor{direct_gen!50}
    \texttt{4o-Image}
      & 48.97 & 30.89 & \cellcolor{direct_gen}39.93
      & 50.19 & 70.67 & \cellcolor{direct_gen}60.43
      & \cellcolor{direct_gen}50.18
      & \cellcolor{direct_gen}79.86  & \cellcolor{direct_gen}120.86 & \cellcolor{direct_gen}100.36
      \\[1ex]
    \midrule
    \textit{\ouralg\ with additional backbones} \\[0.3ex]
    \rowcolor{poster_agent!50}
    \texttt{\ouralg-4o}
      & 52.95 & 49.17                & \cellcolor{poster_agent}51.06
      & 52.29 & \underline{78.42}             & \cellcolor{poster_agent}65.35
      & \cellcolor{poster_agent}58.21
      & \cellcolor{poster_agent}101.87 & \cellcolor{poster_agent}\underline{130.39} & \cellcolor{poster_agent}116.13
      \\[1ex]
    \rowcolor{poster_agent!50}
    \texttt{\ouralg-o4-mini}
      & \textbf{54.21} & \textbf{60.27} & \cellcolor{poster_agent}\textbf{57.24}
      & 51.99 & \textbf{78.87} & \cellcolor{poster_agent}\underline{65.43}
      & \cellcolor{poster_agent}\textbf{61.33}
      & \cellcolor{poster_agent}\textbf{113.76} & \cellcolor{poster_agent}130.05 & \cellcolor{poster_agent}\textbf{121.91}
      \\[1ex]
    \rowcolor{poster_agent!50}
    \texttt{\ouralg-Qwen-7B}
      & 51.81 & 48.79                & \cellcolor{poster_agent}50.30
      & \underline{52.57} & 76.66             & \cellcolor{poster_agent}64.62
      & \cellcolor{poster_agent}57.46
      & \cellcolor{poster_agent}100.35 & \cellcolor{poster_agent}128.94 & \cellcolor{poster_agent}114.65
      \\[1ex]
    \rowcolor{poster_agent!50}
    \texttt{\ouralg-Qwen-72B}
      & \underline{53.65} & \underline{54.61} & \cellcolor{poster_agent}\underline{54.13}
      & \textbf{52.69} & 78.01 & \cellcolor{poster_agent}\textbf{65.35}
      & \cellcolor{poster_agent}\underline{59.74}
      & \cellcolor{poster_agent}\underline{107.76} & \cellcolor{poster_agent}\textbf{130.10} & \cellcolor{poster_agent}\underline{118.93}
      \\
    \bottomrule
  \end{tabular}%
  }
  \caption{\textbf{\kqa\ evaluation with additional backbones.} \texttt{PosterAgent-o4-mini} achieves the highest overall scores, while all variants substantially outperform the baseline. Best scores in each column are \textbf{bolded} and second best are \underline{underlined}.}
  \label{tab:additional_backbones_paperquiz}
\end{table*}

\begin{table*}[t]
  \centering
  \setlength{\tabcolsep}{3pt}
  \resizebox{0.85\textwidth}{!}{%
  \begin{tabular}{l c c c c}
    \toprule
    \textbf{Model}
      & \textbf{Visual Sim}~$\uparrow$
      & \textbf{Overall VLM-as-Judge}~$\uparrow$
      & \textbf{PaperQuiz Raw Overall}~$\uparrow$
      & \textbf{PaperQuiz Aug Overall}~$\uparrow$ \\[0.5ex]
    \midrule
    \textit{Baseline} \\
    \rowcolor{direct_gen!50}
    \texttt{4o-Image}
      & \underline{0.76} & \cellcolor{direct_gen}2.33 & \cellcolor{direct_gen}50.18 & \cellcolor{direct_gen}100.36 \\[1ex]

    \midrule
    \textit{\ouralg\ with additional backbones} \\[0.3ex]
    \rowcolor{poster_agent!50}
    \texttt{\ouralg-4o}
      & 0.75 & \cellcolor{poster_agent}\textbf{3.72} & \cellcolor{poster_agent}58.21 & \cellcolor{poster_agent}116.13 \\[1ex]
    \rowcolor{poster_agent!50}
    \texttt{\ouralg-o4-mini}
      & \underline{0.76} & \cellcolor{poster_agent}3.57 & \cellcolor{poster_agent}\textbf{61.33} & \cellcolor{poster_agent}\textbf{121.91} \\[1ex]
    \rowcolor{poster_agent!50}
    \texttt{\ouralg-Qwen-7B}
      & 0.75 & \cellcolor{poster_agent}\underline{3.66} & \cellcolor{poster_agent}57.46 & \cellcolor{poster_agent}114.65 \\[1ex]
    \rowcolor{poster_agent!50}
    \texttt{\ouralg-Qwen-72B}
      & \textbf{0.78} & \cellcolor{poster_agent}3.56 & \cellcolor{poster_agent}\underline{59.74} & \cellcolor{poster_agent}\underline{118.93} \\
    \bottomrule
  \end{tabular}%
  }
  \caption{\textbf{Summary comparison of key metrics with additional backbones.} All \texttt{\ouralg{}} variants demonstrate strong performance across metrics. Best scores in each column are \textbf{bolded} and second best are \underline{underlined}.}
  \label{tab:additional_backbones_summary}
\end{table*}

\subsection{Poster Generation Paradigm Comparison}\label{append:paradigm_comparison}

To clarify our design choices, we provide a systematic comparison of different poster generation paradigms in Table~\ref{tab:paradigm_comparison}. \texttt{\ouralg{}} adopts a \textit{hybrid} approach that combines the strengths of multiple paradigms: we generate code for precise layout control (coordinates, sizes, layering), then render to PPTX to obtain visual feedback, which the system uses to iteratively refine both layout and content.

\textbf{Coding-only approaches} (e.g., direct HTML/code synthesis or general coding agents) offer exact placement and reproducibility but produce artifacts that are cumbersome for users to edit and cannot naturally "see" the rendered result to correct visual issues like text overflow or alignment problems.

\textbf{GUI-only pipelines} make editing easy and support feedback from the rendered poster. Still, precise, large-scale adjustments require many low-level operations (e.g., clicking, dragging) and are computationally inefficient for automated generation.

\textbf{Template retrieval} can be efficient and produce editable outputs, but it is not true generation from scratch and depends critically on the availability and suitability of templates. For scientific posters, high-quality, non-proprietary, and diverse templates are scarce. Even when strong templates are available, our experiments show that \texttt{PPTAgent-4o}—given six human-designed poster templates with manual selection of the best match—performed noticeably worse than \texttt{\ouralg{}}, underscoring the limitation of template dependence for this task.

By generating code and iterating with rendered visual feedback in PPTX, \texttt{\ouralg{}} inherits precise control, editable outputs, true from-scratch generation, efficient global changes, and feedback-driven refinement—properties we found necessary to meet the dual demands of content accuracy and visual layout quality.

\begin{table*}[t]
  \centering
  \setlength{\tabcolsep}{4pt}
  \resizebox{0.95\textwidth}{!}{%
  \begin{tabular}{l c c c c c}
    \toprule
    \textbf{Paradigm}
      & \textbf{Precise Control}
      & \textbf{Easy User Editing}
      & \textbf{Generate from Scratch}
      & \textbf{Uses Visual Feedback}
      & \textbf{Efficient Generation} \\[0.5ex]
    \midrule
    \rowcolor{multi_agent!20}
    Coding-only
      & \ding{51} & \ding{55} & \ding{51} & \ding{55} & \ding{51} \\[-0.5ex]
    \rowcolor{multi_agent!20}
    \small (e.g., HTML synthesis)
      & \small (exact placement) & \small (code is hard to edit visually) & & & \\[1ex]

    \rowcolor{direct_gen!20}
    GUI-only
      & \ding{55} & \ding{51} & \ding{51} & \ding{51} & \ding{55} \\[-0.5ex]
    \rowcolor{direct_gen!20}
    \small (e.g., UI automation)
      & \small (fine placement is difficult) & & & & \small (many fine-grained actions) \\[1ex]

    \rowcolor{gt_models!50}
    Template retrieval
      & \ding{55} & \ding{51} & \ding{55} & \ding{51} & \ding{51} \\[-0.5ex]
    \rowcolor{gt_models!50}
    \small (e.g., PPTAgent w/ templates)
      & \small (constrained by template) & & \small (depends on template pool) & & \\[1ex]

    \midrule
    \rowcolor{poster_agent!50}
    \textbf{\ouralg{} (hybrid)}
      & \ding{51} & \ding{51} & \ding{51} & \ding{51} & \ding{51} \\[-0.5ex]
    \rowcolor{poster_agent!50}
    \small (code gen + PPTX render)
      & \small (code-based precision) & \small (editable PPTX output) & \small (no template required) & \small (visual-in-the-loop) & \small (parallelizable) \\

    \bottomrule
  \end{tabular}%
  }
  \caption{\textbf{Comparison of poster generation paradigms.} \texttt{\ouralg{}}'s hybrid approach combines code generation (for precise control) with PPTX rendering (for visual feedback and editability), achieving all desired properties: precise control, easy editing, from-scratch generation, visual feedback integration, and efficient generation. \ding{51} indicates the paradigm supports the property well; \ding{55} indicates significant limitations.}
  \label{tab:paradigm_comparison}
\end{table*}

\subsection{VLM-as-Judge Robustness Analysis}\label{append:vlm_judge_robustness}

To verify the stability and reliability of our VLM-as-Judge evaluation, we conducted five independent runs of the complete evaluation on \texttt{\ouralg-4o} across the entire dataset (100 samples).

Table~\ref{tab:vlm_judge_robustness} presents the results across all six fine-grained criteria (three aesthetic and three information dimensions), along with the averaged scores. The results demonstrate exceptional stability: standard deviations are minimal (std $<$ 0.024) across all metrics, with the overall average showing particularly low variance (std = 0.005). The 95\% confidence intervals are extremely narrow.

\textbf{Key observations:}
\textit{(i)} All metrics exhibit high consistency across runs, with a coefficient of variation $<$ 1\% for most measures.
\textit{(ii)} The narrow confidence intervals indicate that a single evaluation run provides reliable estimates for model comparison.
\textit{(iii)} The stability validates our VLM-as-Judge approach as a robust automatic evaluation method for poster generation.

Given the observed stability, we conclude that single-run evaluations are sufficient for practical model comparison, with periodic multi-run audits recommended to verify continued metric stability.

\begin{table*}[t]
  \centering
  \setlength{\tabcolsep}{2pt}
  \resizebox{\textwidth}{!}{%
  \begin{tabular}{l c c c c c c c c c}
    \toprule
    \textbf{Run}
      & \multicolumn{3}{c}{\textbf{Aesthetic}}
      & \multicolumn{3}{c}{\textbf{Information}}
      & \textbf{Aesthetic}
      & \textbf{Information}
      & \textbf{Overall} \\[0.5ex]
    \cmidrule(lr){2-4}\cmidrule(lr){5-7}
      & Element & Layout & Engagement & Clarity & Content & Logic & Avg & Avg & Avg \\[0.5ex]
    \midrule
    \rowcolor{poster_agent!50}
    Run 1 & 3.95 & 3.86 & 2.93 & 4.03 & 3.96 & 3.60 & \cellcolor{poster_agent}3.58 & \cellcolor{poster_agent}3.86 & \cellcolor{poster_agent}3.72 \\[1ex]
    \rowcolor{poster_agent!50}
    Run 2 & 3.95 & 3.90 & 2.96 & 4.02 & 3.96 & 3.59 & \cellcolor{poster_agent}3.60 & \cellcolor{poster_agent}3.86 & \cellcolor{poster_agent}3.73 \\[1ex]
    \rowcolor{poster_agent!50}
    Run 3 & 3.91 & 3.88 & 2.97 & 4.04 & 3.99 & 3.59 & \cellcolor{poster_agent}3.59 & \cellcolor{poster_agent}3.87 & \cellcolor{poster_agent}3.73 \\[1ex]
    \rowcolor{poster_agent!50}
    Run 4 & 3.93 & 3.84 & 2.95 & 4.03 & 3.97 & 3.58 & \cellcolor{poster_agent}3.57 & \cellcolor{poster_agent}3.86 & \cellcolor{poster_agent}3.72 \\[1ex]
    \rowcolor{poster_agent!50}
    Run 5 & 3.93 & 3.85 & 2.93 & 4.01 & 3.95 & 3.64 & \cellcolor{poster_agent}3.57 & \cellcolor{poster_agent}3.87 & \cellcolor{poster_agent}3.72 \\[1ex]
    \midrule
    \rowcolor{poster_agent!30}
    \textbf{Mean} & 3.934 & 3.866 & 2.948 & 4.026 & 3.966 & 3.600 & \cellcolor{poster_agent}3.582 & \cellcolor{poster_agent}3.864 & \cellcolor{poster_agent}3.724 \\[1ex]
    \rowcolor{poster_agent!30}
    \textbf{Std} & 0.017 & 0.024 & 0.018 & 0.011 & 0.015 & 0.023 & \cellcolor{poster_agent}0.013 & \cellcolor{poster_agent}0.005 & \cellcolor{poster_agent}0.005 \\[1ex]
    \rowcolor{poster_agent!30}
    \textbf{95\% CI} & [3.913, & [3.836, & [2.926, & [4.012, & [3.947, & [3.571, & \cellcolor{poster_agent}[3.566, & \cellcolor{poster_agent}[3.857, & \cellcolor{poster_agent}[3.717, \\[-0.5ex]
    \rowcolor{poster_agent!30}
     & 3.955] & 3.896] & 2.970] & 4.040] & 3.985] & 3.629] & \cellcolor{poster_agent}3.598] & \cellcolor{poster_agent}3.871] & \cellcolor{poster_agent}3.731] \\
    \bottomrule
  \end{tabular}%
  }
  \caption{\textbf{Five-run robustness analysis of VLM-as-Judge evaluation for \texttt{\ouralg-4o}.} Results show exceptional stability with minimal variance (std $<$ 0.024) across all metrics.}
  \label{tab:vlm_judge_robustness}
\end{table*}

\section{Detailed Definition of Evaluation Metrics}

We elaborate on the details of all four types of evaluation metrics applied in this study in this section.

\subsection{Visual Quality Metrics} \label{app:vis_qual}

Two metrics fall into this type, namely \textit{Visual Similarity} and \textit{Figure Relevance}.

$\bullet$ \textit{Visual Similarity} is computed as the cosine similarity between the CLIP image embeddings of the generated poster $\hat{P}$ and the ground‑truth poster $P^{\ast}$. Concretely, letting  
\[
z_I(X) \;=\;\mathrm{CLIP}_{\mathrm{image}}(X)
\]
denote the CLIP image encoder, we set
\begin{equation}
    s_{\mathrm{VS}} \;=\;\text{cosine\_similarity}\,\!\bigl(z_I(\hat{P}),\,z_I(P^{\ast})\bigr)\;\in[-1,1].
\end{equation}
By operating at the instance level rather than comparing distributional statistics (e.g., FID~\cite{heusel2017gans}), this measure directly captures semantic alignment and overall content fidelity between individual poster images.

$\bullet$ \textit{Figure Relevance} assesses whether each figure in the generated poster is contextually appropriate. For a set of $N$ figure crops $\{f_i\}_{i=1}^N$ extracted from $\hat{P}$ and their corresponding section text $\{t_i\}_{i=1}^N$ from the original paper, we compute image and text embeddings  
\[
z_I(f_i)\;=\;\mathrm{CLIP}_{\mathrm{image}}(f_i),
\quad
z_T(t_i)\;=\;\mathrm{CLIP}_{\mathrm{text}}(t_i).
\]
We then define
\[
s_{\mathrm{FR}} =
\begin{cases}
\displaystyle \frac{1}{N}\sum_{i=1}^N \text{cosine\_similarity}\,\!\bigl(z_I(f_i),\,z_T(t_i)\bigr),
& N > 0,\\[6pt]
0, & N = 0.
\end{cases}
\]

\subsection{Textual Coherence Metrics}
\label{app:text_coh}

We quantify textual coherence by computing the standard perplexity (PPL) of the poster text under the \texttt{Llama-2-7b-hf} language model.  Specifically, let the poster be tokenized into a sequence \(w_{1:n}\).  The model assigns each token a conditional probability \(p(w_i \mid w_{<i})\).  We then define perplexity as

\[
\mathrm{PPL}
\;=\;
\exp\,\!\Bigl(
   -\frac{1}{n}\sum_{i=1}^{n}\log p(w_i \mid w_{<i})
\Bigr).
\]

Lower values of PPL correspond to more predictable and then more coherent text.  We employ full‑sequence PPL for its simplicity and direct interpretability in capturing overall textual fluency.

\subsection{Holistic Quality Assessment via VLMs (VLM-as-Judge)}
\label{app:vlm_judge}

Each poster is scored on six criteria by a vision–language model.  For each criterion we supply a dedicated prompt in a \texttt{tcolorbox} using the \texttt{prompt\_func} style; the model returns:
\begin{verbatim}
{"reason": "<justification>", "score": <1–5>}
\end{verbatim}

\bigskip
\noindent\textbf{Element Quality.}  
This criterion evaluates the visual clarity, resolution, and stylistic consistency of individual graphic elements (figures, charts, icons).  
\begin{tcolorbox}[prompt_aethetic_element]
\textbf{System Prompt}:
You are an extremely discerning visual-element judge. Scrutinize every figure, chart, and image for any visual or stylistic issue. Always look for even subtle flaws: low contrast, imperfect resolutions, slightly inconsistent styles, crowded or mislabeled legends, etc. Be wary of awarding high scores unless the visuals truly meet the strictest standards.\\[1ex]

\textbf{Instructions}: Five-Point Scale\\[1ex]

1 Point:

  \quad • Graphics are blurry, pixelated, or illegible.

  \quad • Color choices severely hinder interpretation.

  \quad • Visuals may significantly detract from comprehension.\\[1ex]

2 Points:

  \quad • At least one graphic is clear, while others suffer from poor resolution or style.

  \quad • Legends or labels are missing or too small to read comfortably.

  \quad • Color schemes create some confusion or difficulty.\\[1ex]

3 Points:

  \quad • Most graphics are legible and relevant, but have notable issues with consistency, sizing, or clarity.

  \quad • Some mismatches in style or color usage detract from cohesion.

  \quad • Minor but noticeable labeling/legend shortcomings.\\[1ex]

4 Points:

  \quad • High-quality graphics with generally consistent styling.

  \quad • Clear legends and color schemes aid interpretation.

  \quad • Any remaining flaws are slight and do not significantly hinder understanding.\\[1ex]

5 Points:

  \quad • Rarely awarded; strictly reserved for publication-grade visuals.

  \quad • Crisp resolution with no instances of blurriness.

  \quad • Harmonious color palette, impeccable labeling, and an exceptionally consistent style.\\[1ex]

Example Output:
\begin{verbatim}
{"reason": "...", "score": int}
\end{verbatim} \\[1ex]

Think step by step and be conservative with your rating.
\end{tcolorbox}

\medskip
\noindent\textbf{Layout Balance.}  
This criterion assesses the overall arrangement, alignment, and spacing of text and graphics to ensure a coherent and readable poster structure.  

\begin{tcolorbox}[prompt_aethetic_Layout]
\textbf{System Prompt}:  
You are an uncompromising poster-layout judge. Critique the overall arrangement of all visual components (text blocks, headings, figures, white-space, alignment) that affect readability. Always scan for subtle alignment issues, uneven spacing, or any layout feature that might disrupt reader comprehension. Resist giving high scores unless the layout is exceptionally polished.\\[1ex]

\textbf{Instructions}: Five-Point Scale\\[1ex]

1 Point:

  \quad • Highly disorganized layout; elements overlap, making text or graphics illegible.  
   
  \quad • Margins are violated or reading path is nearly impossible to follow.  
   
  \quad • Severely hinders comprehension.\\[1ex]

2 Points:

  \quad • Some semblance of structure (columns/rows) but marred by inconsistent alignment or overcrowded sections.  
   
  \quad • White-space distribution may be haphazard or insufficient.  
   
  \quad • Reading flow is interrupted, though one can still piece it together.\\[1ex]

3 Points:

  \quad • Recognizable structure with mostly consistent alignment and spacing.  
   
  \quad • Some minor layout distractions remain (e.g., slightly cramped text, uneven spacing, small alignment slips).  
   
  \quad • Generally readable but not particularly polished.\\[1ex]

4 Points:

  \quad • Well-organized grid or arrangement; logical reading path that mostly flows.  
   
  \quad • Appropriate font sizes, spacing, and alignment; only subtle layout imperfections.  
   
  \quad • White-space usage clean and deliberate; nearly professional.\\[1ex]

5 Points:

  \quad • Very rarely granted; must be a pristine, professional-grade layout.  
   
  \quad • Seamless alignment, balanced spacing, and expertly guided reading path.  
   
  \quad • Flawless design synergy that maximizes readability and comprehension.\\[1ex]

Example Output:
\begin{verbatim}
{"reason": "...", "score": int}
\end{verbatim}\\[1ex]

Think step by step and be tough on small alignment/spacing issues.
\end{tcolorbox}

\medskip
\noindent\textbf{Engagement.}  
This criterion judges how effectively the poster’s design elements—color, typography, and composition—capture and sustain viewer attention.  

\begin{tcolorbox}[prompt_aethetic_engagement]
\textbf{System Prompt}:  
You are an uncompromising poster-aesthetics judge focusing on engagement. Be extremely critical of color harmony, typography, visual balance, and the poster’s ability to grab and hold attention. Always look for subtle issues—color clashes, overly busy or dull designs, inappropriate font choices, awkward spacing, or anything that might reduce engagement. Reserve high scores for truly exemplary work.\\[1ex]

\textbf{Instructions}: Five-Point Scale\\[1ex]

1 Point:

  \quad • Visually off-putting; clashing colors or crowded design repel viewers.  
   
  \quad • Typography choice is jarring or illegible at a glance.  
   
  \quad • Overall fails to engage or entice.\\[1ex]

2 Points:

  \quad • Some visually appealing elements exist but are overshadowed by dull or inconsistent design moments.  
   
  \quad • Font sizes or styles reduce accessibility or attractiveness.  
   
  \quad • Limited capacity to draw an audience's focus.\\[1ex]

3 Points:

  \quad • Shows generally pleasing color scheme and typography, though lacking a “wow” factor.  
   
  \quad • Balance and visual flow are acceptable but reveal minor weaknesses (e.g., slightly crowded or sparse areas).  
   
  \quad • Engagement is average; neither strong nor particularly weak.\\[1ex]

4 Points:

  \quad • Eye-catching design using mostly harmonious colors and effective typography.  
   
  \quad • Good use of negative space; the layout guides the viewer’s eye effectively.  
   
  \quad • Only minor flaws or bland spots prevent it from being top-tier.\\[1ex]

5 Points:

  \quad • Rarely awarded—reserved for truly striking, magazine-cover-caliber visuals.  
   
  \quad • Flawless color palette and typography; everything works together seamlessly.  
   
  \quad • Immediately captivating design that retains audience interest without any noticeable weakness.\\[1ex]

Example Output:
\begin{verbatim}
{"reason": "...", "score": int}
\end{verbatim}\\[1ex]

Think step by step and be very conservative when scoring.
\end{tcolorbox}

\bigskip
\noindent\textbf{Clarity.}  
This criterion evaluates sentence-level readability, grammar, and phrasing to ensure the text is polished and error-free.

\begin{tcolorbox}[prompt_info_clarity]
\textbf{System Prompt}:  
You are an uncompromising micro-text judge. Critically evaluate sentence-level clarity, grammar, phrasing, and intra-section coherence. Look for even subtle grammatical slips, confusing jargon, or clumsy phrasing. Be slow to award top marks unless the text is impeccably polished.\\[1ex]

\textbf{Instructions}: Five-Point Scale\\[1ex]

1 Point:

  \quad • Rampant grammatical or spelling errors; sentences may be unreadable.  
   
  \quad • Overly technical jargon without explanations; fragments or run-ons predominate.  
   
  \quad • Overall, text quality severely impedes understanding.\\[1ex]

2 Points:

  \quad • Meaning is generally discernible, but multiple grammar or syntax problems appear in each section.  
   
  \quad • Awkward or unclear phrasing disrupts the flow of reading.  
   
  \quad • Only partial clarity is achieved.\\[1ex]

3 Points:

  \quad • Overall readable text with a few noticeable grammar or wording missteps.  
   
  \quad • Occasional awkward phrasing or redundancies appear, but readers can follow without major confusion.  
   
  \quad • Average clarity.\\[1ex]

4 Points:

  \quad • Well-written, mostly free of grammatical or spelling errors.  
   
  \quad • Terminology is used properly; text flows smoothly within paragraphs.  
   
  \quad • Minor slip-ups can be present but do not disrupt understanding.\\[1ex]

5 Points:

  \quad • Exceptional text quality, error-free, and elegantly phrased.  
   
  \quad • Complex ideas conveyed with clear, concise language.  
   
  \quad • Granted only if absolutely no grammatical, spelling, or stylistic flaws are detected.\\[1ex]

Example Output:
\begin{verbatim}
{"reason": "...", "score": int}
\end{verbatim}\\[1ex]

Think step by step.
\end{tcolorbox}

\medskip
\noindent\textbf{Content Completeness.}  
This criterion measures whether all key sections are included and richly detailed, reflecting comprehensive coverage of the paper’s main contributions.

\begin{tcolorbox}[prompt_info_content]
\textbf{System Prompt}:  
You are an uncompromising content-depth judge. Assess whether the poster includes all essential sections and whether each section presents sufficient detail. Look for any missing or under-developed segments; do not hesitate to penalize for insufficient depth. Award the highest scores only if the poster expertly covers every necessary aspect.\\[1ex]

\textbf{Instructions}: Five-Point Scale\\[1ex]

1 Point:

  \quad • Critical sections (e.g., objectives or results) are completely missing or trivial.  
   
  \quad • Data grossly insufficient to comprehend the study or conclusions.  
   
  \quad • Very poor depth that fails to convey essential information.\\[1ex]

2 Points:

  \quad • Most key sections appear but major details (context, data, references) are absent.  
   
  \quad • Lack of elaboration on methods or results leaves big gaps.  
   
  \quad • Overall content too shallow to properly inform.\\[1ex]

3 Points:

  \quad • All standard sections included with fundamental information.  
   
  \quad • Some omissions or scant detail in certain areas (e.g., results or methodology).  
   
  \quad • Only moderate depth; the reader must fill many gaps themselves.\\[1ex]

4 Points:

  \quad • All essential sections present, each treated with adequate-to-strong detail.  
   
  \quad • Robust description of objectives, methods, results, and references.  
   
  \quad • Only minor improvements needed.\\[1ex]

5 Points:

  \quad • Very rarely granted; everything must be comprehensive and thorough.  
   
  \quad • Exhaustive detail on methodology, results (with statistics), interpretation, references, and future work.  
   
  \quad • Leaves readers with minimal unanswered questions.\\[1ex]

Example Output:
\begin{verbatim}
{"reason": "...", "score": int}
\end{verbatim}\\[1ex]

Think step by step.
\end{tcolorbox}

\medskip
\noindent\textbf{Logical Flow.}  
This criterion examines the coherence and progression of ideas across poster sections, ensuring a seamless narrative from introduction to conclusion.

\begin{tcolorbox}[prompt_info_logic]
\textbf{System Prompt}:  
You are an uncompromising macro-logic judge. Examine how well the poster’s major sections (Introduction, Methods, Results, Conclusions, etc.) connect to form a coherent narrative. Pay attention to continuity, how logically each section flows from the previous, and whether there are any abrupt gaps. Only award the highest marks if the storyline is perfectly seamless.\\[1ex]

\textbf{Instructions}: Five-Point Scale\\[1ex]

1 Point:

  \quad • Sections are disjointed; little to no logical connection between them.  
   
  \quad • Key transitions or the central rationale is missing, creating confusion.\\[1ex]

2 Points:

  \quad • General sequence recognizable but important logical steps are weak or missing.  
   
  \quad • Readers must infer key links.\\[1ex]

3 Points:

  \quad • Mostly coherent narrative with minor gaps.  
   
  \quad • Transitions exist but some logical steps are lightly justified.\\[1ex]

4 Points:

  \quad • Well-structured storyline; each section clearly builds on the previous.  
   
  \quad • Transitions are stated, rationale is mostly strong.\\[1ex]

5 Points:

  \quad • Extremely rare; flawless logical flow from introduction to conclusion.  
   
  \quad • Seamless transitions; no inferential leaps.\\[1ex]

Example Output:
\begin{verbatim}
{"reason": "...", "score": int}
\end{verbatim}\\[1ex]

Think step by step and penalize any noticeable logical gap or awkward transition.
\end{tcolorbox}

\bigskip

For each poster, we record all six criterion scores and compute two aggregated metrics:

\begin{align*}
\text{Aesthetic Score}
&= \frac{\text{Element Quality} + \text{Layout Balance} + \text{Engagement}}{3},\\
\text{Information Score}
&= \frac{\text{Clarity} + \text{Content Completeness} + \text{Logical Flow}}{3}.
\end{align*}

\subsection{\kqa}
\label{app:qa}

\noindent\textbf{QA Dataset Curation.}  
Each paper PDF is converted to markdown via our PDF parser. We then prompt o3 to generate 100 multiple‑choice questions per paper, where we have 50 verbatim and 50 interpretive questions as follows:
\begin{itemize}
  \item \emph{Verbatim questions (50)}: directly answerable from the paper text, covering 13 orthogonal content aspects (e.g., objectives, methodology, key results).
  \item \emph{Interpretive questions (50)}: requires high‑level comprehension beyond verbatim text, spanning 10 conceptual dimensions (e.g., motivation, contribution synthesis, implication analysis).
\end{itemize}

The exact prompts that are applied to generate the questions are given below, for verbatim and interpretive questions, respectively.

\begin{tcolorbox}[prompt_verbatim_question]
\textbf{System Prompt}:  
You are a Question‑Generation agent for academic posters. Your task is to read the supplied Markdown text (\texttt{document\_markdown}) and produce \textbf{exactly 50 multiple-choice QA items} whose answers can be located verbatim or nearly verbatim in that text. The questions must be suitable for conference‑poster readers: avoid deep theoretical proofs, reference lists, or citation minutiae. Follow all guidelines below precisely.\\[1ex]

\textbf{Instructions}:  

1. Carefully read the Markdown in \texttt{document\_markdown}. 

  \quad • Each question must map to one clear sentence or phrase in the poster text.  
  
  \quad • No duplicate or near‑duplicate wording.\\[1ex]

2. Write 50 factual, answerable‑from‑text questions.  
  
  \quad • Vary difficulty from easy “headline” facts to specific numeric or procedural details.\\[1ex]

3. Distribute the 50 questions across the following poster‑friendly aspects, aiming for 2–5 questions per aspect and ensuring each aspect appears at least once:  

  \quad A. Title \& authorship (title, author names, affiliations, keywords)  
  
  \quad B. Motivation / problem statement / research gap  
  
  \quad C. Objectives or hypotheses  
  
  \quad D. Dataset(s) or experimental materials  
  
  \quad E. Methodology (algorithms, model architecture, workflow steps)  
  
  \quad F. Key parameters or hyper‑parameters (values, settings)  
  
  \quad G. Evaluation metrics or criteria  
  
  \quad H. Quantitative results (numbers in tables, charts) 
  
  \quad I. Qualitative findings, figures, or illustrative examples  
  \quad J. Comparative or ablation study results  
  
  \quad K. Conclusions, implications, or contributions  
  
  \quad L. Limitations or future work  
  
  \quad M. Definitions of domain‑specific terms or abbreviations\\[1ex]

4. EXCLUDE references, citations, author acknowledgements, and any text that would not appear on a standard poster.\\[1ex]

5. Use the following JSON‑for‑each format (exact spelling \& casing):  
\begin{verbatim}
{
  "Question X": {
    "aspect": "<A‑M>",
    "question": "<single sentence>",
    "options": [
      "A. <choice 1>",
      "B. <choice 2>",
      "C. <choice 3>",
      "D. <choice 4>"
    ],
    "answer": "<Letter>. <exact correct option text>"
  },
  …
}
\end{verbatim}\\[1ex]

6. Output **only** the final JSON object containing 50 items—no additional commentary.\\[1ex]

7. Balance the correct answers roughly equally among options A–D.\\[1ex]

Example Output:  
\begin{verbatim}
{"Question 1": {…}, "Question 2": {…}, …, "Question 50": {…}}
\end{verbatim}\\[1ex]

Think step by step and ensure full compliance with every guideline.
\end{tcolorbox}

\begin{tcolorbox}[prompt_interpretive_question]
\textbf{System Prompt}:  
You are a Question-Generation agent.  
Your task is to read the supplied Markdown text (\texttt{document\_markdown}) and create \textbf{exactly 50 multiple-choice questions} that capture a *high-level understanding* of the work—its purpose, novelty, core approach, and overall findings. Every question must still be answerable by locating explicit sentences or phrases in the text; do not require inference that is absent from the poster-style content.\\[1ex]

\textbf{Instructions}:  

1. Read the Markdown in \texttt{document\_markdown} closely.  

  \quad • Each question must map to explicit content in the text.  
  
  \quad • Do not require inference beyond presented poster-level information.\\[1ex]

2. Draft 50 factual questions probing the reader’s global grasp (e.g., “What problem does the study address?”).  

  \quad • Avoid low-level numeric settings, code snippets, or reference lists.  
  
  \quad • Vary wording and avoid duplicates.\\[1ex]

3. Cover all of the following *high-level* aspects—each must appear at least twice to guarantee breadth:  

  \quad A. Research domain \& background context  
  
  \quad B. Central problem / motivation / research gap  
  
  \quad C. Primary goal, hypothesis, or research question  
  
  \quad D. Key contributions or novelty statements  
  
  \quad E. Overall methodology or workflow (summarized)  
  
  \quad F. Principal findings or headline quantitative results  
  
  \quad G. Qualitative insights or illustrative examples  
  
  \quad H. Implications, applications, or significance  
  
  \quad I. Limitations or future-work directions  
  
  \quad J. Main conclusions or take-home messages\\[1ex]

4. EXCLUDE citations, granular hyper-parameters, precise numeric tables, and acknowledgements—stick to poster-level overview content.\\[1ex]

5. Return the questions in the following *strict* JSON schema:  
\begin{verbatim}
{
  "Question X": {
    "aspect": "<A-J>",
    "question": "<one concise sentence>",
    "options": [
      "A. <choice 1>",
      "B. <choice 2>",
      "C. <choice 3>",
      "D. <choice 4>"
    ],
    "answer": "<Letter>. <exact correct option text>"
  },
  ...
}
\end{verbatim}\\[1ex]

6. Produce **only** the final JSON object with 50 entries—no commentary, headers, or extra lines.\\[1ex]

7. The number of correct answers should be approximately balanced across A–D.\\[1ex]

\textbf{Document Markdown}:  
\texttt{\{\{ document\_markdown \}\}}\\[1ex]

\textit{Output ONLY the JSON with 50 questions below}
\end{tcolorbox}

\medskip
\noindent\textbf{Evaluation Workflow.}  
For each poster image, we query six VLM reader models to answer curated questions. These models include three open-source models (LLaVA-OneVision-Qwen2-7B-ov-hf, Phi-4-multimodal-instruct, and Llama-4-Scout-17B-16E-Instruct) and three closed-source models (o3, GPT-4o mini, and Gemini 2.0 Flash). Their outputs are evaluated according to two enforced rules:
\begin{itemize}
  \item \textbf{No external knowledge.} Models must base answers solely on information present in the poster image.
  \item \textbf{Visual citation.} Each answer must include a reference to the poster region supporting it (e.g., “See Figure 2 caption”); if no region contains the answer, the model responds “NA.”
\end{itemize}

\begin{tcolorbox}[prompt_answer_agent]
\textbf{System Prompt}:  
You are an answering agent. You will be provided with:  

  1. An image of a poster.  
  
  2. A JSON object called “questions” which contains multiple questions. Each question has four possible answers: A, B, C, or D.  

Your goal is to analyze the poster thoroughly and answer each question based on the information it provides. You should **NOT** use any external knowledge or context beyond the poster image. You must rely solely on the content of the poster to answer the questions.  

For each question:  

  \quad • If you find enough evidence in the poster to decide on a specific option (A, B, C, or D), then choose that option and include a brief reference to the part of the poster that supports your answer (e.g., “Top-left text”, “Event date section”, etc.).  
  
  \quad • If the poster does not offer sufficient information to confidently choose any of the options, respond with "NA" for both the answer and the reference.\\[1ex]

\textbf{Instructions}:  
1. Study the poster image along with the “questions” provided.\\[1ex]  

2. For each question:  

  \quad • Decide if the poster clearly supports one of the four options (A, B, C, or D). If so, pick that answer.  
  
  \quad • Otherwise, if the poster does not have adequate information, use "NA" for the answer.\\[1ex]  
  
3. Provide a brief reference indicating where in the poster you found the answer. If no reference is available (i.e., your answer is "NA"), use "NA" for the reference too.\\[1ex]  

4. Format your output strictly as a JSON object with this pattern:  
\begin{verbatim}
{
  "Question 1": {
    "answer": "X",
    "reference": "some reference or 'NA'"
  },
  "Question 2": {
    "answer": "X",
    "reference": "some reference or 'NA'"
  },
  ...
}
\end{verbatim}\\[1ex]  

5. Do not include any explanations or extra keys beyond the specified structure.\\[1ex]  

6. You must provide an answer entry for all questions in the “questions” object.\\[1ex]

\textbf{Example Output}:  
\begin{verbatim}
{
  "Question 1": {
    "answer": "B",
    "reference": "Description on the top-right of the poster"
  },
  "Question 2": {
    "answer": "NA",
    "reference": "NA"
  }
}
\end{verbatim}
\end{tcolorbox}

\medskip
\noindent\textbf{Scoring Metrics.}  
Let \(s_R\) be the raw accuracy (fraction of correctly answered questions) and \(l\) the token count of the poster text. We define the \emph{density‑augmented score}
\[
s_A \;=\; s_R \;\Bigl(1 + \frac{1}{\max(1,\,l/w)}\Bigr),
\]
where \(w\) is the median text length of ground‑truth posters. The density multiplier is capped at 2 to penalize verbosity and reward concise, information‑dense designs.

\section{Human Evaluation Protocol}\label{append:human_eval}
\textbf{Instructions.} Each human evaluator follows the instructions as follow,
\begin{itemize}
    \item You will be given a poster, as well as 6 text files containing the criteria to judge the poster.
    \item You need to read the poster and provide your scores according to the 6 text files' criteria.
\end{itemize}
\textbf{Criteria.} The criteria are the same as those outlined in \kqa\ \ref{app:qa}.

\section{Error Analysis}

Generating a scientific poster requires tight coupling of language understanding, visual synthesis, and spatial layout reasoning. Across the five pipelines we evaluate—\texttt{4o-Image}, \texttt{4o-HTML}, \texttt{OWL-4o}, \texttt{PPTAgent}, and our proposed \texttt{\ouralg{}}—we consistently observe four high-level failure modes: text integrity issues, visual / layout flaws, missing visuals, and overflow issues. Below, we describe each class of error and highlight representative examples.

\subsection{Text Integrity Issues}

\begin{figure*}[!t]
  \centering
  \begin{tabular}{cc}
    \begin{subfigure}{0.48\linewidth}
      \centering
\includegraphics[width=\linewidth]{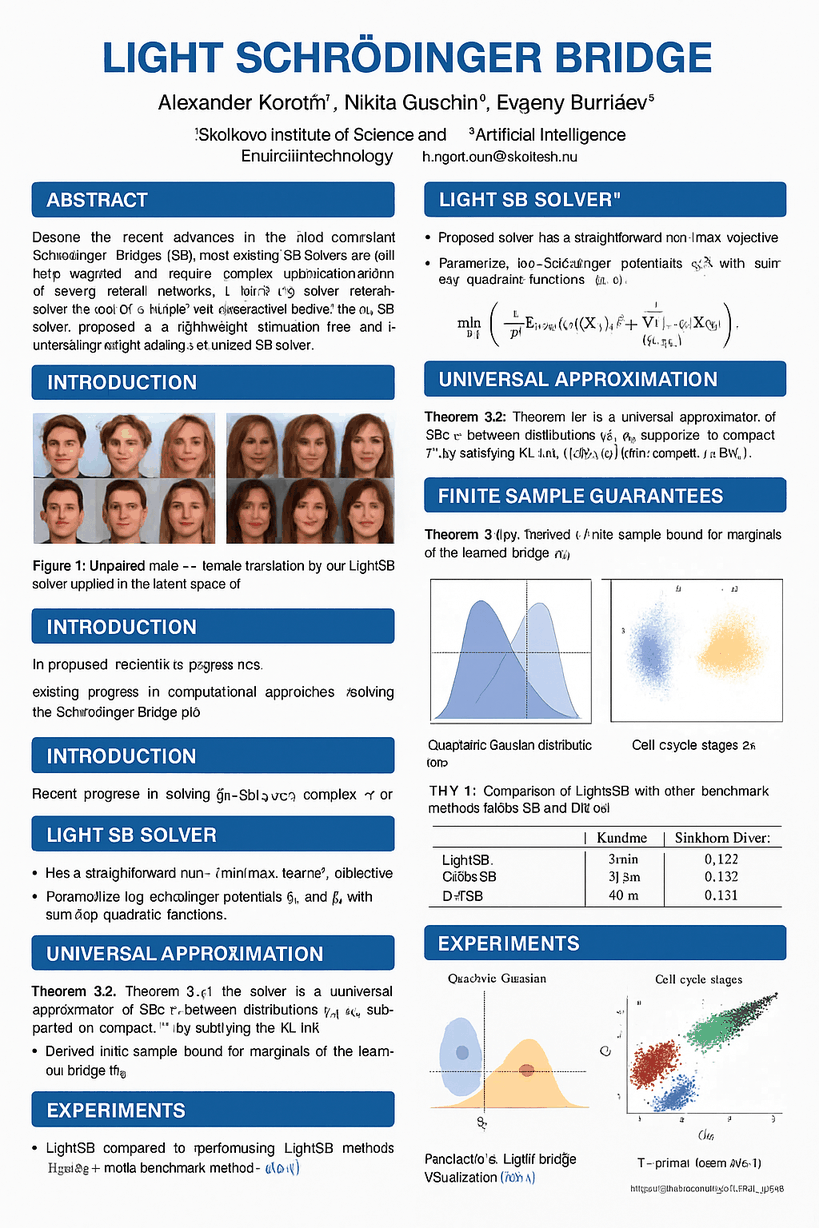}
      \subcaption{A poster generated by \texttt{4o-Image}, where substantial corrupted text is generated.}
      \label{fig:error_corrupted_text_a}
    \end{subfigure}
    &
    \begin{subfigure}{0.48\linewidth}
      \centering
 \includegraphics[width=\linewidth]{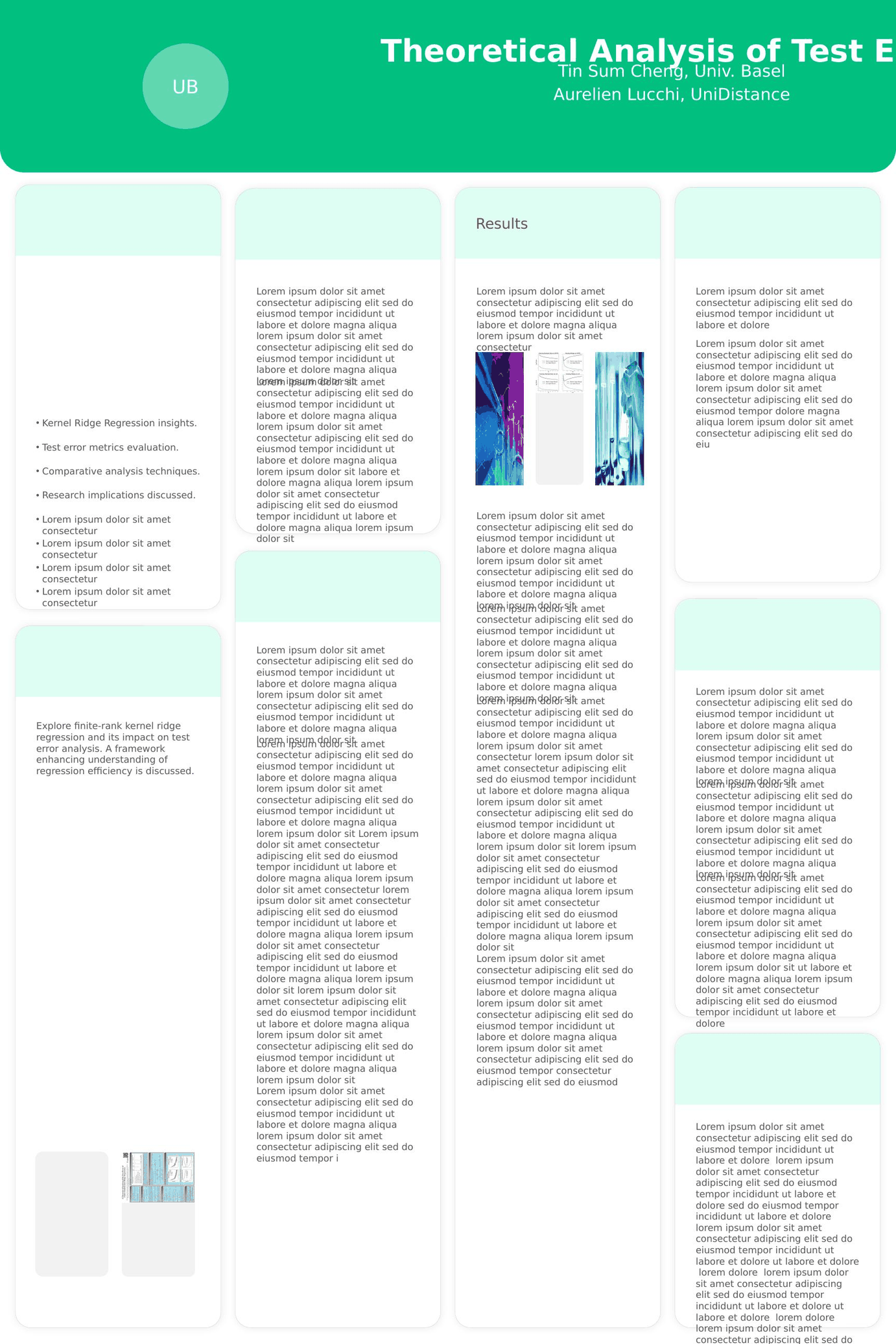}
      \subcaption{A poster generated by \texttt{PPTAgent}, where meaningless template placeholder text is remained.}
      \label{fig:error_corrupted_text_b}
    \end{subfigure}
  \end{tabular}
  \caption{Examples of posters with corrupted text.}
  \label{fig:error_corrupted_text}
\end{figure*}

Legible text is crucial for conveying a paper’s content. In image–only generation (\texttt{4o-Image}), posters often contain garbled or unreadable text (Fig.\ref{fig:error_corrupted_text_a}) because pixel-level synthesis struggles with high-resolution typography, underscores the fragility of text rendering when no explicit semantic control is applied. \texttt{PPTAgent}, as a template-based method, exhibits a different variant: placeholders are left intact or partly overwritten (Fig.\ref{fig:error_corrupted_text_b}), producing semantically “corrupted” content.

\subsection{Visual / Layout Flaws}

\begin{figure*}[!t]
  \centering
  \begin{tabular}{cc}
    \begin{subfigure}{0.48\linewidth}
      \centering
\includegraphics[width=\linewidth]{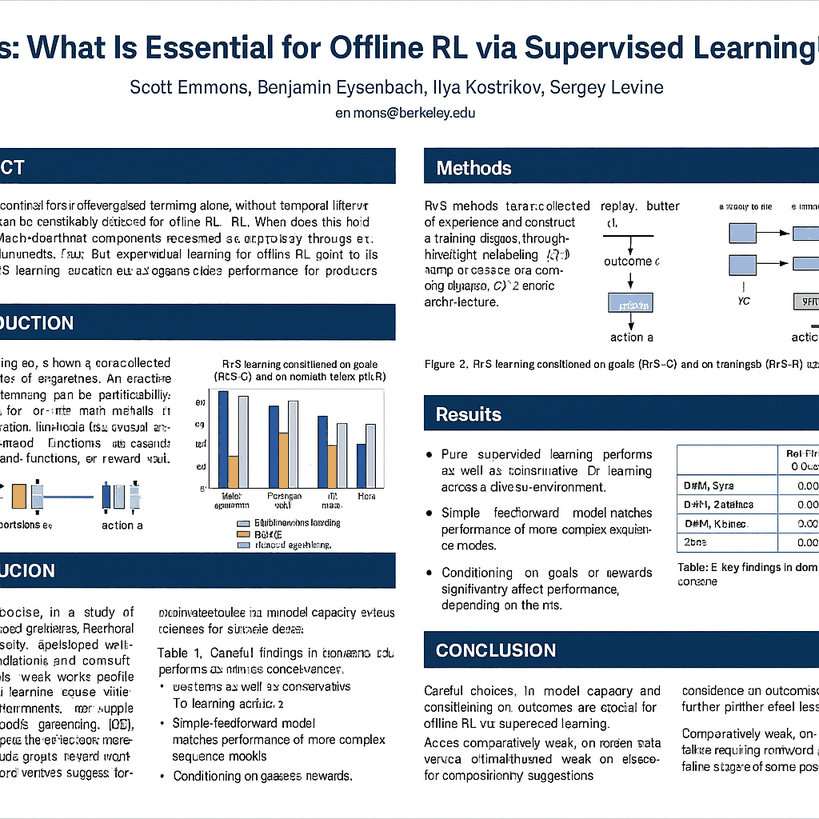}
      \subcaption{A poster generated by \texttt{4o-Image}, where the poster is cutoff horizontally due to incomplete generation.}
    \label{fig:error_cutoff_a}
    \end{subfigure}
    &
    \begin{subfigure}{0.48\linewidth}
      \centering
 \includegraphics[width=\linewidth]{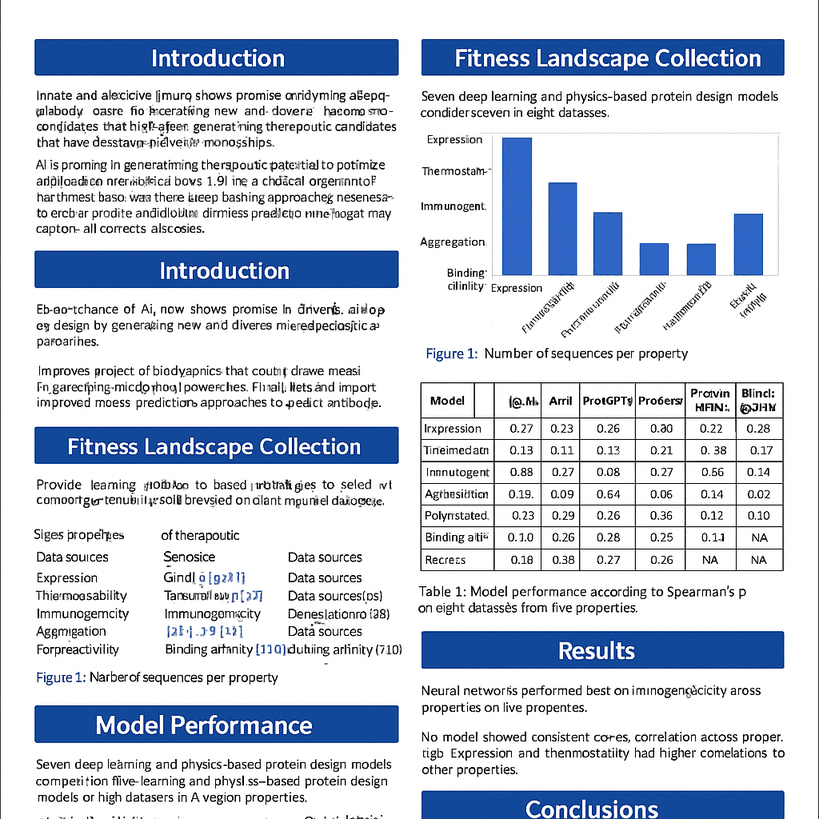}
      \subcaption{A poster generated by \texttt{4o-Image}, where the poster is cutoff vertically due to incomplete generation.}
      \label{fig:error_cutoff_b}
    \end{subfigure}
  \end{tabular}
  \caption{Examples of posters with cutoff.}
  \label{fig:error_cutoff}
\end{figure*}

\begin{figure*}[!t]
  \centering
  \begin{tabular}{cc}
    \begin{subfigure}{0.48\linewidth}
      \centering
\includegraphics[width=\linewidth]{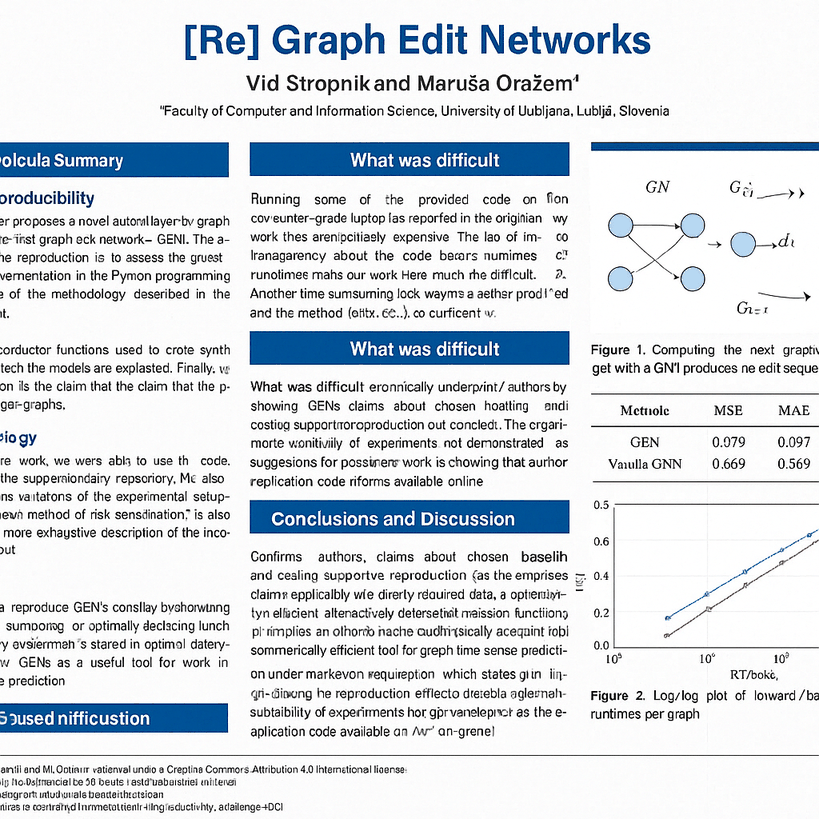}
      \subcaption{A poster produced by \texttt{4o-Image}, featuring a figure that is low-resolution, visually corrupted, and unintelligible.}
      \label{fig:error_obsure_figure_a}
    \end{subfigure}
    &
    \begin{subfigure}{0.48\linewidth}
      \centering
 \includegraphics[width=\linewidth]{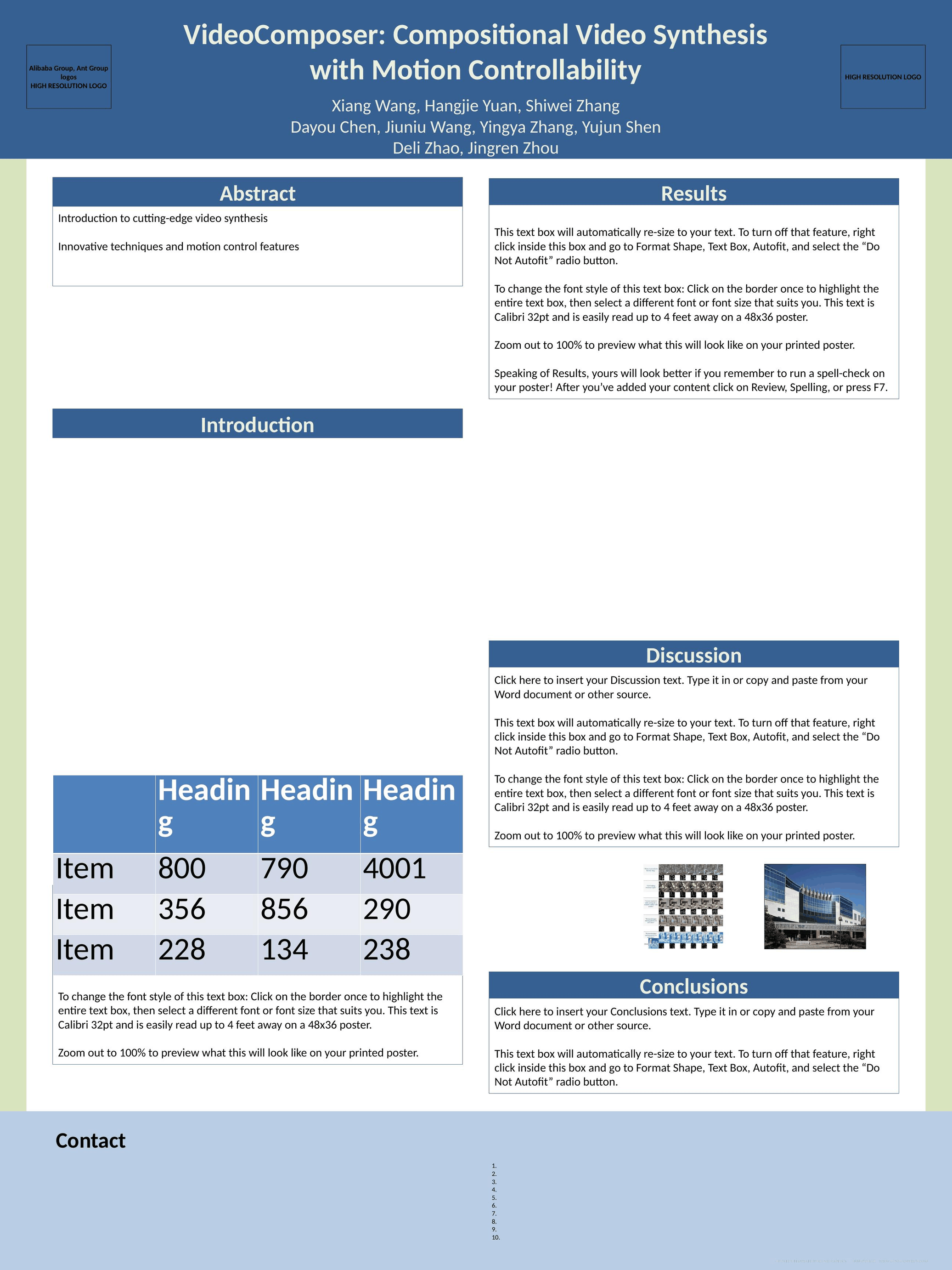}
      \subcaption{A poster generated by \texttt{PPTAgent}, where figures are rendered too small to be legible.}
      \label{fig:error_obsure_figure_b}
    \end{subfigure}
  \end{tabular}
  \caption{Examples of posters with obscure figures.}
  \label{fig:error_obscure_figure}
\end{figure*}

\begin{figure*}[!t]
  \centering
  \begin{tabular}{cc}
    \begin{subfigure}{0.48\linewidth}
      \centering
\includegraphics[width=\linewidth]{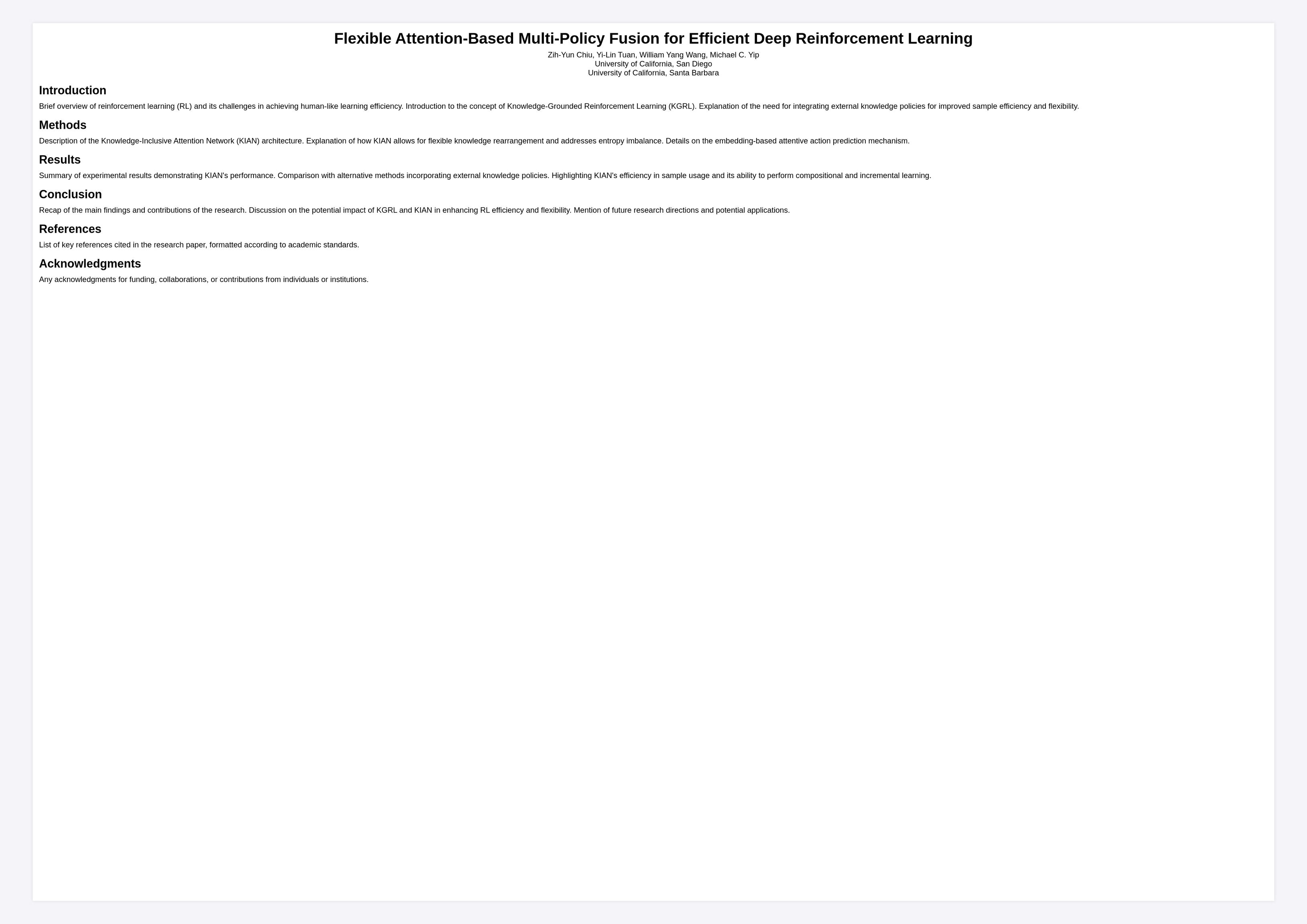}
      \subcaption{A poster generated by \texttt{OWL-4o}, where there are large blanks on the poster.}
      \label{fig:error_large_blanks_a}
    \end{subfigure}
    &
    \begin{subfigure}{0.48\linewidth}
      \centering
 \includegraphics[width=\linewidth]{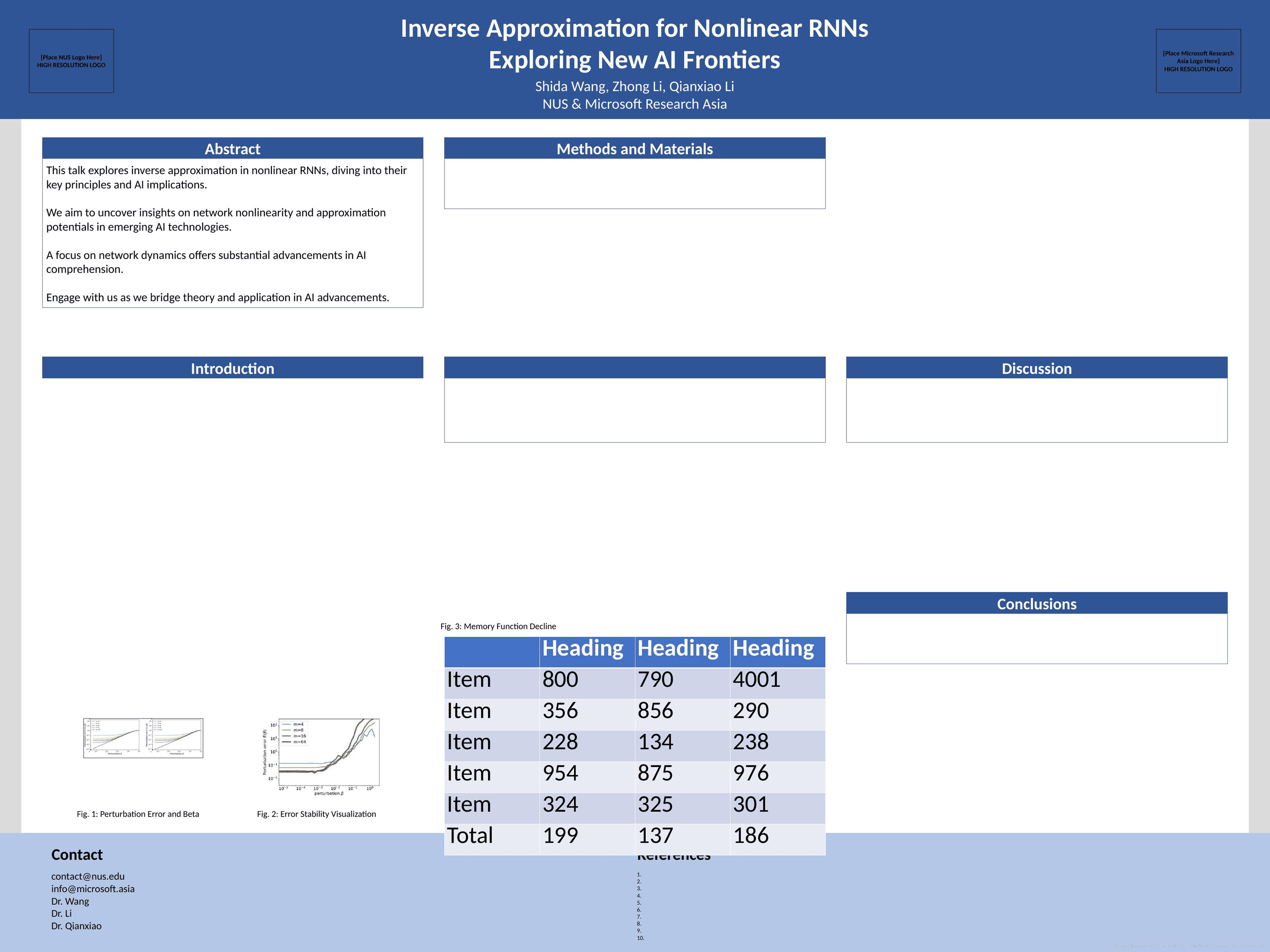}
      \subcaption{A poster generated by \texttt{PPTAgent}, where there are large blanks on the poster.}
      \label{fig:error_large_blanks_b}
    \end{subfigure}
  \end{tabular}
  \caption{Examples of posters with large blanks.}
  \label{fig:error_large_blanks}
\end{figure*}

Pipelines without robust visual feedback frequently misplace or distort content. \texttt{4o-Image} outputs can be truncated horizontally or vertically (Fig.\ref{fig:error_cutoff_a}, \ref{fig:error_cutoff_b}) because the generator lacks hard spatial constraints. The same model sometimes hallucinates nonsensical figures (Fig.\ref{fig:error_obsure_figure_a}). Even with a predefined template, \texttt{PPTAgent} may insert figures at unusably small scales (Fig.\ref{fig:error_obsure_figure_b}), or leave substantial blank regions when text or images are partially generated (Fig.\ref{fig:error_large_blanks_b}). HTML-based agents such as \texttt{OWL-4o} also suffer from large empty areas (Fig.~\ref{fig:error_large_blanks_a}) when their sequential code lacks iterative, visual validation.

\subsection{Missing Visuals}

\begin{figure*}[!t]
  \centering
  \begin{tabular}{cc}
    \begin{subfigure}{0.48\linewidth}
      \centering
\includegraphics[width=\linewidth]{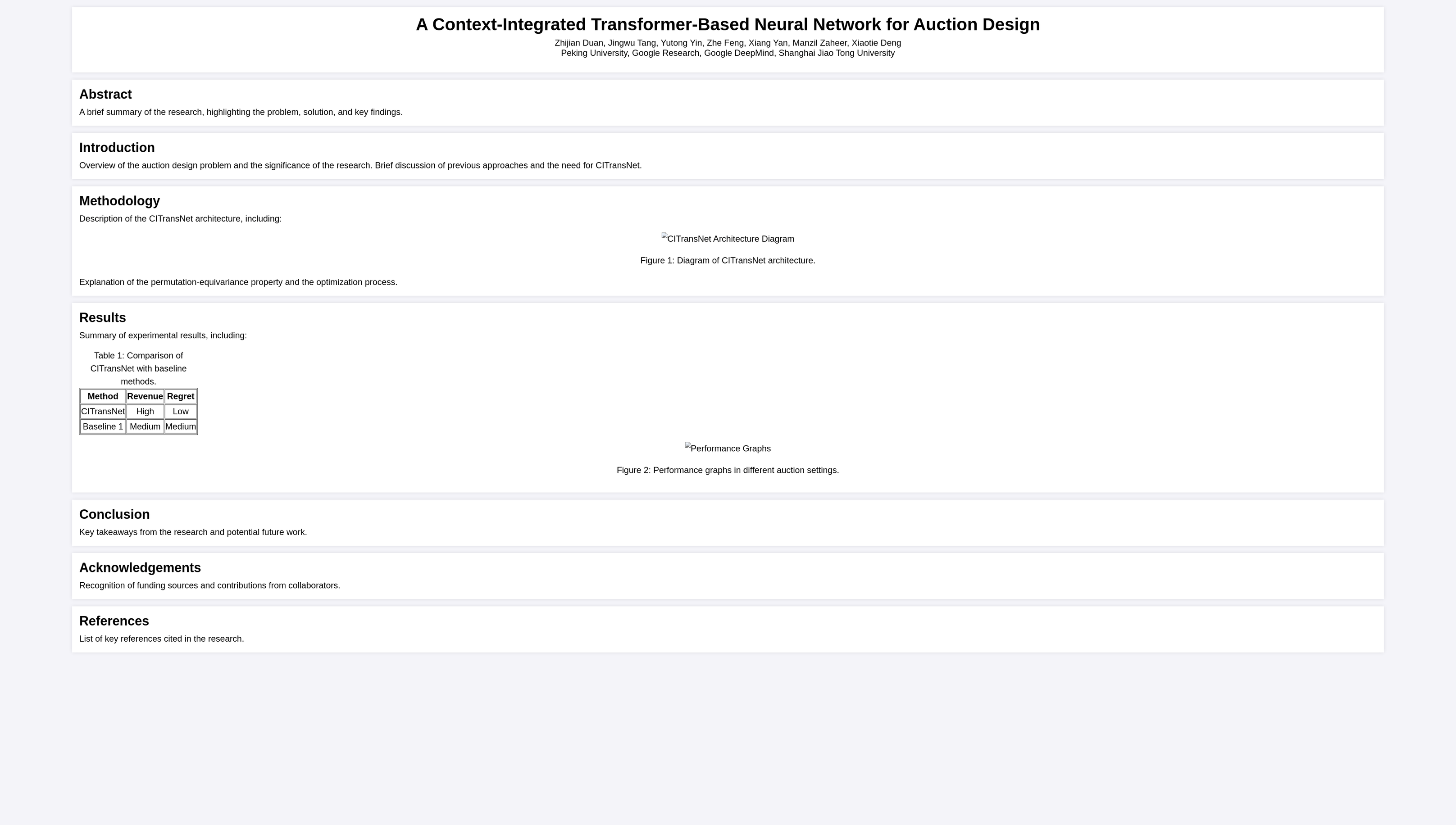}
      \subcaption{A poster generated by \texttt{OWL-4o}, where no figures are inserted into poster.}
      \label{fig:error_no_image_a}
    \end{subfigure}
    &
    \begin{subfigure}{0.48\linewidth}
      \centering
 \includegraphics[width=\linewidth]{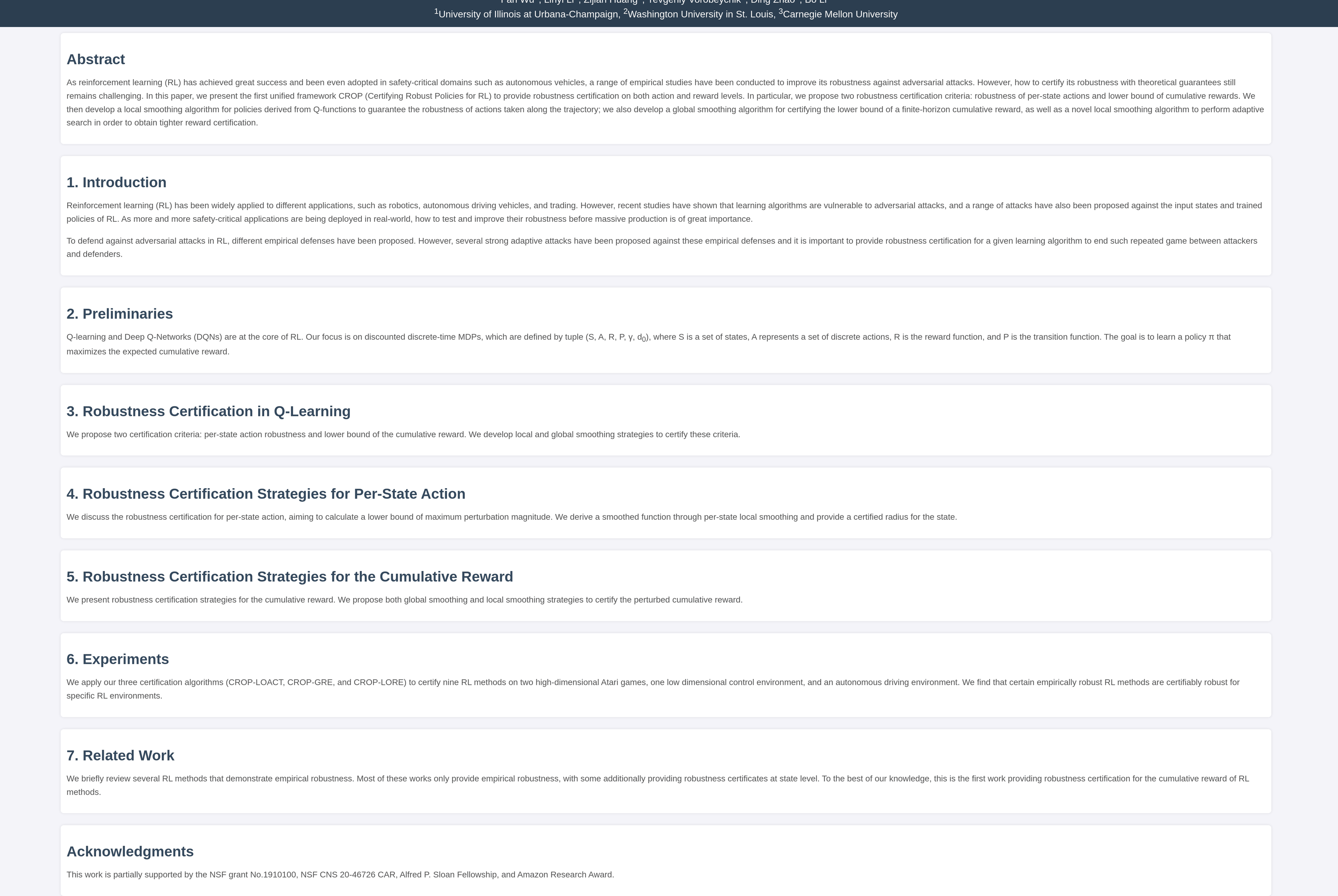}
      \subcaption{A poster generated by \texttt{4o-HTML}, where no figures are inserted into poster.}
      \label{fig:error_no_image_b}
    \end{subfigure}
  \end{tabular}
  \caption{Examples of posters without figures.}
  \label{fig:error_no_image}
\end{figure*}

Although \texttt{OWL-4o} is, in principle, able to invoke external toolkits for figure extraction, it fails to complete the full retrieval-insert cycle; the resulting posters remain purely textual (Fig.~\ref{fig:error_no_image_a}) On the other hand, \texttt{4o-HTML} \ref{fig:error_no_image_b}) by design is text-only, leading to similar issues.

\subsection{Overflow Issues}

\begin{figure*}[!t]
  \centering
  \begin{tabular}{cc}
    \begin{subfigure}{0.48\linewidth}
      \centering
\includegraphics[width=\linewidth]{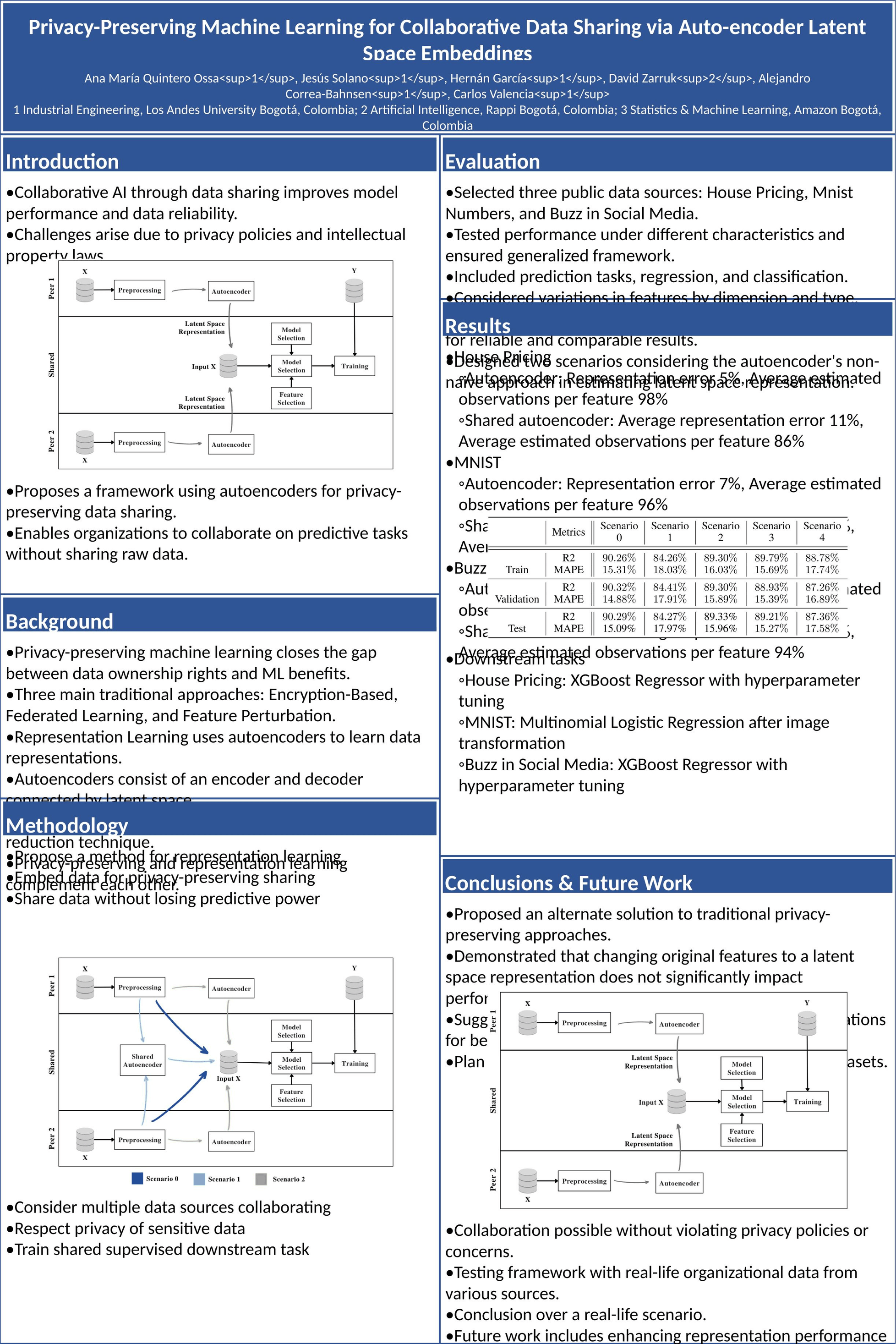}
      \subcaption{A poster generated by \texttt{\ouralg{}-Qwen}, where there is text overflowing outside textbox.}
      \label{fig:error_overflow_a}
    \end{subfigure}
    &
    \begin{subfigure}{0.48\linewidth}
      \centering
 \includegraphics[width=\linewidth]{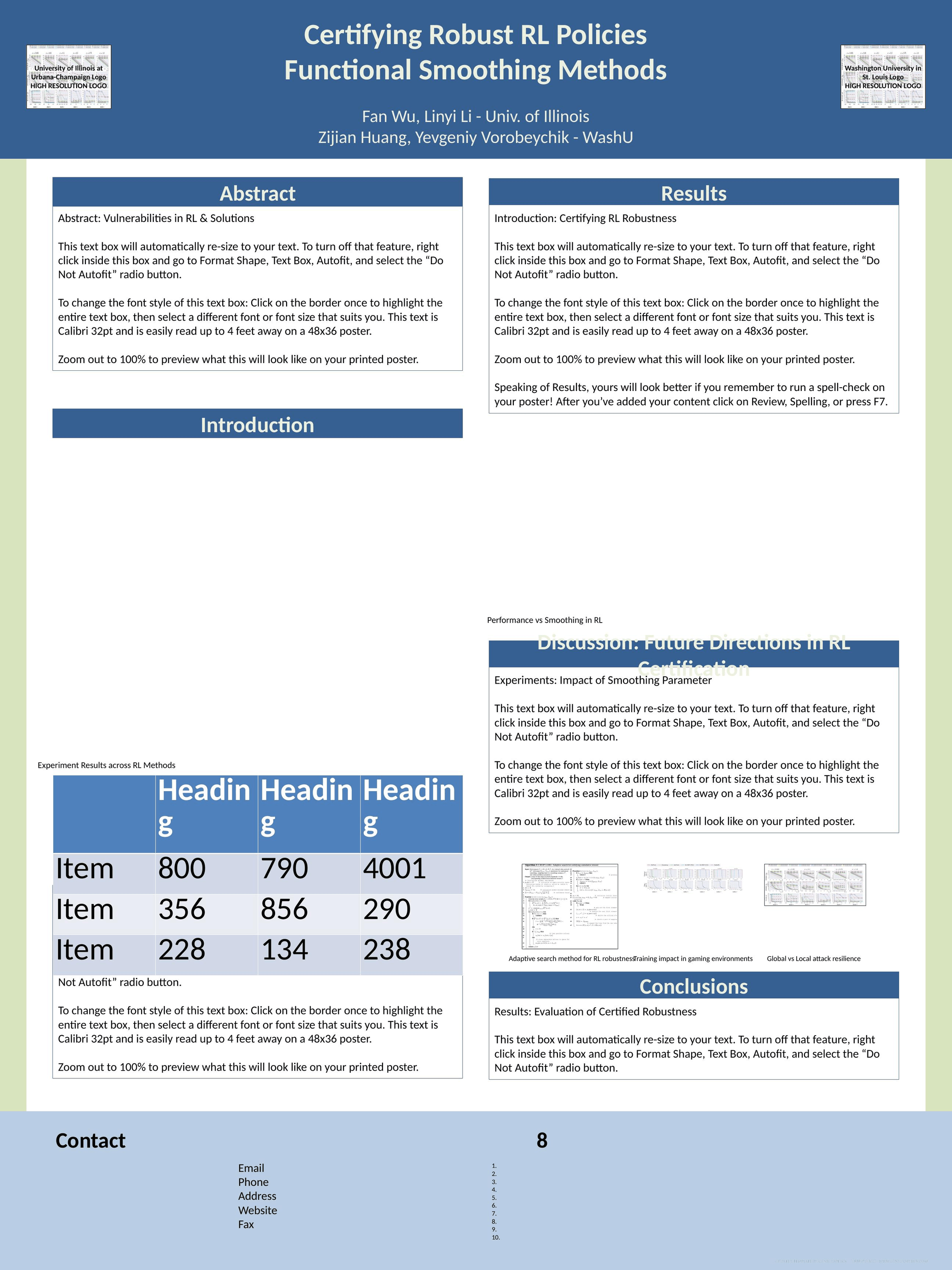}
      \subcaption{A poster generated by \texttt{PPTAgent}, where there is text overflowing outside textbox.}
      \label{fig:error_overflow_b}
    \end{subfigure}
  \end{tabular}
  \caption{Examples of posters with textual overflow.}
  \label{fig:error_overflow}
\end{figure*}

Unlike HTML, where nested boxes naturally clip overflow, the \texttt{PPTX} format lacks strict parent–child containment. Consequently, both \texttt{PPTAgent} and \texttt{\ouralg{}} sometimes produce text that spills beyond panel boundaries (Fig.~\ref{fig:error_overflow_b}, \ref{fig:error_overflow_a}). Among the \texttt{\ouralg{}} variants, the problem is relatively more pronounced in the \texttt{Qwen} variant, whose backbone (\texttt{Qwen2.5-VL-7b}) provides weaker visual grounding than \texttt{GPT-4o}, making its visual-feedback loop less reliable.

\FloatBarrier

\section{Prompt Templates}

\subsection{Baseline Prompts}
We exhibit the prompt templates used to generate baselines: \texttt{4o-Image}, \texttt{4o-HTML}, and \texttt{OWL-4o}.

\begin{tcolorbox}[prompt_4o_image]
Carefully analyze the provided research paper and design a professional, visually appealing academic conference poster. Include clear, informative text summaries, relevant figures, and tables that are neatly arranged and aligned. The poster should accurately represent the key findings, methods, and conclusions as if created by the original authors for presentation at a scientific conference. Ensure the design includes all essential elements commonly found in academic posters. The layout should be engaging, easy to follow, and visually attractive, balancing textual clarity with graphic effectiveness.  The poster should be of width {width}px and height {height}px. Generate through image generation.
\end{tcolorbox}

\begin{tcolorbox}[prompt_owl_4o]
Read the PDF file from: \\
\texttt{paper\_path}/paper.pdf
\\

Carefully analyze the provided research paper and design a professional, visually appealing academic conference poster. 
Include clear, informative text summaries, relevant figures, and tables that are neatly arranged and aligned.
The poster should accurately represent the key findings, methods, and conclusions as if created by the original authors for presentation at a scientific conference. 
Ensure the design includes all essential elements commonly found in academic posters. 
The layout should be engaging, easy to follow, and visually attractive, balancing textual clarity with graphic effectiveness. 

You should approach the task by generating and executing python-pptx code to create a single-slide PowerPoint presentation.
You should save your code, as well as the generated PowerPoint file.
\end{tcolorbox}

\begin{tcolorbox}[prompt_4o_html]
\textbf{System Prompt:} \\
\\
You are a document-to-poster generation agent.  
  Your task is to read the supplied Markdown text (\texttt{document\_markdown}) and
  design a professional, visually appealing academic conference poster by
  generating an HTML file.  
  Follow the guidelines below precisely.
\\

\textbf{Instructions}
\\

1. Carefully read the Markdown in \texttt{document\_markdown}.

  2. Design a full-page academic conference poster in HTML + CSS:
  
     \quad • Include a prominent header with title, authors, and affiliations.  [1ex]
     
     \quad • Break content into logical sections (Introduction, Methods, Results, Conclusions, etc.). 
     
     \quad • Provide clear, informative text summaries. 
     
     \quad • Embed relevant figures and tables, neatly arranged and aligned.  
     
     \quad • Accurately represent key findings, methods, and conclusions. 
     
     \quad • Ensure the layout is engaging, easy to follow, and visually attractive. 
     
     \quad • Include all essential poster elements commonly found at scientific conferences.
     
  3. Write complete HTML code (with inline or embedded CSS) that, when rendered,
     produces the poster layout.
     
  5. The poster width should be \texttt{poster\_width} px and height should be \texttt{poster\_height} px.
  
  4. **Output only** a JSON object with a single key "HTML", whose value is
     the entire HTML code for the poster.
\end{tcolorbox}

\subsection{Parser Prompts}
We exhibit prompt templates used for parser: (1) The LLM summarization prompt; (2) The figure filtering prompt.

\begin{tcolorbox}[prompt_parser_summarizer]
\textbf{System Prompt:} \\

You are a document content divider and extractor specialist, expert in dividing and extracting content from various types of documents and reorganizing it into a two-level json format for later poster generation.
\\

\textbf{Instruction:} \\

Based on given markdown document, generate a JSON output for later poster generation, make sure the output is concise and focused.
Step-by-Step Instructions:
1. Identify Sections and Subsections in document and identify sections and subsections based on the heading levels and logical structure.

2. Divide Content: Reorganize the content into sections and subsections, ensuring that each subsection contains approximately 500 words.

3. Refine Titles: Create titles for each section with at most 3 words.

4. Remove Unwanted Elements: Eliminate any unwanted elements such as headers, footers, text surrounded by "$\sim\sim$" indicating deletion.

5. Refine Text: For content, you should keep as much raw text as possible. Do not include citations.

6. Length: you should control the length of each section, according to their importance according to your understanding of the paper. For important sections, their content should be long.

7. Make sure there is a poster title section at the beginning, and it should contain information like paper title, author, organization etc.

8. The "meta" key contains the meta information of the poster, where the title should be the raw title of the paper and is not summarized.

9. Ther **must** be a section for the poster title.

Example Output:
\begin{verbatim}
{
    "meta": {
        "poster_title": "raw title of the paper",
        "authors": "authors of the paper",
        "affiliations": "affiliations of the authors"
    },
    "sections": [
        {
            "title": "Poster Title & Author",
            "content": "content of poster title and author"
        },
        {
            "title": "title of section1",
            "content": "content of section 1"
        },
        {
            "title": "title of section2",
            "content": "content of section 2"
        }
    ]
}
\end{verbatim}
\end{tcolorbox}

\begin{tcolorbox}[prompt_parser_filter]
\textbf{System Prompt:}\\

You are an assistant that reviews a poster's JSON layout (\texttt{json\_content}), along with corresponding \texttt{image\_information} and \texttt{table\_information}. Your task is to filter out any image or table entries that are irrelevant to the content described in \texttt{json\_content} (for instance, if their captions or any provided details do not align with the topics, sections, or content in the poster).\\

  Specifically:
  1. Read through the full poster data described in \texttt{json\_content}. \\
  
  2. Examine each entry within \texttt{image\_information} and \texttt{table\_information}. \\
  
  3. Decide if each entry is relevant based on its caption, path, or any other information provided. \\
  
    \quad - For example, if an image has a caption that obviously does not fit into any section or does not relate to the poster's content outline, deem it “unimportant.” \\
    
  4. Keep only those images/tables you consider "important" for the poster (i.e., relevant to the topics, sections, or discussions mentioned in \texttt{json\_content}). \\
  
  5. Produce an output containing just two keys: \\
  \texttt{"image\_information"} for the filtered images, and \texttt{"table\_information"} for the filtered tables. Each of these keys should map to an array of filtered objects. \\

  You must output valid JSON containing only:
  \begin{verbatim}
    {
      "image_information": {...},
      "table_information": {...}
    }
 \end{verbatim}

\textbf{Instructions:} \\

The user will provide JSON: \\

    1. \texttt{"json\_content"}: The content of the poster (sections, text, etc.). \\
    
    2. \texttt{"image\_information"}: A dict of images (each with caption, path, size constraints).  \\
    
    3. \texttt{"table\_information"}: A dict of tables (each with caption, path, size constraints). \\

  Your task: \\
  
    1. Read the poster outline (\texttt{json\_content}). \\
    
    2. Filter \texttt{image\_information} and \texttt{table\_information} so that only entries relevant to the poster content remain.  \\
    
      \quad • Relevance is determined by matching or relating their captions to the poster’s sections or content. \\
      
      \quad • If an image or table does not clearly match or support any content in \texttt{json\_content}, remove it. \\
      
    3. Return a JSON with the structure:
    \begin{verbatim}
       {
         "image_information": <filtered image information JSON>,
         "table_information": <filtered table information JSON>
       }
    \end{verbatim}

  Output Format: \\
  
  Just return a JSON object with the two keys:\\
  
  "\texttt{image\_information}" and "\texttt{table\_information}" — each containing the filtered data.
  No additional keys or text. Both "\texttt{image\_information}" and "\texttt{table\_information}" should present even if they are empty. \\

  Note:\\
  
  \quad • If no entries remain for either images or tables, just return an empty dict for that key. \\
  
  \quad • Keep at most 5 entries in \texttt{image\_information} and \texttt{table\_information} respectively. \\
  
  \quad • Make sure the JSON you output is valid.\\

  Please provide only the JSON object as your final output.
\end{tcolorbox}

\section{Planner Prompts}
We present the prompts used by the planner module, covering three components: (1) the asset matching prompt; (2) the painter prompt; and (3) the commenter prompt.

\begin{tcolorbox}[prompt_planner_matching]
\textbf{System Prompt}:

You are an expert assistant tasked with assigning images or tables to the most relevant poster sections. You will be given:
\begin{itemize}
  \item JSON content of the poster outline, including each section's title and a brief description.
  \item A list of images (image\_information) with captions and size constraints.
  \item A list of tables (table\_information) with captions and size constraints.
\end{itemize}

Your goal is to produce a JSON mapping of each top-level section to exactly zero or one image/table that best fits that section’s content. For each top-level section (named in the provided JSON “json\_content”), decide:
\begin{itemize}
  \item Whether an image or table (or none) is most relevant to the section's theme or description.
  \item If relevant, select the single most appropriate image or table to assign.
  \item Base this selection on the conceptual content described in the section (“research methods”, “results”, “conclusion”, etc.) and compare it with the captions of the provided images or tables, choosing whichever fits best.
  \item If assigning an image, specify “image”: \textless id\textgreater, where \textless id\textgreater is the identifier of the chosen image from “image\_information”.
  \item If assigning a table, specify “table”: \textless id\textgreater, where \textless id\textgreater is the identifier of the chosen table from “table\_information”.
  \item Include an additional “reason” field briefly explaining why this assignment was made (e.g., how the image/table relates to the section content).
  \item If no image or table is assigned to a given section, omit that section from the final JSON (i.e., only list sections where you actually assign something).
\end{itemize}

\textbf{Important Notes}:
\begin{itemize}
  \item The assignment should not be arbitrary. It must be logically consistent with the section’s description and the provided caption for the image or table.
  \item Do not produce any layout properties or subsections here.
  \item The final output must be a single JSON object, mapping from section names to the chosen image/table ID plus the “reason” field.
  \item If multiple images or tables are suitable, select the single best one and assign only that.
  \item If “image\_information” or “table\_information” is empty, you may end up assigning nothing to any section.
\end{itemize}

\textbf{Instructions}:
\begin{enumerate}
    \item Read and analyze the poster's top-level sections from \{\{ json\_content \}\}.
    \item Look at \{\{ image\_information \}\} and \{\{ table\_information \}\}. Determine content-fit:
      \begin{itemize}
       \item If a section's description or subject matter matches well with a given image/table caption, consider assigning it.
       \item If multiple images or tables seem relevant, choose the single best fit.
       \item If none of the images or tables are relevant, or if none are provided, do not assign anything for that section.
      \end{itemize}
    \item Produce a single JSON object. Each key is the exact name of a top-level section (e.g., "Introduction", "Methods", "Results"), and the value is an object with:
      \begin{itemize}
        \item \texttt{"image": image\_id} or \texttt{"table": table\_id}
        \item \texttt{"reason": short explanation describing why the image/table is assigned}
      \end{itemize}
    \item If no assignment is made for a section, exclude that section from the JSON.
    \item No image can be reused for multiple sections. Each image/table can only be assigned to one section.
    \item Ensure your final response strictly follows JSON syntax with no extra commentary.
\end{enumerate}

\textbf{Example Output Format}:
\begin{verbatim}
{
  "Introduction": {
    "image": 1,
    "reason": "Image 1 depicts the central concept introduced
    in this section."
  },
  "Results": {
    "table": 2,
    "reason": "Table 2 summarizes the key metrics discussed
    in the results."
  }
}
\end{verbatim}
\end{tcolorbox}

\begin{tcolorbox}[prompt_planner_painter]
\textbf{System Prompt}:

You are an expert assistant tasked with producing bullet-point summaries for a given poster section. You will be given:
\begin{itemize}
  \item A JSON object \texttt{summary\_of\_section} that contains:
    \begin{verbatim}
    {
      "title": "<section title>",
      "content": "<full text description>"
    }
    \end{verbatim}
  \item An integer \texttt{number\_of\_textboxes}, which can only be \texttt{1} or \texttt{2}.
\end{itemize}

Your goal is to produce a JSON object representing the bullet-point text for this poster section. Each “textbox” key (\texttt{textbox1} or \texttt{textbox2}) maps to a list of bullet-point entries. Each bullet-point entry must be a JSON object of the form:
\begin{verbatim}
{
  "alignment": "left",            
  "bullet": true,                 
  "level": <indent_level>,        
  "font_size": <integer>,         
  "runs": [
    {
      "text": "<bullet point text>"
      # optionally "bold": true or "italic": true if needed
    }
  ]
}
\end{verbatim}

\textbf{Instructions}:
\begin{enumerate}
    \item If \texttt{number\_of\_textboxes} = 1, your final output must only have:
\begin{verbatim}
{
  "title": [ section title ],
  "textbox1": [ ... array of bullet items ... ]
}
\end{verbatim}

    \item If \texttt{number\_of\_textboxes} = 2, then you must produce \emph{two} keys: \texttt{textbox1} and \texttt{textbox2}, and each must have the same number of bullet items. For example:
\begin{verbatim}
{
  "title": [ section title ],
  "textbox1": [... N bullet items ...],
  "textbox2": [... N bullet items ...]
}
\end{verbatim}
    where both arrays have \emph{identical length}.

    \item Each bullet point is a JSON object with the structure shown above; you can create as many bullet points as needed (following the constraint about textbox count).

    \item Make sure your final output is valid JSON, with no extra keys or additional formatting.

    \item Return only the JSON object, nothing else.
\end{enumerate}

\textbf{Example Output}:

\textit{Example when \texttt{number\_of\_textboxes} = 1:}

\begin{verbatim}
{
  "title": [
    {
      "alignment": "left",
      "bullet": false,
      "level": 0,
      "font_size": 60,
      "runs": [
        {
          "text": "Methodology",
          "bold": true
        }
      ]
    }
  ],
  "textbox1": [
    {
      "alignment": "left",
      "bullet": true,
      "level": 0,
      "font_size": 48,
      "runs": [
        {
          "text": "Key point about domain-invariant component analysis."
        }
      ]
    },
    {
      "alignment": "left",
      "bullet": true,
      "level": 1,
      "font_size": 48,
      "runs": [
        {
          "text": "Supporting detail.",
          "bold": true
        }
      ]
    }
  ]
}
\end{verbatim}

\textit{Example when \texttt{number\_of\_textboxes} = 2:}

\begin{verbatim}
{
  "title": [
    {
      "alignment": "left",
      "bullet": false,
      "level": 0,
      "font_size": 60,
      "runs": [
        {
          "text": "Experimental results",
          "bold": true
        }
      ]
    }
  ],
  "textbox1": [
    {
      "alignment": "left",
      "bullet": true,
      "level": 0,
      "font_size": 48,
      "runs": [
        {
          "text": "Primary finding, bullet 1."
        }
      ]
    },
    {
      "alignment": "left",
      "bullet": true,
      "level": 0,
      "font_size": 48,
      "runs": [
        {
          "text": "Primary finding, bullet 2."
        }
      ]
    }
  ],
  "textbox2": [
    {
      "alignment": "left",
      "bullet": true,
      "level": 0,
      "font_size": 48,
      "runs": [
        {
          "text": "Additional commentary, bullet 1."
        }
      ]
    },
    {
      "alignment": "left",
      "bullet": true,
      "level": 0,
      "font_size": 48,
      "runs": [
        {
          "text": "Additional commentary, bullet 2."
        }
      ]
    }
  ]
}
\end{verbatim}
\end{tcolorbox}

\begin{tcolorbox}[prompt_planner_commenter]
\textbf{System Prompt}:
You are an agent that is given three images:
\begin{itemize}
    \item \textbf{Negative Example}: This image shows a bounding box with text overflowing outside it (i.e., text crossing or cut off by the box).
    \item \textbf{Positive Example}: This image shows a bounding box with text that fits completely (i.e., no text crossing or cut off).
    \item \textbf{Target Image}: This is the final image you must analyze.
\end{itemize}

From the first two images, you learn to interpret:
\begin{enumerate}
    \item Whether text is overflowing (text crossing, cut off, or otherwise cannot fully fit in the box).
    \item Whether there is too much blank space in the bounding box (i.e., the text is significantly smaller than the box, leaving large unused space).
    \item Whether the text and bounding box are generally well-aligned (no overflow, no large blank space).
\end{enumerate}

Then, for the \textbf{Target Image}, you must:
\begin{itemize}
    \item If there is any overflow text, return \texttt{"1"}.
    \item If there is too much blank space, return \texttt{"2"}.
    \item If the text fits well (no overflow, no large blank space), return \texttt{"3"}.
\end{itemize}

\vspace{1ex}
\textbf{Instructions}:
\begin{enumerate}
    \item You are provided three images (negative example, positive example, and target).
    \item Refer to the first two images (negative and positive examples) to understand:
        \begin{itemize}
            \item What text overflow looks like
            \item What too much blank space in a bounding box means
            \item How a generally well-fitted bounding box appears
        \end{itemize}
    \item Analyze the third (Target) image's bounding box to check:
        \begin{itemize}
            \item If there is overflow text, return \texttt{"1"}.
            \item If there is too much blank space, return \texttt{"2"}.
            \item Otherwise (if everything looks good), return \texttt{"3"}.
        \end{itemize}
\end{enumerate}
\end{tcolorbox}

\section{Failure by Diffusion Models}
\begin{figure}[!h]
  \centering
  \includegraphics[width=0.6\linewidth]{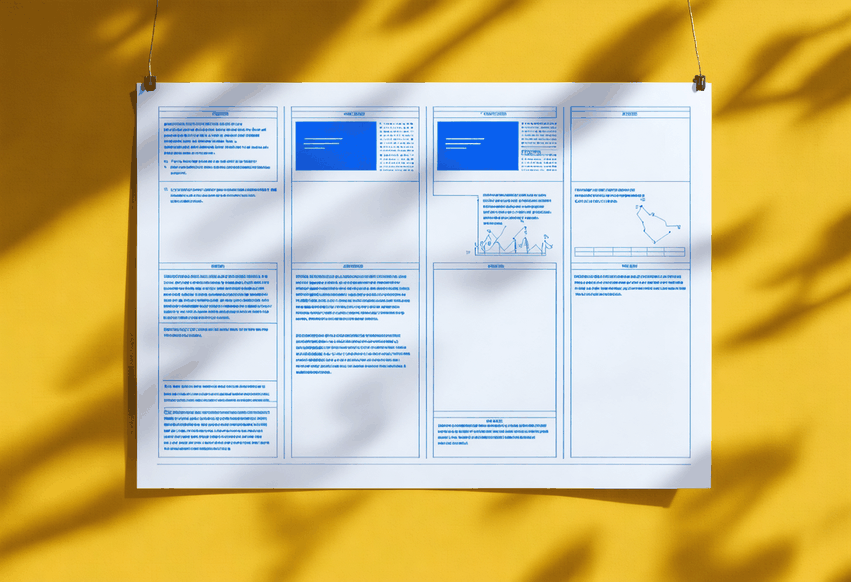}%
  \\[1ex]

  \includegraphics[width=0.6\linewidth]{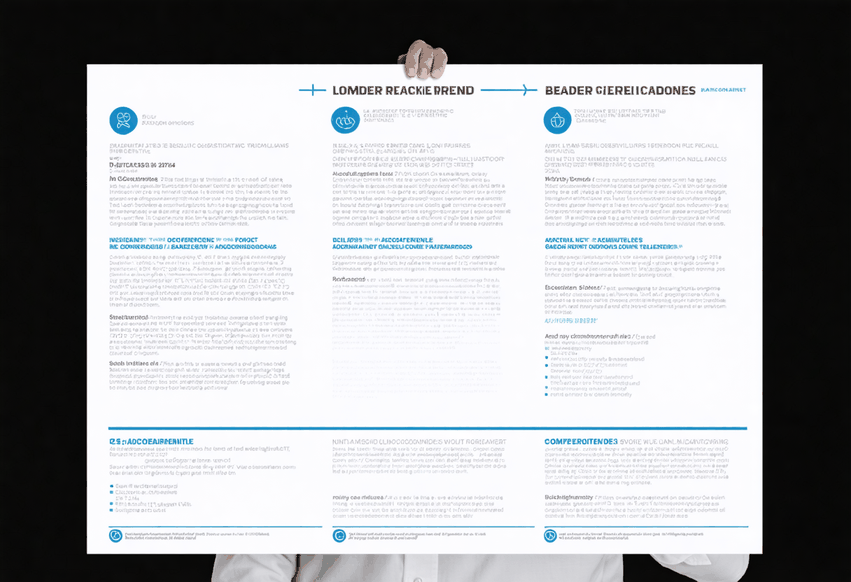}%
  \\[1ex]

  \includegraphics[width=0.6\linewidth]{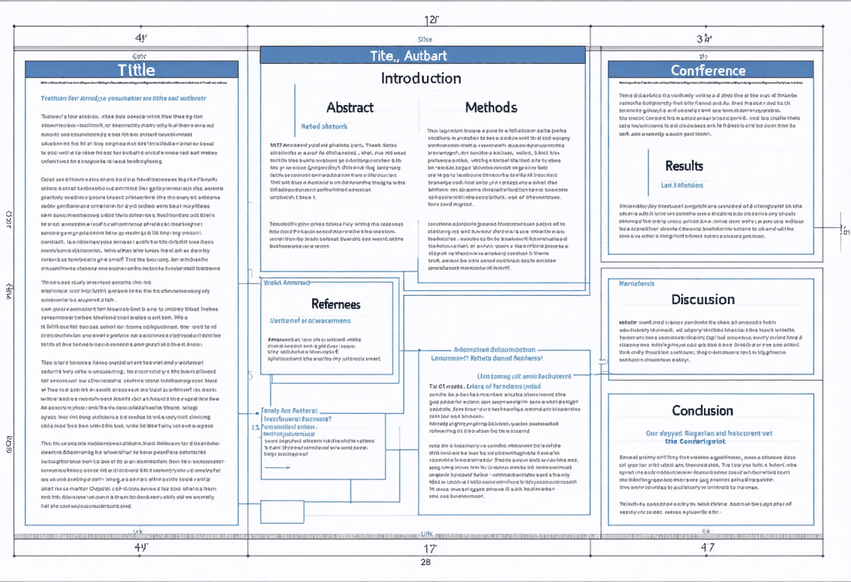}%

  \caption{Failure generation examples by Stable Diffusion Ultra model~\cite{stableultra}.}
  \label{fig:diffusion}
\end{figure}

In Fig.~\ref{fig:diffusion}, we illustrate failure cases of Stable Diffusion Ultra~\cite{stableultra}.
We found that diffusion models suffer from the issues listed below and remain far from adequate for academic poster generation:
\textbf{(i) Severely inaccurate text rendering} – Generated text often appears blurry, misspelled, or semantically incoherent, failing to meet title, body, and caption requirements.
\textbf{(ii) Unpredictable layouts} – Models cannot reliably partition the page or align content blocks, resulting in a disorganized visual hierarchy.
\textbf{(iii) Inconsistent styling} – Fonts sizes, spacing lack controllable parameters, making it impossible to conform to template guidelines.

\section{Illustration of In-context reference for Commenter}
\begin{figure}[!h]
  \centering
  \begin{subfigure}[b]{0.45\linewidth}
    \centering
    \includegraphics[width=\linewidth]{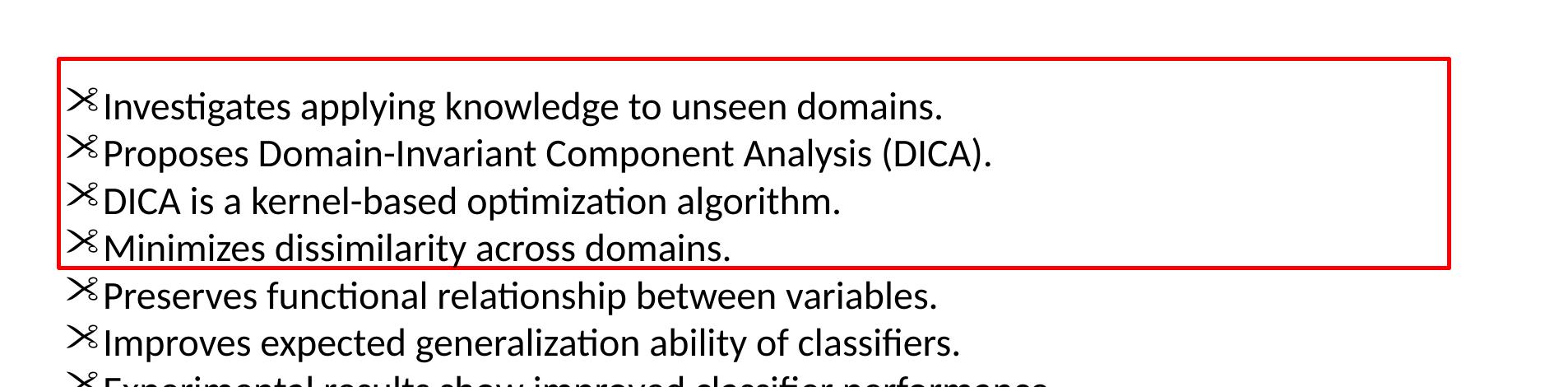}
    \caption{Negative examples}
    \label{fig:img1}
  \end{subfigure}
  \hfill
  \begin{subfigure}[b]{0.45\linewidth}
    \centering
    \includegraphics[width=\linewidth]{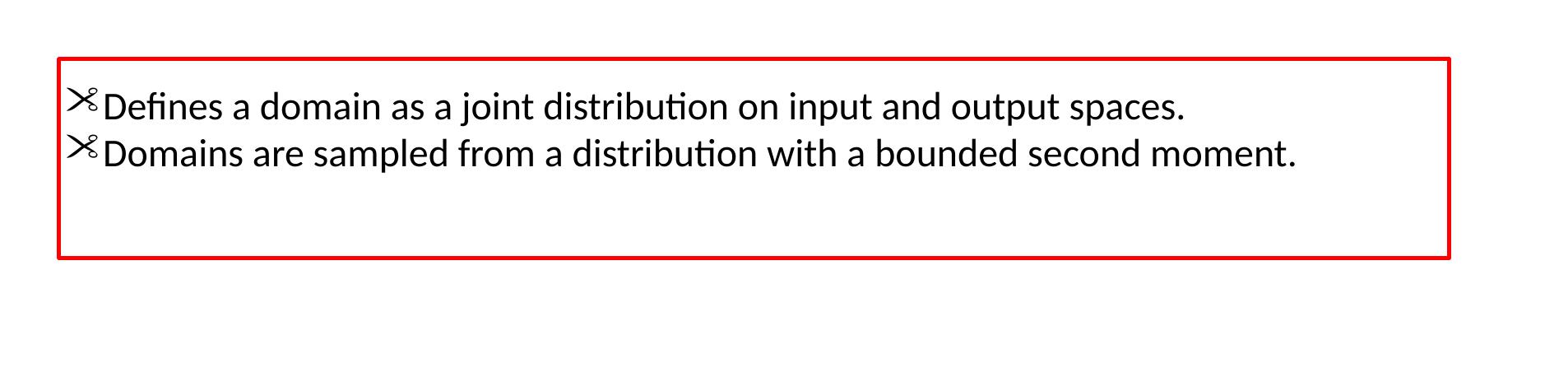}
    \caption{Positive examples}
    \label{fig:img2}
  \end{subfigure}
  \caption{In-context references for the commenter help the VLM better identify whether the current panel falls into a failure case.}
  \label{fig:incontext}
\end{figure}

In Fig.~\ref{fig:incontext}, we illustrate the in-context references used by our commenter during panel refinement to avoid undesirable cases such as “overflow,” “too blank,”.
These examples are highlighted by a red box as a visual prompt.

\end{document}